\documentclass[10pt,twocolumn,letterpaper]{article}

\makeatletter
\@namedef{ver@everyshi.sty}{}
\makeatother

\usepackage[pagenumbers]{cvpr} 
\usepackage{graphicx}
\usepackage{amsmath}
\usepackage{amssymb}
\usepackage{booktabs}

\usepackage[most]{tcolorbox}
\newtcbox{\mymath}[1][]{%
    nobeforeafter, math upper, tcbox raise base,
    enhanced, colframe=blue!30!black,
    colback=blue!30, boxrule=1pt,
    #1}
\usepackage{empheq} 
\definecolor{lightgreen}{HTML}{90EE90}

\usepackage[rightcaption]{sidecap}
\usepackage[ruled]{algorithm2e} 

\usepackage{enumitem}

\usepackage[pagebackref,breaklinks,colorlinks]{hyperref}

\usepackage[capitalize]{cleveref}
\crefname{section}{Sec.}{Secs.}
\Crefname{section}{Section}{Sections}
\Crefname{table}{Table}{Tables}
\crefname{table}{Tab.}{Tabs.}

\def\cvprPaperID{} 
\def\confName{Automation in Construction}
\def\confYear{2022}

\begin{document}

\title{Vitruvio: 3D Building Meshes via Single Perspective Sketches}
\author{Alberto Tono\\
Stanford University \& Computational Design Institute\\
{\tt\small atono@stanford.edu \& alberto.tono@cd.institute}
\and
Heyaojing Huang\\
Stanford University\\
{\tt\small hhyj4495@stanford.edu}
\and
Ashwin Agrawal\\
Stanford University\\
{\tt\small ashwin15@stanford.edu}
\and
Martin Fischer\\
Stanford University\\
{\tt\small fischer@stanford.edu}
}

\twocolumn[{
\maketitle

\begin{center}
\captionsetup{type=figure}
    \includegraphics[ width=1 \textwidth]{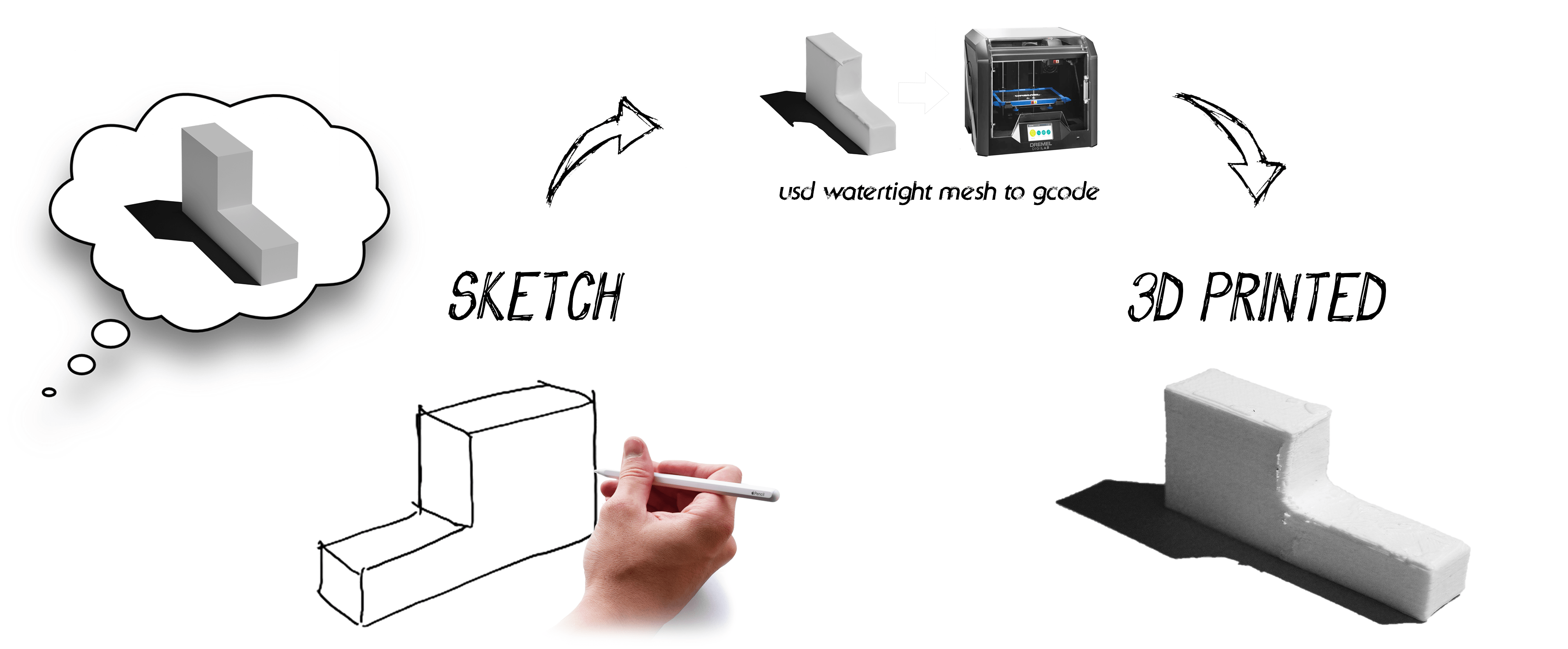}
    \captionof{figure}{Vitruvio converts a single perspective sketch to a 3D watertight mesh in universal Scene Description (USD) format. Vitruvio's final output consists of a 3D printable model. In the figure above, the model has been printed with DREMEL 3D45, PLA material from a $.gcode$ shape file. The users envision a 3D building mass (the ground truth building on the left, representing the "D2X" building in the dataset). Then, they sketch it on an iPad with a single-line style, centered on a squared canvas. This diagram wants to show some of the assumptions and limitations. The model has been trained only with a single-line synthetic style for the assumptions. For the limitations, the output mesh lacks accurate dimensions and proportions, as presented in Section $\ref{limitation}$.}
\end{center}
}]




\begin{abstract}

At the beginning of a project, architects convey design ideas via quick 2D diagrams, front views, floor plans, and sketches. Consequently, many stakeholders have difficulty visualizing the 3D representation of the building mass, leading to varied interpretations and inhibiting a shared understanding of the design. To alleviate the challenge, this paper proposes a deep learning-based method, Vitruvio, for creating a 3D model from a single perspective sketch. This allows designers to automatically generate 3D representations in real-time based on their initial sketches and thus communicate effectively and intuitively to the client. Vitruvio adapts Occupancy Network to perform single view reconstruction (SVR), a technique for creating 3D representations from a single image. Vitruvio achieves : (1) 18\% increase in the reconstruction accuracy and (2) 26\% reduction in the inference time compared to Occupancy Network on 1k buildings provided by the New York municipality. This research investigates the effect of the building orientation in the reconstruction quality, discovering that Vitruvio can capture fine-grain details in complex buildings when their native orientation is preserved during training, as opposed to the SVR's standard practice that aligns every building to its canonical pose. Finally, we release the code.

\end{abstract}

    



\section{Introduction}

The design process in Architecture, Engineering, and Construction (AEC) industry involves many stakeholders, including professionals such as engineers, architects, planners, and non-specialists such as clients, citizens, and users. Each stakeholder contributes all the design aspects, which Vitruvius called `Firmitas, Utilitas, Venustas', which translates as solidity, usefulness, and beauty. Early in the design process, all parties must reach a shared understanding of these Vitruvian values to avoid any misrepresentation later on \cite{bouchlaghem2005visualisation, yu2022systematic}. A critical factor in establishing a shared understanding is the ability to convey the information quickly, using a medium all stakeholders can understand \cite{sketchdanhaivecaitlin}.

However, during the initial meetings, design ideas are shared with mediums such as 2D diagrams \cite{buildingan}, front views, floorplans \cite{housegan,housegan++,SketchOpt, housediffusion }, and sketches on papers \cite{Encodedmemory}. These mediums often represent the design information in a few lines, leading to partial and incomplete representations of the overall mass. As a result, many stakeholders need help to visualize the actual 3D representation of the building, resulting in varied interpretations, which ultimately inhibits a shared understanding of design. \cite{pakhale2020digital} notes that the inability of the stakeholders to interpret 2D designs leads to reductions in productivity, reworks, wastage, and cost overruns. \cite{khanzode2000effect} points out that this mode of design practices leads to difficulties in the communication of designs since these representations lack of the 3D information ( such as proportion, volume, overall mass, and others) needed during later phases.

To alleviate this challenge, this research aims to generate 3D geometries from sketches, grounding its theory on Sketch Modeling, an active area of research since the 90's \cite{teddy1999,sketchsurvey}. Sketch modeling has two major approaches: Learning-based methods and Non-learning-based methods. The Non-learning based methods require specific and defined inputs to construct 3D geometries. As a result, this method operates with fixed viewpoints and specific sketching styles, thus reducing the designer's flexibility. Therefore learning-based methods have been employed to resolve these issues, allowing for more flexibility, as detailed later in Section \ref{method}. 
Learning-based methods, also called data-driven, generate a 3D shape from a partial sketch by learning a joint probabilistic distribution of a shape and sketches from the dataset. 


Currently, these techniques have only focused on decontextualized shapes such as furniture and mechanical parts, where positions and orientations do not directly affect their representation.
However, this is different for buildings. Their design is affected by the building's location and orientation



Therefore, this research provides a step forward in this direction to consider location and orientation within the reconstruction process for deep learning models. Indeed, Vitruvio is a \textit{flexible}, and \textit{contextual} method that reconstructs a 3D representation of buildings from a single perspective sketch. It provides the flexibility to generate a building mass from a partial sketch drawn from any perspective viewpoint. To accomplish this tasks we build our own dataset (Section \ref{dataset} ) dubbed Manhattan1k. Manhattan1k preserves  the contextual information of building, specifically their locations and orientations. 

To summarize, the contribution is threefold: 

\begin{enumerate}
    \item We explain the use of a learning-based method for sketch-to-3D applications where 
    the final 3D building shapes depend on a single perspective sketch.  
     \item We develop Vitruvio adapting Occupancy Network (\textit{OccNet}) \cite{Occnet2019} to our buildings dataset and improving its accuracy and efficiency ( Section \ref{accandeff} ).
    \item We show qualitatively and quantitatively that the building orientation affects the reconstruction performance of our network (Section \ref{context}). 
   
\end{enumerate}

After presenting the related works and their limitations in Section \ref{relatedwork}, the remainder of this paper introduces our methods in Sections \ref{method}, \ref{dataset} and the relative experiments to validate the above-mentioned hypothesis and claims in Section \ref{experiments}. Finally, Sections  \ref{discussion} ,\ref{limitation}, and \ref{conclusion} present the discussion and limitations of our method. The code has been released open source: \url{https://github.com/CDInstitute/Vitruvio}

\section{Related Work}\label{relatedwork}

This section introduces previous work in sketch modeling and the limitations that led to a learning-based approach. 

For most, sketch modeling has been an active area of research since the late 90's \cite{teddy1999, sketchsurvey}. This modeling process can be represented in two ways: as a series of operations or as a Single View Reconstruction (SVR) \cite{whatsvrlearn, SVRsurvey}.

The former is typically adopted by CAD software. It requires specific and defined inputs, such as strokes, to construct 3D geometries. Through a series of sketches from different viewpoints \cite{Delanoy20203DJohanna} , or a series of procedures \cite{Nishida2016, procedural_model, Free2CAD, sketchrnn,vitruvion, Li2020Sketch2CAD}, these simplified CAD interfaces , provide complete control of the 3D geometry, at the expense of the artistic style. Thus, the models developed have been view, and style-dependent \cite{deepsketchmodeling}, operating with fixed viewpoints and specific sketching styles.

 The latter, SVR, leverages a more \textit{flexible} approach. It uses computer vision techniques to reconstruct 3D shapes from a single sketch without the mandatory requirement of a digital surface. SVR has recently gained attention thanks to the advances in learning-based methods \cite{differentiablerendering, neuralrendering,freehandreconstruction,OpenSketch19,deepsketchmodeling,zhong2020towards, deepsketchmodeling,sketch2mesh}, inspired by image-based reconstruction where the geometric output is represented in two main ways: explicitly or implicitly. An explicit representation is composed of meshes \cite{image3drecon,atlasnet,wen2019pixel2mesh} , point clouds \cite{pointnet,pointnet++,dgcnn,psgn16}, voxels \cite{OctNet,3dr2n2,marrnet}, sets of semantically meaningful parts \cite{Superquadrics}, constructive solid geometries (CSG) \cite{Free2CAD, Li2020Sketch2CAD}  coons patch-based \cite{smirnov2021learning}, superquadrics \cite{Superquadrics}. The implicit ones are represented as occupancy function \cite{Occnet2019,dreamfields_zeroshot,im-net}, neural fields, or signed distance fields \cite{deepSDF,Takikawa2022SDF,disn,ansari2021doccnet,ldif,pifu,oechsle2021unisurf,reiser2021kilonerf,xu2022sinnerf,lin2022neurmips,mueller2022instantNerf}. 


Previous SVR approaches focused on decontextualized shapes \cite{chang2015shapenet,BuildingNet,modelnet}. Indeed, furniture \cite{chang2015shapenet,modelnet} and mechanical parts \cite{mechanicalpartspurdue,abc,fusion360gallery} can be positioned anywhere and do not have to be designed for an exact location. However, this is different for buildings. Their construction and design have specific constraints and regulations that vary based on location. In a specific neighborhood, the buildings share similar limitations, features, and characteristics. In the initial design phases, the project location is known, and it is used in common data-driven approaches \cite{datadrivendesign} with Geographical Information Systems (GIS) \cite{GISbaseddesign,comocambridge} to inform the design process, thereby \textit{contextualized}. In fact, the location and orientation highly impact the design. Therefore, the 3D building mass generation from a sketch should account for those factors, and they should be present in the dataset. For example, considering a specific building in New York, the wind impacts its design: a five-degree rotation affects its energy and structural performance. 



Due to these desiderata, a deep learning model. It captures the underlying correlation between the 2D sketch and the 3D building shape from a bayesian perspective, without the need to follow a deterministic process and with the ability to encode additional information such as building location and orientation. Hence, the dataset is the key to these data-centric learning-based approaches. Unfortunately, recent single view sketch-based methods focused on datasets like ShapeNet \cite{chang2015shapenet}, and only two targeted the reconstruction of buildings \cite{Nishida2016} and \cite{Delanoy20203DJohanna} from perspective sketches. These examples either targeted the content generation communities (gaming and mapping scenarios) \cite{Nishida2016}, or did not allow a 3D reconstruction based on a single perspective sketch \cite{Delanoy20203DJohanna, SketchOpt,sketchdanhaivecaitlin}, sub-optimal for AEC's workflows. While previous generative design approaches required an explicit formulation of constraints and parameters to generate new solutions, our method can synthesize and learn the generative process from existing buildings \cite{multimodalcondsynthesis}.
For this reason, we develop Vitruvio. Vitruvio is a deep generative model: a learning-based approach approximating a specific dataset's probability distribution. Our dataset comprises existing 3D building shapes, sketches, and contextual information (orientation).


\section{Method}\label{method}

In this research, we adopt a learning-based \cite{componet2021} method that better aligns with our desiderata, as previously described. Deep generative models, Variational Auto-Encoder (VAE) \cite{vae,stocbackvi,neuralprocesses,infovae}
, Auto-Encoder (AE), Generative Adversarial Network (GAN) \cite{buildingan,marrnet}, Flow-based \cite{cflow}, Energy-based, Score-Based or Diffusion Model \cite{pointclouddiffusion,pointvoxeldiffusion}) aim to approximate an unknown joint probability distribution. In fact, our approach does not learn to directly map 2D images to 3D \cite{atlasnet,pixel2mesh,3dr2n2}. It estimates the joint distribution \cite{marrnet,deepSDF,vae,Occnet2019,cflow}  of three main random variables $p( x, y, \phi)$ : the building shapes $x^{(i)}$, the respective sketches $y_{ij}$, and the contextual information $\phi_{il}$ ( building orientation and position). This process enables to sample new shapes from this learned distribution described as $p_D$ (where $D$ is the dataset $x, y, \phi$).

\begin{figure}[ht]
  \centering
  \graphicspath{ {./images/} }
  \includegraphics[width=1\linewidth]{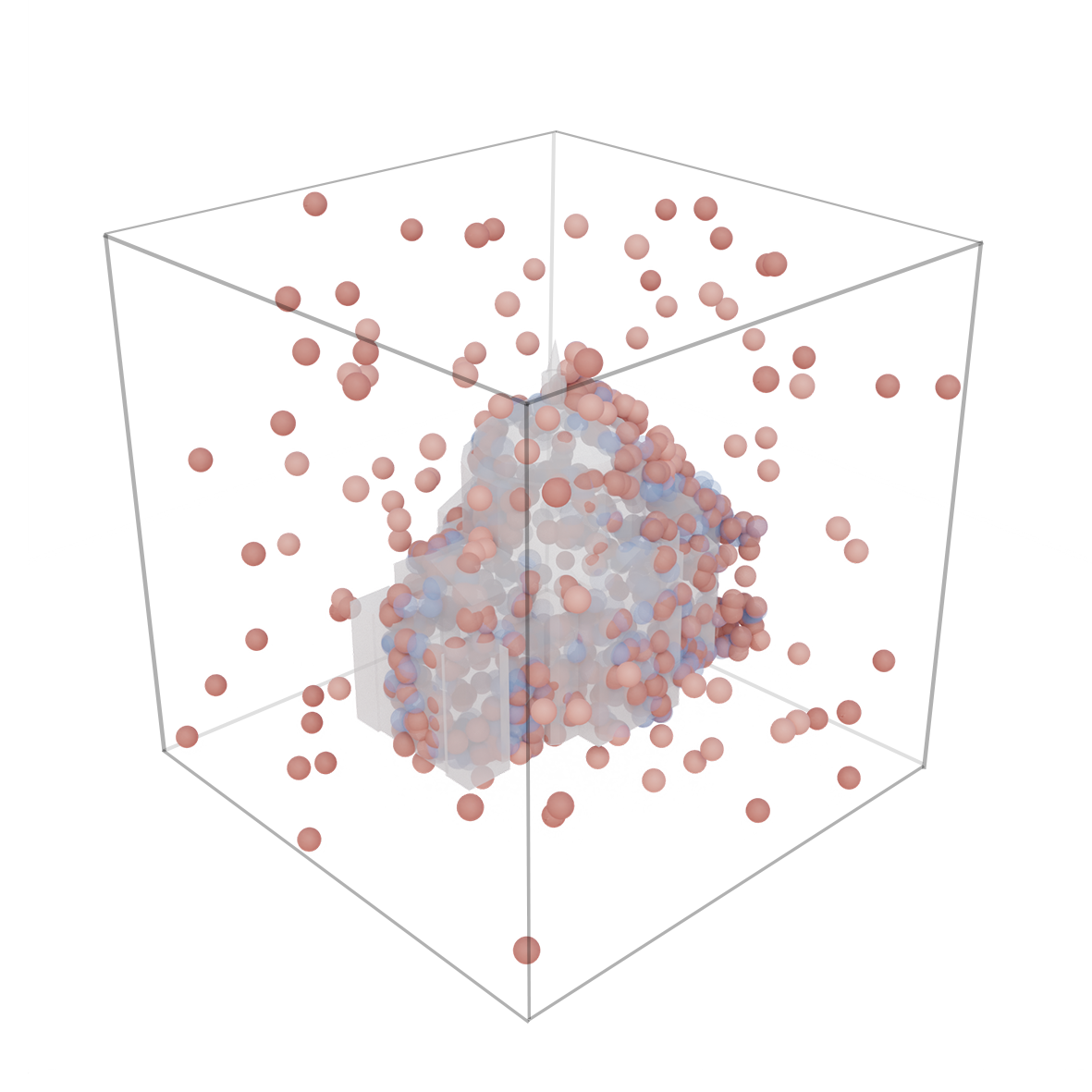}
   \caption{Sampling and visualization of the unsigned occupancy function. Blue points correspond to internal and red to external samples. In this image, points are represented as spheres with a 0.01 unit radius. The building shape has been previously normalized in a unit interval.}
   \label{fig:sdf}
\end{figure}

\begin{figure*}[ht]
  \centering
  \graphicspath{ {./images/} }
  \includegraphics[width=1\linewidth]{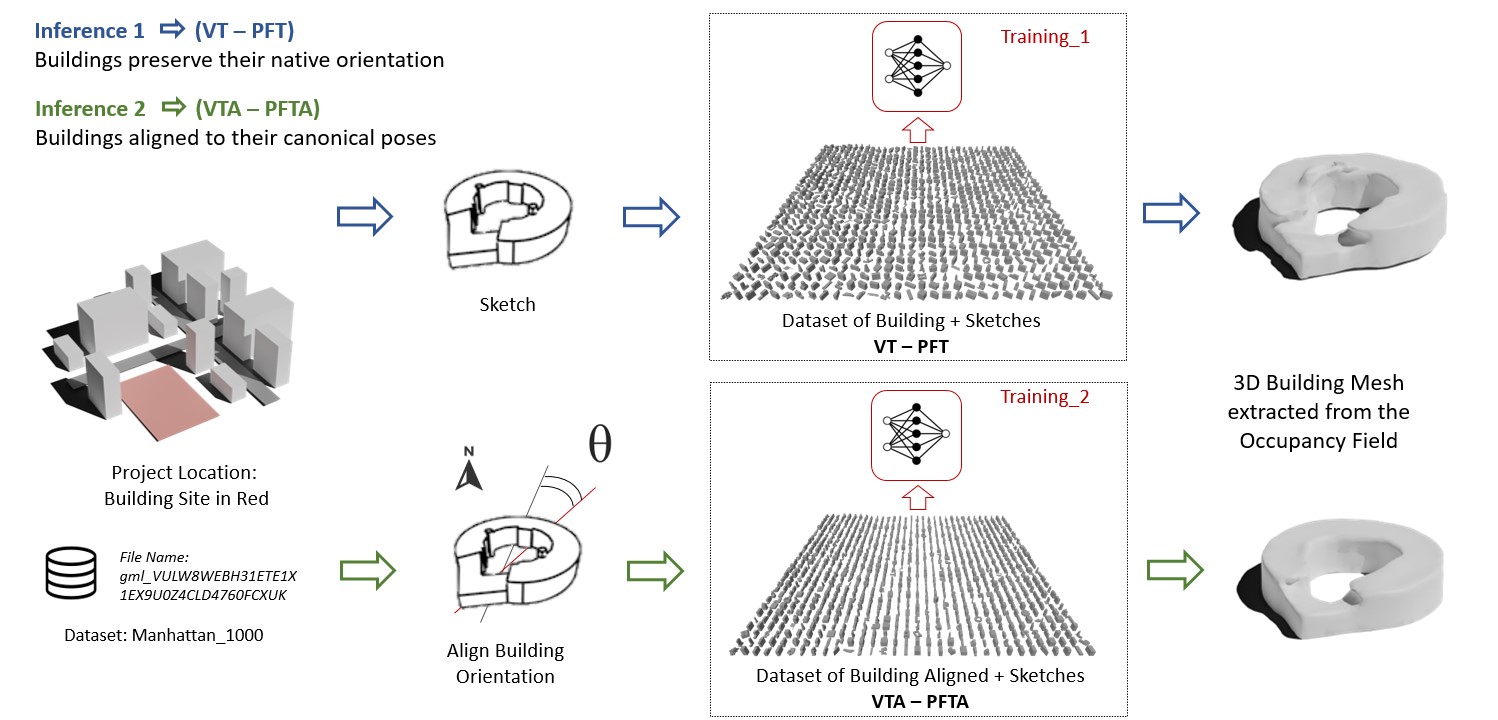}
   \caption{Diagram of our envisioned design workflow. The project starts from a known project location. Then, the designer sketches multiple design options. Vitruvio converts them to 3D printable meshes. We execute the training process with two different datasets. One preserves the initial orientation, and the other where the building orientation $\theta$ is stored and the buildings are aligned in a canonical pose.}
   \label{fig:workflow}
\end{figure*}

First the $x^{(i)}$ represents the  3D building shape $i^{th}$ in our dataset composed by $n$ independently sampled buildings  $\left\{x^{(1)}, \ldots, x^{(n)}\right\}$. For this approach, we follow \cite{Occnet2019} and $x \in \mathbb{R}$, where each 3D point ( $\mathbb{R}^3$ ) is represented by an occupancy function that outputs $1$ if that location is inside the shapes, and $0$ otherwise. This function is represented as a classification problem: a Neural Network outputs 1 or 0 values from $xyz$ coordinates ( or $p $  where $p$ is $p_{i k} \in \mathbb{R}^{3}, k=1, \ldots, K$ in the building shape $i^{th}$ ) and a sketch.

Second, the sketches are represented as $y_{ij} \in \mathbb{R}^2$ (as grayscale representations),  where $i$ represents the building and $j$ is the sketch viewpoint.

Third,  $\phi_{il}$ constitutes the contextual, surrounding factors $l^{th}$; in our case, we considered the orientation and location of the building $i^{th}$. 

The buildings are represented as independent 3D shapes by the occupancy network $f_{\theta}(x)$ where $
f_{\theta}: \mathbb{R}^{3} \times \mathbb{R}^{2}\rightarrow[0,1]
$, but after properly encoding the sketch and the orientation into a finite latent variable $z$ (encoder for $\phi, y$), the probability of the building shape $x$ based on the NN parameters $\theta$ can be represented  $p(x;\theta) = \sum_z p(x,z;\theta) $. We can derive the marginal and conditional distribution from this joint probability approximation to better serve downstream inference tasks such as sketch modeling and shape generation \cite{crossmodal3Dshapegen}.


This framework allows the generation of new samples (Predictive Posterior) that mimic the training distribution (Likelihood). Thus, new building shapes are generated without developing specific generative algorithms, with constraints and parameters, but by simply sampling from the learned distribution.

This sampling procedure is conditioned on a sketch (similar to conditional inference processes such as Pumarola et al.  \cite{cflow} Chan et al. \cite{EG3D}, or Ramesh et al. \cite{dalle2}).

For the chain rule, as in VAE\cite{neuralprocesses,stocbackvi,vae,generativeordiscriminative,cvae,3DP3,GSGN} , $p(x)$ is conditioned to the latent vector $z$ ( representing the sketches  \cite{VAEbottlenecked}) :  $$ p( y, \phi , x; \theta) =  p( z, x; \theta) = p( z | x; \theta)p(x; \theta) = p( x; \theta | z)p(z)  $$



Since the posterior $p( z | x; \theta)$ (represented with an encoder) is often intractable,  VAE uses a deep network to represent $p( x; \theta | z)$ as the decoder.

In \textit{OccNet} \cite{Occnet2019} and IM-NET \cite{im-net, neuralfieldsvision}, $ p( x; \theta | z)p(z)  = p_{\theta}(x, z) \equiv p(z) p_{\theta}(x \mid z)$  the decoder $p(x|z)$ overlooked additional factors such as structure \cite{structureNET} physics \cite{Phys-DeepSDF} or contextual variables that influence the latent variable $z$  (embedding). This could be the cause of their poor generalization. The decoder produces the 3D shape $x$ conditioned on the embedding $z$ (probabilistic latent variable model), which encodes the sketch and other information.

The inputs of the \textit{OccNet}'s encoder are $p_{ij}$ and the occupancy values $o_{i j}$, $ij$ as in Fig \ref{fig:sdf}. This encoder predicts the mean and the standard deviation of a Gaussian distribution (posterior distribution) $q_{\psi}\left(z \mid\left(p_{i j}, o_{i j}\right)_{j=1: K}\right)$ on $z \in \mathbb{R}^{L}$ with $L$ representing the dimension of the embedding and $z$ the conditioning on the sketch. In this way, it is assumed that $p(z)$ has a simple (Gaussian $\mathcal{N}\left(\mu, \sigma^{2}\right)$) prior distribution over the features.


Our goal, as in \textit{OccNet} \cite{Occnet2019},  is to optimize the variation of the Evidence Lower Bound (ELBO) of the Negative Log-Likelihood (NLL) of a generative model \ref{supplementary}, allowing a joint training of the encoder and decoder network. 
$p\left(\left(o_{i j}\right)_{j=1: K} \mid\left(p_{i j}\right)_{j=1: K}\right)$:

$$
\begin{array}{r}
\mathcal{L}^{\text {gen }}(\theta, \psi)=\sum_{j=1}^K \mathcal{L}(\underbrace{f_\theta\left(p_{i j}, z_i\right)}_{\text {decoder }}, o_{i j}) \\
+\mathrm{D}_{\mathrm{KL}}(\underbrace{q_\psi\left(z \mid\left(p_{i j}, o_{i j}\right)_{j=1: K}\right)}_{\text {encoder }} \| p_0(z))
\end{array}
$$

where $\mathrm{D_{K L}}$ denotes the KL-divergence, between $p_{0}(z)$ and $z_{i}$. Here $p_{0}(z)$ [marginal of $p(x,z;\theta) $] is the prior distribution on the latent variable $z_{i}$ (typically Gaussian, with reparametrization trick to ensure the differentiability of the sampling). And $z_{i}$ is sampled according to $q_{\psi}\left(z_{i} \mid\left(p_{i j}, o_{i j}\right)_{j=1: K}\right)$.

$$
\log P(x \mid c)-D_{K L}[Q(z \mid x, c) \| P(z \mid x, c)]= \\
$$
$$
\underbrace{E[\log P(x \mid z, c)]}_{decoder}-\underbrace{D_{K L}[Q(z \mid x, c) \| P(z \mid c)] }_{encoder}
$$

Here we maximize the conditional log-likelihood, noticing that the goal is not only to model the complex distribution of the shapes for buildings but also to make a discriminative prediction based on the input sketch and contextual information. Specifically, different buildings could be generated from the same image based on their different contextual information, such as location and orientation. For example, a sketch of a cube could generate different 3D models based on the weather conditions of the building site. In a warm location, the cube could have an atrium \cite{atrium} to guarantee more sunlight, air circulation, and shading \cite{danhaive21}. In a cold location, to better preserve the heat, the atrium is not recommended. While the rest of the method follow the exact implementation of \cite{Occnet2019} with the same training procedure, losses and inference, our experiments are designed to validate our initial hypothesis and to show the potential of \textit{OccNet} for applications related to building design.


\section{Dataset}\label{dataset}

As mentioned in Section \ref{method}, a data-driven learning-based method is employed in this research. Hence, it requires a dataset to be trained on to approximate a joint distribution of a training corpus $D$ composed by $x, y, \phi$ . Initially, Federova et al.'s Synthetic Dataset \cite{federova} \textit{BuildingNet} \cite{BuildingNet}, \textit{Random3DCity} \cite{random3dcity}, \textit{3DCityDB (Berlin)} \cite{3dcitydb}  and \textit{Realcity3D} \cite{realcity3d} have been analyzed. BuildingNet's dataset and Federova et al.'s  \cite{federova} provide buildings with proper segmentation, but unfortunately, they lack contextual information, are too detailed for conceptual design phases, and misrepresent existing buildings. Instead, \textit{Realcity3D} \footnote{NY website https://www.nyc.gov/site/planning/data-maps/open-data/dwn-nyc-3d-model-download.page and AI4CE https://github.com/ai4ce/RealCity3D.} does not have sketch representations. We built a custom dataset of building masses with respective sketches and contextual information to validate our method.

\begin{figure}[hb]
  \centering
  \graphicspath{ {./images/} }
  \includegraphics[width=1\linewidth]{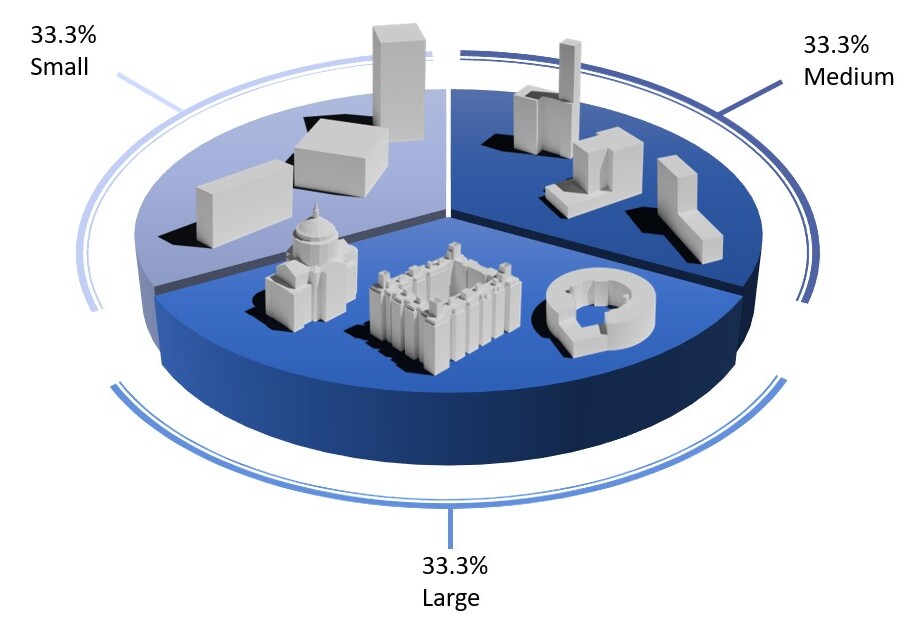}
   \caption{Dataset division of the 1k models based on file size. We used 333 small size (12Kb), 333 medium size (12-300Kb), and 334 large size ($\geq{334Kb}$)}
   \label{datasettable}
\end{figure}

 Therefore, from \textit{Realcity3D}, we extracted the $.obj$ of $46k$ buildings belonging to the municipality of New York. From these $46k$ ($45847$) buildings, we selected, based on their file size, a subset of only one thousand shapes from Manhattan ( we called this dataset \textit{Manhattan 1K}, in the GitHub repository we released the filenames of the buildings adopted). We divided the buildings into three main file sizes ( correlated to the levels of details 'LOD' \cite{3dfier, LOD} categories, the more details a shape has, the larger the file size to store that amount of information) to reduce this variation and minimize model variance within each class. For simplicity, we split the dataset based on the file size: small ($333$), medium ($334$), and large ($333$), as in Table [\ref{datasettable}] and data separation provided in the repository. Moreover, we randomly composed the training/validation/test sets in $700/ 100/ 200$. The training set, with $700$ buildings, used $16800$ synthetic sketches ($24$ sketches for each 3D building shape).

 Adapting the \textit{OccNet}'s approach \cite{Occnet2019}, we defined these building shapes as implicit representations \cite{neuralfieldsvision}.  The points are sampled uniformly from the bounding box of the mesh as depicted in Fig. \ref{fig:sdf}. 

 A function represented by a neural network constructs the iso-surface determining if the point is inside or outside the building mesh. This association is performed only with a watertight mesh (e.g., for measuring IoU). Following \textit{OccNet}'s approach, we implemented the code provided by \cite{Stutz_2018_CVPR}, which performs TSDF-fusion on random depth renderings of the object to create watertight versions of the meshes. We centered and re-scaled all meshes for voxelizations from \textit{3DR2-N2} \cite{3dr2n2}. The 3D bounding box of the mesh is centered at $0$, and its longest edge has a length of $1$. To find the iso-surface we sampled 100k points in the unit cube centered at $0$ with additional slight padding of $0.05$ units on both sides. After the sampling procedure, an algorithm counts the number of triangles that a $z$-ray (parallel to the $z$-axis) intersects. If this number is even, the point lies outside the mesh. Otherwise, it lies inside $o: \mathbb{R}^{3} \rightarrow\{0,1\}$, following a classification approach typical of a Bernoulli probability distribution. 
 
 On the contrary, during training, only a subsample of $2048$ points is employed from the original 100k points. While pre-processing our dataset accordingly to Stutz and Geiger's \cite{meshfusion,Occnet2019, meshlab}, some artifacts in the watertight meshes are produced, and more information is provided in the supplementary material \ref{supplementary}.

 Once our neural network has implicitly defined the geometrical representation of the building, we focused on the conditional features, such as contextual information and sketches, crucial to verify the hypothesis presented in the previous Section \ref{method}.

\textbf{Location and Orientation}. During the initial conceptual design phase, it is crucial to consider the building’s location and orientation. Location is influenced by the geographical, cultural, and geopolitical context in which the building is located, providing a more holistic view of the project. 
Instead, the orientation drives the buildings' energy efficiency and the occupants' indoor comfort, amongst other building characteristics. Furthermore, taking advantage of natural daylight impacts the heating/cooling systems increasing savings and human well-being. 
For the building location, the filename of each model in the dataset is associated with their exact position on Earth following the epsg:4326 convention according to the GeoDataFrame \cite{cityGML} Coordinate Reference System (CRS) \cite{proj}.  
For the orientation, we aligned the buildings with the bounding box to simplify the training process. Principal Component Analysis (PCA) based alignment has limitations when principal axes are not consistently oriented. These axes are unstable when principal values are similar. However, despite these limitations, it would be a good approach for our case, but an even more straightforward approach to extracting the larger side of the Oriented Bounding Box (OBB) containing the building works. We released a quick algorithm \ref{alg:aligment} \footnote{Trimesh \cite{trimesh}} to align the building to a canonical pose, keeping track of its orientation. A different approach could be performed, ensuring object-level rotation equivariance via deep network \cite{yu2022rotationally} within the network to directly bypass the \textit{alignment} phase \cite{Cosmo2022, condor, vectorneurons}.
 
This uniform canonical pose avoids creating artifacts in the voxelization.

\begin{algorithm}[hb]
\caption{Alignment algorithm } \label{alg:aligment}

\SetKwComment{Comment}{/}{/}
\KwIn{ $n$ 3D buildings  $\left\{x^{(1)}, \ldots, x^{(n)}\right\}$ }
\KwOut{ $n$ 3D buildings Aligned $\{ x_{\text {aligned }}^{(1)}, \ldots, x_{\text {aligned }}^{(n)} \}$ }

\While{$i \neq n$}{
         $V_{obb} $$\leftarrow$$ x_{obb}^{(i)}$ \Comment*{get OBB} 
        \For{$v$ in V}  
            {
                $v_{xy} \gets set$ $v_{z}=0 $\; 
            }
         $V_{xy} \leftarrow V_{obb}$ \Comment*[r]{ proj verts to $xy$}
         $P_{xy} = \int_{A_{xy}} (x,y) d A $ \Comment*[r]{inertia axis} 
         $I_{oy}, I_{ox} = P_{xy} $ \;
        $ \bar{y} = I_{oy} $ \;

        \eIf{ $\bar{y}_{y} , \bar{y}_{x} > 0$ or $\bar{y}_{y} < 0 \And  \bar{y}_{x} > 0$}
            {
             $-\bar{y}$ \;
             $  -R_{\theta}  \gets \theta_{y\bar{y}} $\;
             $ -R_{\theta}(x^(i)) = x_{aligned}^{(i)}$ \;
            }
            {
            $\bar{y}$ \;
             $  R_{\theta}  \gets \theta_{y\bar{y}} $\;
            $R_{\theta}(x^(i)) = x_{aligned}^{(i)}$ \;
            }

}
\end{algorithm}

\textbf{Synthetic sketches}: conscious of the limitations highlighted by previous work \cite{zhong2020towards,Wang:2021:Tracing,deepsketchmodeling}, and due to the cost of collecting a real sketch dataset, we adopted synthetic sketches. We followed the dataset generation from \cite{chang2015shapenet,3dr2n2} and we changed the renderer. Therefore, the synthetic sketches have been obtained after rendering the initial polygonal meshes \cite{polygen2020} in Blender \cite{blender}, combining a Lambertian representation of the building with a Suggestive Contours' filter \cite{suggestivecontours} to highlight the edges. Other approaches \cite{vinker2022clipasso,informativedrawings} did not provide a desirable representation and yielded poor results. Finally, we moved the cameras around the object (instead of moving the object), extracting $24$ camera extrinsic $R$, $T$ (Rotation, Translation), and intrinsic parameters $K$. Our light source was aligned with the camera position, and the building had identical appearance values (luminance) independent of the viewpoint, passing the photo-consistency test (with Lambertian's properties). We centered the building to their barycenter and scaled them to a unit sphere, as in DeepSDF \cite{deepSDF}.


\section{Experiments} \label{experiments}

Our experiments are designed to validate our two hypotheses:  first, that the network is capable of reconstructing diverse 3D building shapes (curved, with different topology, etc.) conditioning on a single perspective sketch in real-time measuring its accuracy and computational efficiency; second, to show how the building orientation affects the reconstruction metrics, the relevance of the orientation.

\textbf{Baseline implementation details}. We adopted \textit{OccNet} as a baseline and backbone for our experiments. Our initial version of \textit{OccNet} had 13,414,465 parameters with a model size of 161Mb. Testing OccNet directly on our sketches does not produce any meaningful results. Therefore we re-trained it over the Manhattan 1K dataset ( as in the next Section \ref{dataset}) with two methods: 
\begin{enumerate}
    \item Transfer learning with a Pre-training and Fine-Tuning technique (PFT). 
    \item Training our model from scratch (Vanilla Training - VT).
\end{enumerate}
 For the former, we fine-tuned their pre-trained model with 980k iterations over ShapeNet, freezing part of the network and retraining only the last layers. For the latter, we retrained the model from scratch, but unlike \textit{OccNet}, we applied the Adam optimizer with a learning rate of $\eta=10^{-4}$ and weight decay. For other hyperparameters of Adam \cite{adam} we used: $\beta_{1}=0.9, \beta_{2}=0.999$ and $\epsilon=10^{-8}$. Our batch size operates with size 32 due to the high memory cost of 3D shapes. We subsampled the shape with 2048 points. The main loss metric used was the Intersection over Union (IoU) between the generated occupancy grid and the voxel ground truth. A voxel resolution of 32 was adopted, and the latent embedding as $c$ had a size of 256. We reduced the 24 sketches per shape to a resolution of 224x224 pixels. We replaced ReLU with LeakyReLU and GeLU as nonlinear layers and different optimization strategies without visible qualitative improvement. The training took six hours using an RTX 2080 GPU and GCP cloud A100. In total, we estimated 20.000 Watts consumed for all trainings, producing around 10 tons of $CO_2$ emissions. We calculated and changed the initial dataset normalization: previously, the mean and standard deviation was based on ImageNet \cite{deng2009imagenet, chang2015shapenet}. In our dataset, the images are all black and white since they do not have an RGB channel.


\subsection{ First Experiment: Accuracy and Efficiency} \label{accandeff}

In this experiment, we validate the first hypothesis: that it is possible to generate 3D building shapes conditioned on a single perspective sketch accurately and efficiently for real-time design applications. Furthermore, our experiments introduce a relative baseline to test the accuracy and computational performance in single sketch 3D building reconstruction tasks.

We re-trained and fine-tuned the baseline to improve the \textbf{accuracy}. As shown in Table \ref{ref:accuracy}, adopting OccNet directly with a different dataset \cite{domainadaptation} produces unusable reconstructed outputs (Pre-Trained Model). Therefore, we pre-trained and fine-tuned the model (PFT) to assess its metrics, providing a new baseline for our model. During this assessment, we noticed initial overfitting in the training and validation curves as shown in the Supplementary Material \ref{supplementary} and as previous findings \cite{whatsvrlearn} confirmed for the ShapeNet dataset \cite{Occnet2019}. Our vanilla trained model (VT) with MobileNetV3 showed similar limited generalization capabilities but with a 38.5\% improvement for accuracy defined as the mean distance of points on the reconstructed mesh $\hat{y}$ to their nearest neighbors on the ground truth mesh $y_{nn}$, and a 41.3\% for completeness (define like accuracy, but in the opposite direction). We saved different model checkpoints at the early epochs and adopted early stopping to limit and avoid overfitting. Furthermore, we attenuated this overfitting behavior in other models with regularization losses. We tested with L2 regularization $\lambda R(f)$, adding weight decay to the Adam optimizer (weight decay= $10^{-2}$, and even higher) $\max_{f} \sum_{i=1}^{n} V\left(f\left(x_{i}\right), y_{i}\right)+\lambda R(f)$, for $\max _{f}$ is intended to max the $IoU_{voxels}$. 

To improve the \textbf{efficiency} (Table \ref{ref:efficiency}), for this experiment, we adopted MobileNetV3 instead of ResNet-18 as in \cite{Occnet2019} as the image encoder based on the smaller dataset at our disposal to reduce the network size. If this modification benefits our dataset, it should be re-confirmed while scaling it with more buildings. Instead, replacing the marching cubes algorithm with neural dual contouring (NDC) \cite{neuralcontours} increases the mesh reconstruction efficiency and perfectly scales with the dataset size.

Our experiments collected \textbf{quantitative} results about computational speed and accuracy. The quantitative analysis considered the model size, parameters, and time to complete each operation. 
Different encoding architectures and mesh reconstruction algorithms have been tested for the model performance, as shown in Table \ref{ref:efficiency}. The model size has been reduced by 30\% with MobileNetV3, impacting the model reconstruction quality minimally. The inference time of \textit{OccNet} was around 0.4s / mesh. It was slower than other methods (3D-R2N2 \cite{3dr2n2}: 11ms, PSGN \cite{psgn16}: 10ms, Pixel2Mesh \cite{pixel2mesh}: 31ms). Furthermore, implementing an end-to-end pipeline with NDC benefitted the mesh reconstruction's efficiency but increased the model size.




\begin{table*}[t]

\centering
%
\begin{tabular}{l|ccc|cc} 
 \toprule

   \textbf{Efficiency}          & \multicolumn{3}{c}{Computational}          & \multicolumn{2}{c}{Memory}                 \\ 
\cmidrule(lr){2-4}\cmidrule(lr){5-6}
\multicolumn{1}{c}{}         & Encoding  $\downarrow$             & Point Evaluation   $\downarrow$   & Mesh Reconstruction  $\downarrow$            & Model Parameters    $\downarrow$            & M.Size       $\downarrow$          \\ 
\toprule 
~ ~      ResNet101               &   0.006s    & 0.28s           & 0.155s & 26M    & 314Mb  \\
~ ~      ResNet18                &   0.002s    & 0.34s           & 0.157s & 13M    & 161Mb\\
~ ~      MobileNetV3 (ours)      &   0.012s    & $\textbf{0.12s}$  & 0.239s & 4M     & $\textbf{52Mb}$  \\
~ ~      NDC (ours)              &   -         &   -             & $\textbf{0.142s}$ & 0.17M  & $\textbf{0.7Mb*}$
\\\toprule
\end{tabular}
    \caption{Measuring network efficiency through inference, the current model has size 161Mb, and the largest mesh has around eight million vertices.  We increased the dimensionality of the conditional encoding in the linear layer (1000, $c$ $sim$),  adapting MobileNet to $\textit{OccNet}$. The conditional dimension affects the model size and number of parameters since the $\beta$ and $\gamma$ in the batch normalization layer are related by this parameter. The experiments report the average of ten trials tested on the same machine}
\label{ref:efficiency}

\end{table*}





\begin{table*}[t]
\centering
\begin{tabular}{l|ccc} 
 \toprule
   \textbf{Accuracy}          & \multicolumn{2}{c}{Model Reconstruction Metrics}                         \\ 
 \cmidrule(lr){2-3}
 \multicolumn{1}{c}{}     & Accuracy $\uparrow$         & Completeness\textbf{} $\uparrow$             \\  
\toprule
 ~ ~   Pre-Trained Model & NaN & NaN  \\ 
  ~ ~  Pre-Trained Fine-Tuned (ours) & 0.0401  & 0.0324 \\
 ~ ~   Vanilla Training MobileNetV3 (ours) & \textbf{0.0553}[+38.5\%]  & \textbf{0.0458} [+41.3\%]
\\\toprule
\end{tabular}

\caption{Measuring network accuracy for reconstruction tasks. The Pre Trained Model, has been trained with ShapeNet v2 Dataset with rendered images. If we use that model out of the box, the mesh reconstructed is completely flat and does not provide any useful 3D information}
\label{ref:accuracy}

\end{table*}

For the reconstruction accuracy of our approach with employed Chamfer-L1 distance $d_{C D}$, volumetric IoU, and an average consistency score.
The Chamfer-L1 distance is defined as the mean of an accuracy (mean distance of points on $\hat{y}_{recon}$ to their nearest neighbors on the $y_{recon}$) and a completeness metric (from $y_{recon}$ to $\hat{y}_{recon}$ as per the formula above). 
Volumetric IoU ($IoU_{voxels}$) is defined as in \cite{Occnet2019}.  
The Normal consistency exploits higher-order information. It represents the mean absolute dot product of the normals in one mesh and the normals at the nearest neighbors in the ground truth.

\begin{figure*}[ht]
  \centering
  \graphicspath{ {./images/} } 
  \includegraphics[width=1\linewidth]{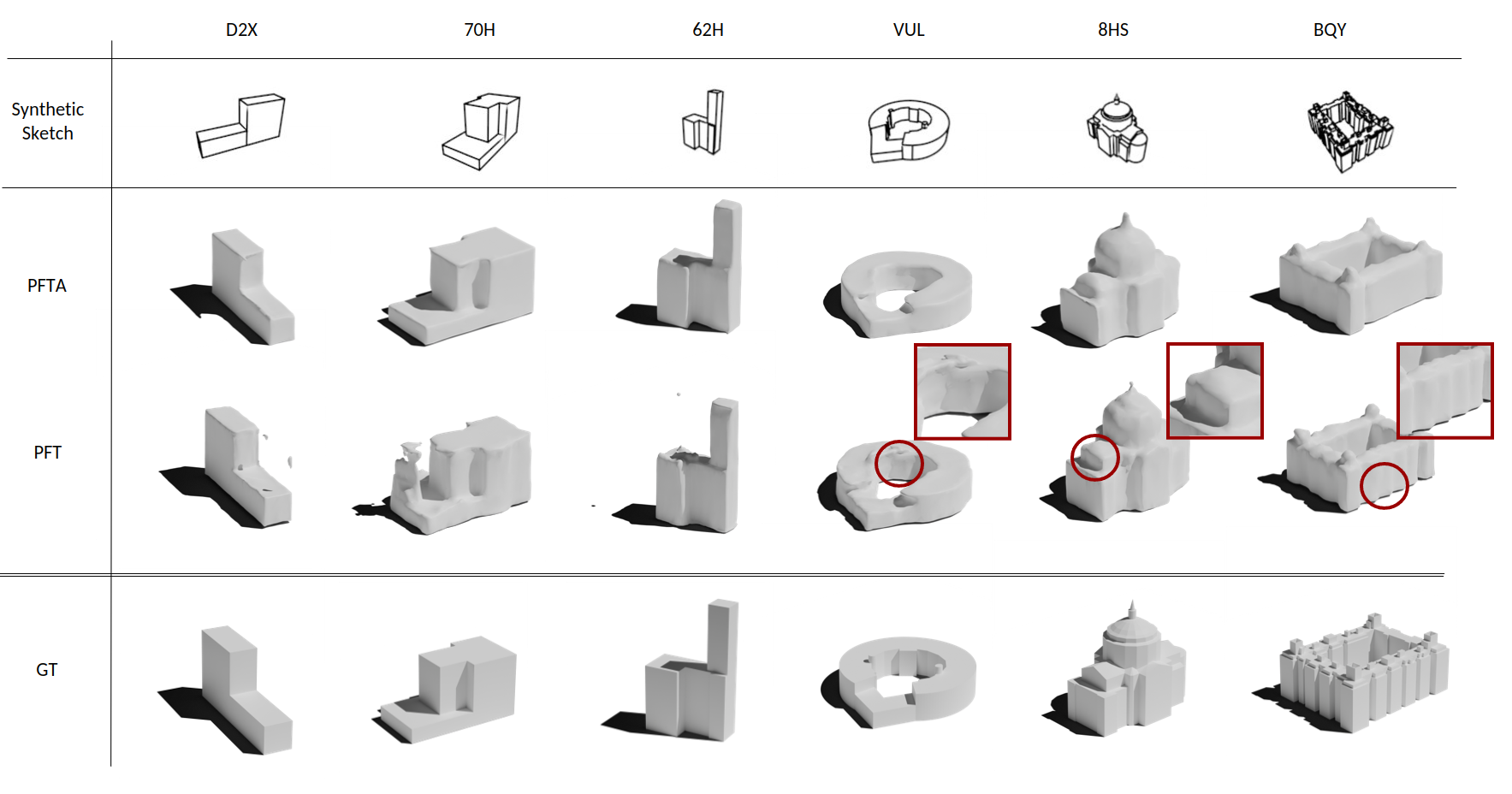}
   \caption{Qualitative comparison between PFT and PFTA. For simple geometries, aligning the buildings in the canonical pose hinders performance. Instead, the alignment helps with the more complex shapes. The shapes are extracted from the training set, showing the network capabilities.}
    \label{comparisonPFT-PFTA}
\end{figure*}

\subsection{Second Experiment: Relevance of Orientation} \label{context}
In Section \ref{method} we explained the importance of considering the building orientation in this generative task. We performed a total of four tests. We trained the models with two different settings as shown in Fig \ref{fig:workflow}: 
\begin{enumerate}
    \item with all the buildings perfectly aligned to the voxelization grid (Pre Trained Fine-Tuning Aligned - PFTA and Vanilla Training Aligned - VTA ) 
    \item with all the buildings preserving their native orientation (Pre Trained Fine-Tuning - PFT, Vanilla Training - VT).
\end{enumerate}

If Pre Trained Fine-Tuning (PFTA) and Pre Trained Fine-Tuning (PFT) are models pre-trained on ShapeNet and fine-tuned with Manhattan 1K, the other Vanilla Training Aligned (VTA) and Vanilla Training (VT) have been trained from scratch on Manhattan 1K, in Table \ref{tab:evaluation}. Based on the observed training curve,  Pre Trained Fine-Tuning (PFT) and Pre Trained Fine-Tuning Aligned (PFTA) models immediately overfit, instead Vanilla Training (VT) and Vanilla Training Aligned (VTA), with L2 regularization losses, provide a smoother training procedure. We also noticed that Conditional Batch Normalization (CBN) increases qualitative reconstruction performance from the samples analyzed confirming previous OccNet's findings. 

We provide an extensive \textbf{qualitative} analysis of the results in the supplementary material with synthetic sketches and nine real sketches. For 200 buildings in the test set, we randomly sample 30 and run the model on those analyzing their reconstruction performances. The quantitative results do not show a clear winner regarding loss and metrics. However, the aligned model offers a more robust reconstruction without artifacts like those without the model aligned. Furthermore, when we tested with real sketches, we discovered a better reconstruction when adding L2 regularization in training, considering the slightly different view and sketching style. Those sketches were drawn on an Ipad, with the same line weights, squared image, and white background.

\begin{table*}[t]

\centering
\begin{tabular}{l|ccc} 
\toprule

   \textbf{Relevance of Orientation}          & \multicolumn{3}{c}{Accuracy - Model Reconstruction Metrics}                          \\ 
\cmidrule(lr){2-4}
\multicolumn{1}{c}{}    &     Chamfer Distance$\downarrow$ &  Intersection over Union (IoU) $\uparrow$ & Normal Consistency $\uparrow$        \\ 

\textit{Preserving Original Orientation } & \multicolumn{1}{c}{} & \multicolumn{1}{c}{}  \\
~ ~      Pre Trained Fine Tuned (PFT) &  0.0362  & 0.6396 & 0.7788 \\ 
~ ~         Vanilla Trained (VT) &  0.0359  & 0.6500 & 0.7936 \\ 
\toprule

\textit{Canonical Pose Alignment} & \multicolumn{1}{c}{} & \multicolumn{1}{c}{} & \multicolumn{1}{c}{} \\
        Pre Trained Fine Tuned Aligned (PFTA)  & \textbf{0.0242}  & \textbf{0.7369} & \textbf{0.8438} \\ 
        Vanilla Trained Aligned (VTA) & 0.0268  & 0.7223 & 0.8427 \\
\toprule

\end{tabular}
 \caption{Vitruvio's benchmark: it showed the overall quantitative metrics. The aligned shapes (PFTA, VTA) provide better results on Chamfer Distance (CD), Intersection over Union (IoU), and Normals (N). Instead for accuracy and completeness, smaller models (MobileNetV3) present better metrics.}
 \label{tab:evaluation}

\end{table*}



\section{Discussion} \label{discussion}


In this section we discuss the results obtained from the experiment section ( Section \ref{experiments} ). Our experiments showed not only an improvement in the accuracy metrics (Table \ref{tab:evaluation}) from previous state-of-the-art and in the efficiency (Table \ref{ref:efficiency}), but also the importance of tracking orientation for the sketch-conditional 3D generation of buildings. We employed occupancy fields to implicitly represent actual building shapes, allowing the reconstruction to be independent of a fixed topology \cite{pixel2mesh} or a stringent parametrization. For example, this representation allows the generation of complex geometries (such as curved buildings, building with atriums, or openings) without constraining the masses to a planar surface, see ``VUL" and ``BHS" as depicted in Fig. \ref{comparisonPFT-PFTA}. While this could be detrimental for a fine-grained representation, it is suited to representing building masses. The watertight 3D mesh output provides volumetric information about the project that could better support other design phases where additional information is required. 
In this sketch to 3D workflow, the computational power is focused on generating information suitable for the Development Design phase (DD) instead of exploring endless design options \cite{performanceinformeddesign}. Furthermore, Vitruvio allows for a better synthesis of the building representation, now compressed to a 2048-dimensional vector. In fact, our algorithm encodes 3D features, producing a latent space that can be conditioned to others related to contextual information \cite{sketchdanhaivecaitlin,StructuralCausal3DRec,danhaive21,vae}: location and orientation, but also energy, or wind performance, being aware of the curse of dimensionality. 
Finally, this research showed qualitatively and quantitatively the alignment's effect. Forcing the dataset to a canonical pose during training shows worse behavior with simple shapes ( first three: ``D2X", ``70H" and ``62H", while it captures better details in complex buildings (last three ) \ref{comparisonPFT-PFTA}.

\section{Limitations} \label{limitation}
While Vitruvio presents promising future research directions, it has three main limitations: absence of real sketches in the dataset, limited metrics, and incomplete final 3D representation \ref{supplementary}. 
Firstly, the synthetic dataset of sketches \cite{deepsketchmodeling} does not capture the distribution of real sketches, making our method sensitive to the viewpoint and style of the sketch representation. It is possible to address this limitation by collecting a real dataset of architectural sketches and associated building masses. Secondly, the reconstruction metrics, CD, EMD, and others are not directly correlated to building design metrics and the creative, generative process ( coverage, COV, minimal matching distance MMD, and 1-NNA \cite{zeng2022lion} ). While on one side, the designers are trying to communicate the ideas they have in mind, in the preliminary phases are searching for inspirations and suggestions; therefore, the generative metrics should take this into account instead of focusing on achieving a perfect representation.  \cite{GET3D,pointclouddiffusion,pointvoxeldiffusion} present better metrics to evaluate the quality of the generative process, capturing the predictive posterior distribution, and these metrics could be further explored in future work. For example, when the designer draws a simple cube, our network can sample multiple options: it can generate a building with an atrium in the middle as one of the possible outputs, providing further suggestions and alternatives to the designer (Section \ref{method}). Currently, there are no specific metrics to evaluate multiple design outcomes provided by the sampling procedure of deep generative models \cite{dalle2,imagen,parti} for buildings. In our case, the final output consists of a watertight triangular mesh that provides volumetric information but lacks direct interoperability with Building Information Modeling and Virtual Design and Construction's (BIM/VDC) workflows. It would be better to modify the final output or convert it to CSG, B-Rep, polygonal mesh \cite{Free2CAD,sketch2mesh,polygen2020,point2cyl}. The BIM can be produced as a post-processing task using the overall mass as a starting point to extract floor and facade information. 
Regarding possible \textbf{ethical considerations}, we believe that the dataset should better represent all the buildings and designers worldwide to remove potential cultural biases. Simultaneously, it should be conscious of possible environmental issues in training large models \cite{foundationmodels}. The possibility of enhancing human capabilities has inspired this research, and we hope this research direction could facilitate the design of more sustainable buildings. Finally, while employing deep generative models such as diffusion models \cite{nam20223dldm} for a conditional generation based on more general inputs, it should carefully evaluate pros and cons \cite{tono2022transorgen}.


\section{Conclusion} \label{conclusion}

This paper presents Vitruvio: a conditional deep generative model that reconstructs a 3D printable building mesh (in USD format) from a single perspective sketch. For this purpose, we introduced a scalable pipeline for generating a dataset composed of geolocated buildings with respective synthetic sketch representations. This method shows promising new research directions to serve owners and designers better, providing a technology that better aligns with business expectations \cite{agrawal2022digital, agrawal2022digital1}. Furthermore, this research shows the effects of inductive biases produced by the alignment of buildings to a canonical pose, and the implications of overlooking these "contextual" information (Section \ref{limitation} ). Indeed, if the alignment improves the quantitative reconstruction metrics, it hinders the qualitative results, decreasing the generalization capabilities of the network. This behavior highlights the need for consistent evaluation metrics and a better understanding of the relationships between a building and its surroundings \cite{StructuralCausal3DRec}.

\section*{Acknowledgements}

This work has been supported by the Center for Integrated Facility Engineering at Stanford and the Computational Design Institute. A special thank you to Joseph Leybovich, Yulia Gryaditskaya, and Jacopo Borga for the comments and review. 

{\small
\bibliographystyle{ieee_fullname}
\bibliography{egbib}
}


\onecolumn

\begin{center}
\textbf{\large Supplemental Materials}
\end{center}
\subsection{Derivation of Conditional Generative Model Framework} \label{supplementary}

During the generative process we want to learn the distribution of the building shapes on Earth, to easily sample from it. When we sample from a statistical distribution we generate a building shape, and its realism is correlated to how well we approximate that distribution. So normally we want to learn the 3D building shape distribution $p(x)$ that it also depends by other random variables. Since our 3D shapes of buildings are represented as an Occupancy Network (Neural Network) with parameters $\theta$, we need to estimate the values of these parameters via an optimization process. In our case as seen in [ Section Intro] the distribution of the building shapes is also join with the sketches and contextual information that need to be encoded in order to be computed efficiently. Furthermore, during the encoding we want to guarantee that a set of latent variable $z$ is generated from a prior Gaussian distribution $p_\theta(z)$( parametrization trick in VAE, etc ) and instead the main dataset is generated by the generative distribution of the $p(x) = p(x, z ; \theta)$ to generate $p(x | y, \phi )$, where $ y, \phi$ can be encoded in $z$ creating a conditional generative model where the shape $x$ are parameterized by the Occupancy Network with parameters $\theta$.

In our model we can start from the assumption that the contextual information are not relevant (remove $\phi$ ).

Let $Q$ be a distribution over the possible values of $z$. That is, $\sum_{z} Q(z)=1$, $Q(z) \geq 0$ and for Jensen's Inequality the last step

$$
\begin{aligned}
\log p(x ; \theta) &=\log \sum_{z} p(x, z ; \theta) \\
&=\log \sum_{z} Q(z) \frac{p(x, z ; \theta)}{Q(z)} \\
& \geq \sum_{z} Q(z) \log \frac{p(x, z ; \theta)}{Q(z)}
\end{aligned}
$$

Then we know that $\frac{p(x, z ; \theta)}{Q(z)}=c$. Here Q is proportioned to the joint distribution of $z$ and $x$, so for some constant $c$ that does not depend on $z$. This is easily accomplished by choosing $Q(z) \propto p(x, z ; \theta)$ .

Actually, since we know $\sum_{z} Q(z)=1$ as a distribution, this further tells us that
$$
\begin{aligned}
Q(z) &=\frac{p(x, z ; \theta)}{\sum_{z} p(x, z ; \theta)} \\
&=\frac{p(x, z ; \theta)}{p(x ; \theta)} \\
&=\frac{p(z \mid x ; \theta)p(x ; \theta)}{p(x ; \theta)} \\
&=p(z \mid x ; \theta)
\end{aligned}
$$
Thus, we simply set the $Q$ 's to be the posterior distribution of the $z$ 's given $x$ and the setting of the parameters $\theta$.

Indeed, we can directly verify that when $Q(z)=p(z \mid x ; \theta)$, we have an equality because
$$
\begin{aligned}
\sum_{z} Q(z) \log \frac{p(x, z ; \theta)}{Q(z)} &=\sum_{z} p(z \mid x ; \theta) \log \frac{p(x, z ; \theta)}{p(z \mid x ; \theta)} \\
&=\sum_{z} p(z \mid x ; \theta) \log \frac{p(z \mid x ; \theta) p(x ; \theta)}{p(z \mid x ; \theta)} \\
&=\sum_{z} p(z \mid x ; \theta) \log p(x ; \theta) \\
&=\log p(x ; \theta) \sum_{z} p(z \mid x ; \theta) \\
&=\log p(x ; \theta) \quad\left(\text { because } \sum_{z} p(z \mid x ; \theta)=1\right)
\end{aligned}
$$
We denote the evidence lower bound (ELBO) by
$$
\operatorname{ELBO}(x ; Q, \theta)=\sum_{z} Q(z) \log \frac{p(x, z ; \theta)}{Q(z)}
$$

And since we want to maximize the ELBO we repeat until convergence the EM process where the (E-step) consists in set for each $i$,  $Q_{i}\left(z^{(i)}\right):=p(z^{(i)} \mid x^{(i)} ; \theta)$
and (M-step) where 
$\theta:=\arg \max _{\theta} \sum_{i=1}^{n} \operatorname{ELBO}\left(x^{(i)} ; Q_{i}, \theta\right) =\arg \max _{\theta} \sum_{i} \sum_{z^{(i)}} Q_{i}\left(z^{(i)}\right) \log \frac{p\left(x^{(i)}, z^{(i)} ; \theta\right)}{Q_{i}\left(z^{(i)}\right)} $ . Here it would be possible to solve the EM problem if we know the exact distribution of $z$ or $Q(z)$ but, in most of the cases, since we are using a NN and encoders, we don't know, so we use VAE to enforce that the distribution of $z$ is Gaussian, so we assume that 
$$
\begin{aligned}
z & \sim \mathcal{N}\left(0, I_{k \times k}\right) \\
x \mid z & \sim \mathcal{N}\left(g(z ; \theta), \sigma^{2} I_{d \times d}\right)
\end{aligned}
$$

VAE extends EM algorithms to high-dimensional continuous latent variables, like in our occupancy function, with non-linear models that uses Neural Network. In our case $x |z $ it is intractable since we have a NN, non linear model. We can not compute the exact the posterior distribution $p(z|x;\theta)$ for $Q(z)$. So we aim to find an approximation of
the true posterior distribution, called predictive posterior distribution. 

Recall that $\mathrm{EM}$ can be viewed as alternating maximization of $\mathrm{ELBO}(Q, \theta)$. Here instead, we optimize the EBLO over $Q \in \mathcal{Q}$ \cite{variationinference}
$$
\max _{Q \in \mathcal{Q}} \max _{\theta} \operatorname{ELBO}(Q, \theta)
$$

\begin{empheq}[box={\mymath[colback=white!30,drop lifted shadow, sharp corners]}]{equation*}
\begin{aligned}
\mathcal{L}_{\mathcal{B}}^{\text {gen }}(\theta, \psi)=& \frac{1}{|\mathcal{B}|} \sum_{i=1}^{|\mathcal{B}|}\left[\sum_{j=1}^{K} \mathcal{L}\left(f_{\theta}\left(p_{i j}, z_{i}\right), o_{i j}\right)+\mathrm{KL}\left(q_{\psi}\left(z \mid\left(p_{i j}, o_{i j}\right)_{j=1: K}\right) \| p_{0}(z)\right)\right]
\end{aligned}
\end{empheq}

And for CVAE 


$$
\begin{aligned}
\log p_\theta(\mathbf{y} \mid \mathbf{x}) &=K L\left(q_\phi(\mathbf{z} \mid \mathbf{x}, \mathbf{y}) \| p_\theta(\mathbf{z} \mid \mathbf{x}, \mathbf{y})\right)+\mathbb{E}_{q_\theta(\mathbf{z} \mid \mathbf{x}, \mathbf{y})}\left[-\log q_\phi(\mathbf{z} \mid \mathbf{x}, \mathbf{y})+\log p_\theta(\mathbf{y}, \mathbf{z} \mid \mathbf{x})\right] \\
& \geq \mathbb{E}_{q_\phi(\mathbf{z} \mid \mathbf{x}, \mathbf{y})}\left[-\log q_\phi(\mathbf{z} \mid \mathbf{x}, \mathbf{y})+\log p_\theta(\mathbf{y}, \mathbf{z} \mid \mathbf{x})\right] \\
&=\mathbb{E}_{q_\phi(\mathbf{z} \mid \mathbf{x}, \mathbf{y})}\left[-\log q_\phi(\mathbf{z} \mid \mathbf{x}, \mathbf{y})+\log p_\theta(\mathbf{z} \mid \mathbf{x})\right]+\mathbb{E}_{q_\phi(\mathbf{z} \mid \mathbf{x}, \mathbf{y})}\left[\log p_\theta(\mathbf{y} \mid \mathbf{x}, \mathbf{z})\right] \\
&=-K L\left(q_\phi(\mathbf{z} \mid \mathbf{x}, \mathbf{y}) \| p_\theta(\mathbf{z} \mid \mathbf{x})\right)+\mathbb{E}_{q_\phi(\mathbf{z} \mid \mathbf{x}, \mathbf{y})}\left[\log p_\theta(\mathbf{y} \mid \mathbf{x}, \mathbf{z})\right]
\end{aligned}
$$

\begin{figure}[hb]
  \centering
  \graphicspath{ {./images/} }
  \includegraphics[width=1 \textwidth]{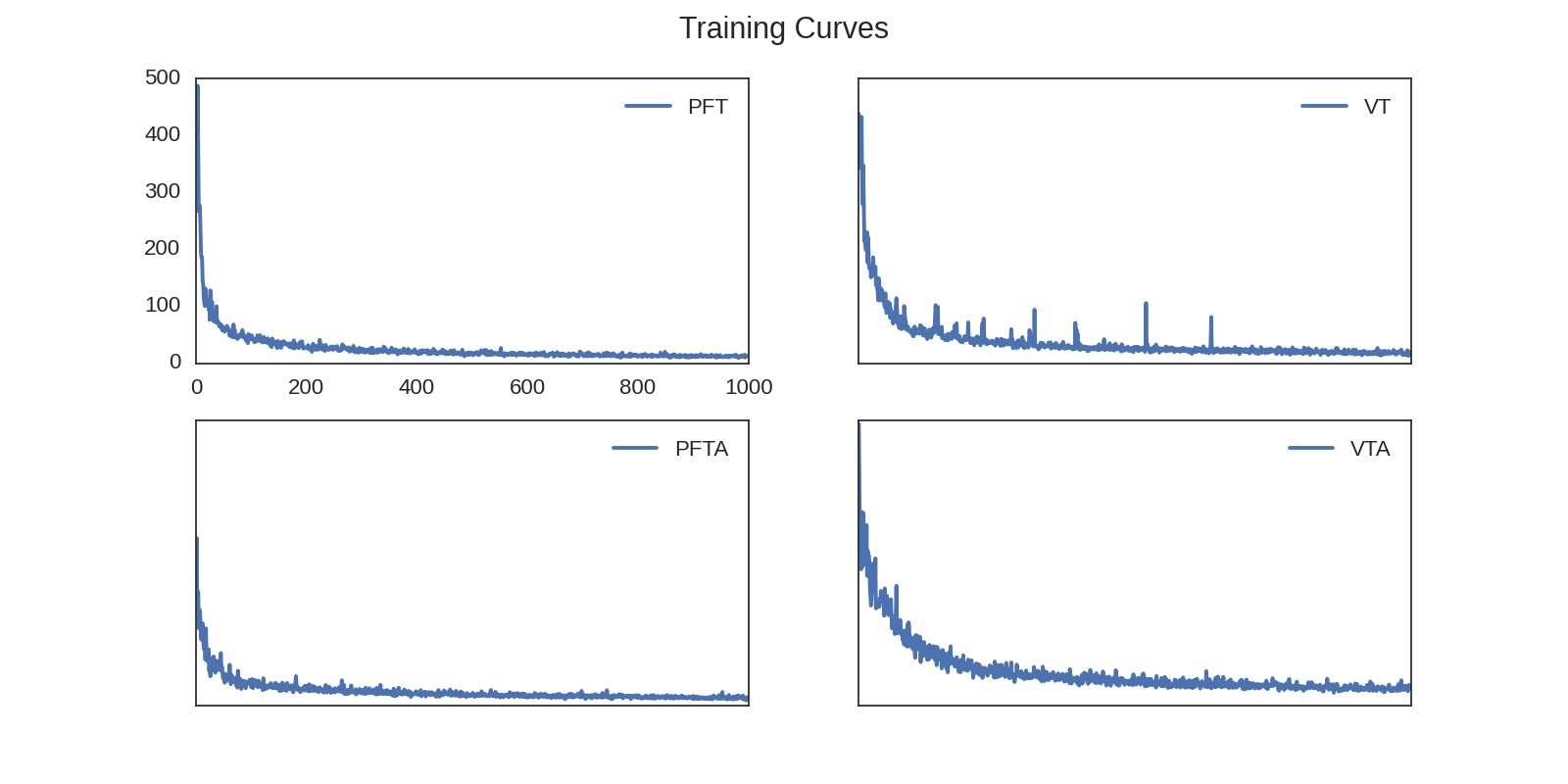}
   \caption{Initial training and validation curves. In our main 4 experiments. Pre-train Fine Tuned and Vanilla Training, with buildings aligned and not.}

\end{figure}

\begin{figure*}
  \centering
  \graphicspath{ {./images/} }
  \includegraphics[width=0.8 \textwidth]{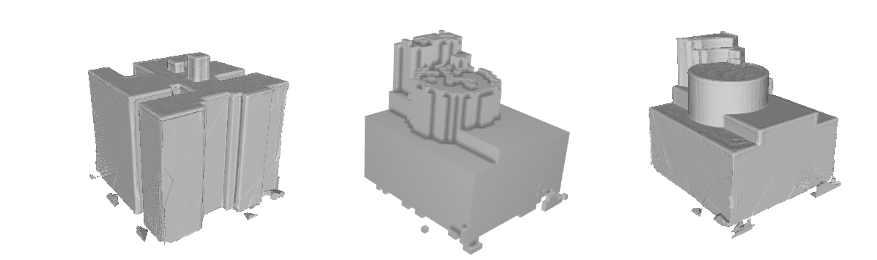}
   \caption{Generating watertight meshes, produces these artifacts to be aware. This issue could be solved with \cite{huang2020manifoldplus} \cite{Vicini2022sdf}. }

\end{figure*}

\begin{table*}[]
\centering
\resizebox{\textwidth}{!}{%
\begin{tabular}{ccccc}
 \\ \hline
  \includegraphics[scale=0.35]{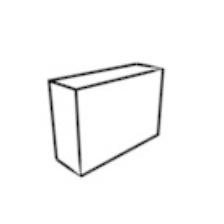} &
  \includegraphics[scale=0.35]{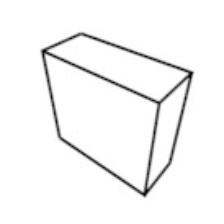} &
  \includegraphics[scale=0.35]{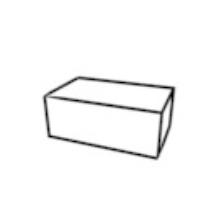} &
  \includegraphics[scale=0.35]{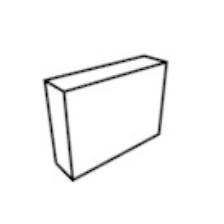} &
  \includegraphics[scale=0.35]{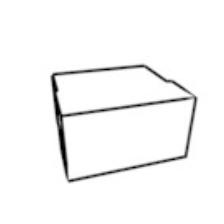} 
  \\ 
  \includegraphics[scale=0.15]{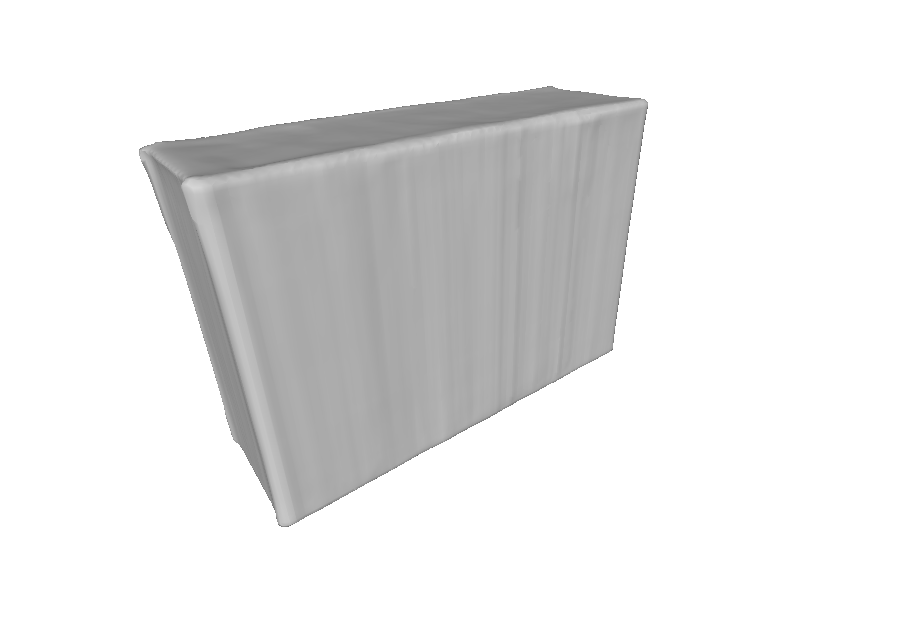} &
  \includegraphics[scale=0.15]{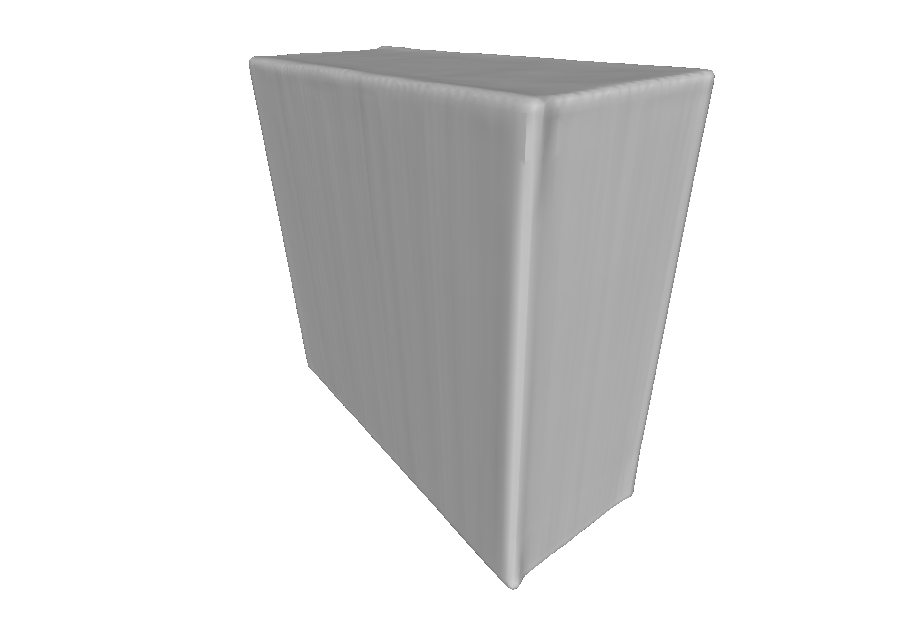} &
  \includegraphics[scale=0.15]{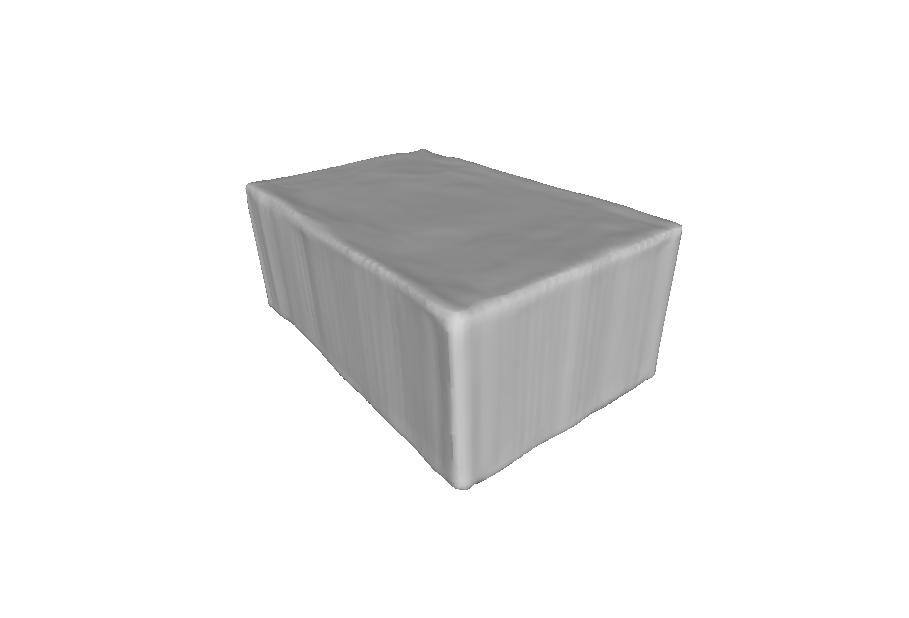} &
  \includegraphics[scale=0.15]{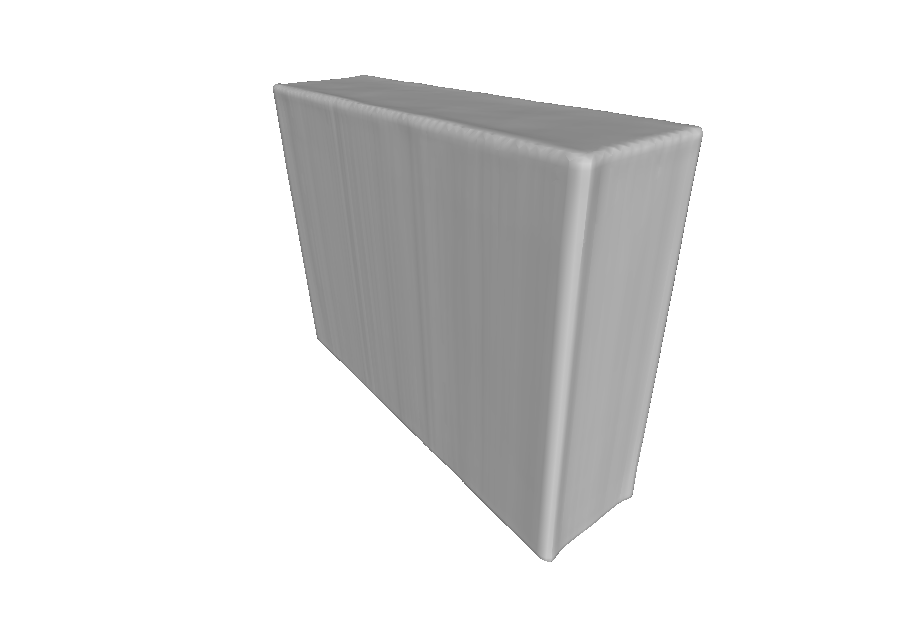} &
  \includegraphics[scale=0.15]{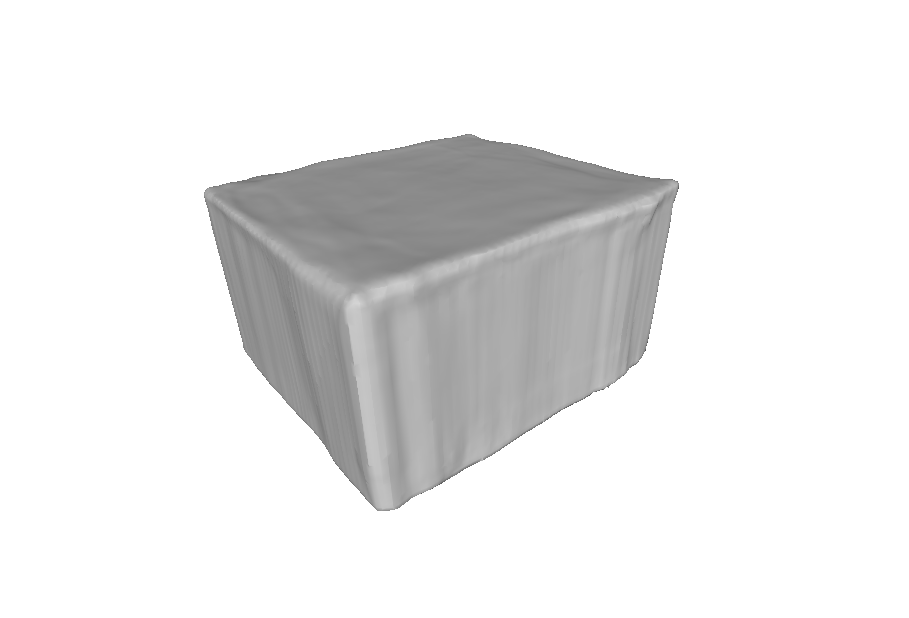} 
  \\ \hline
  
  \includegraphics[scale=0.35]{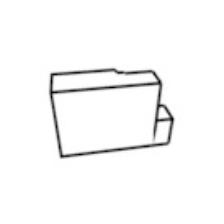} &
  \includegraphics[scale=0.35]{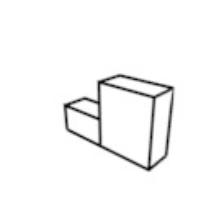} &
  \includegraphics[scale=0.35]{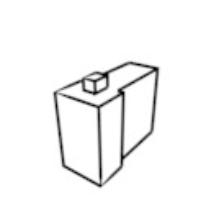} &
  \includegraphics[scale=0.35]{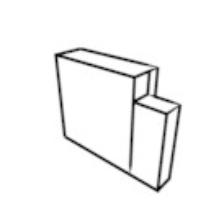} &
  \includegraphics[scale=0.35]{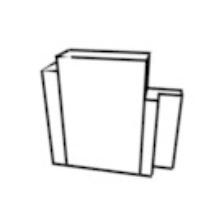} 
  \\
  \includegraphics[scale=0.15]{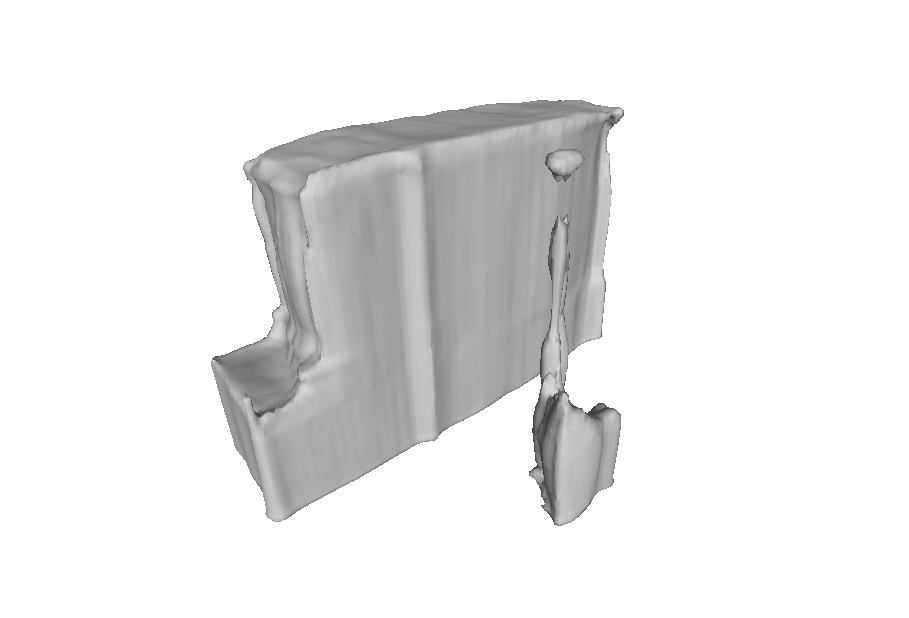} &
  \includegraphics[scale=0.15]{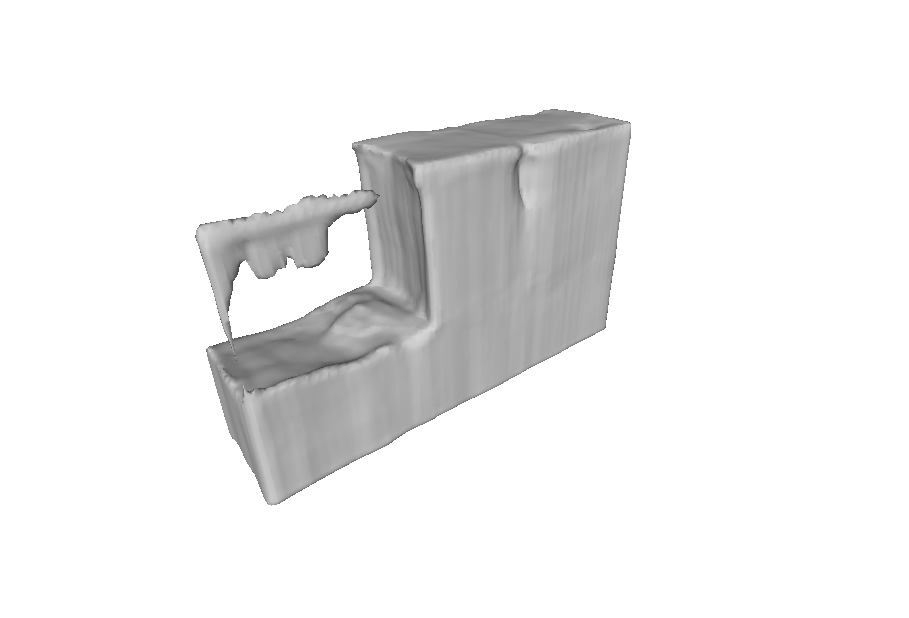} &
  \includegraphics[scale=0.15]{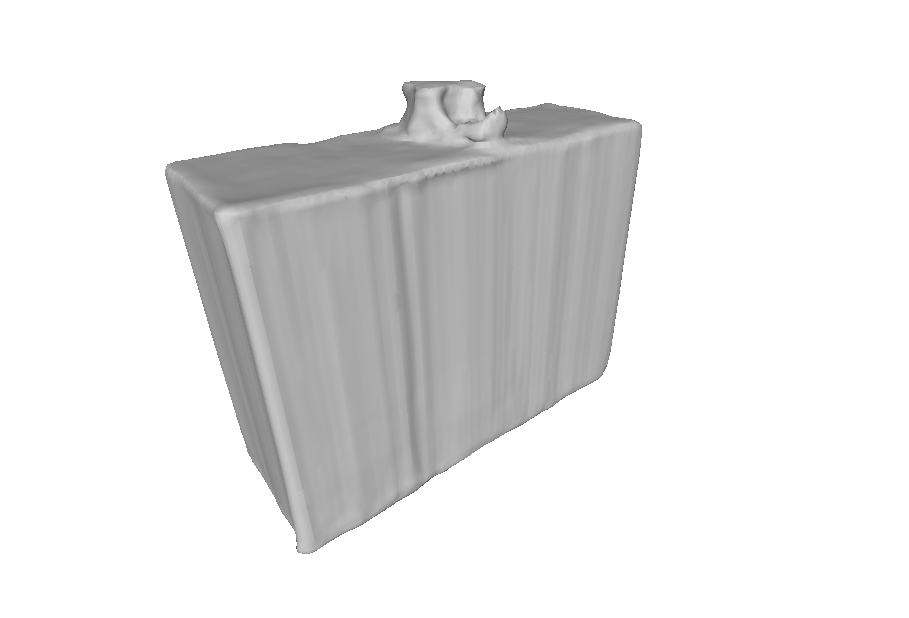} &
  \includegraphics[scale=0.15]{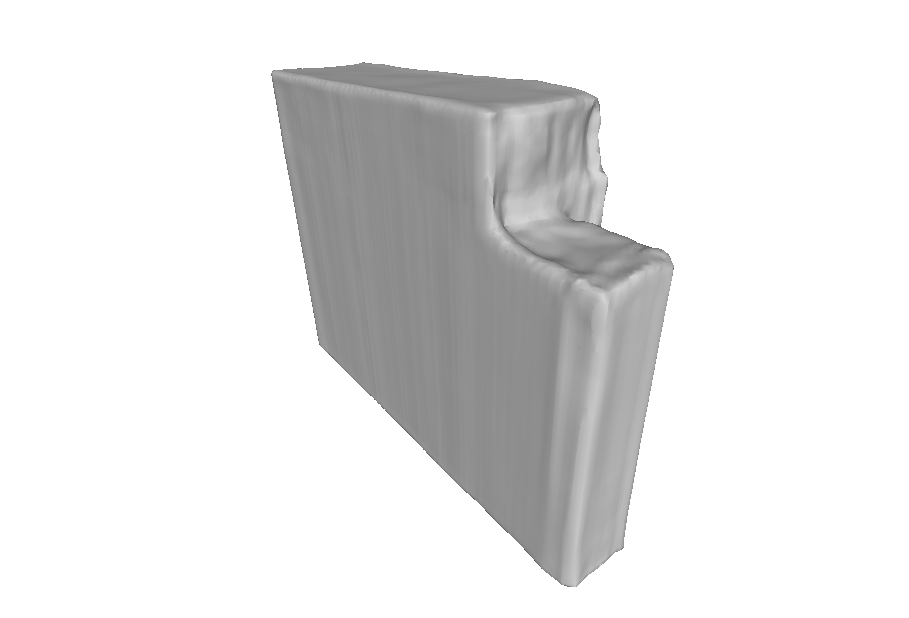} &
  \includegraphics[scale=0.15]{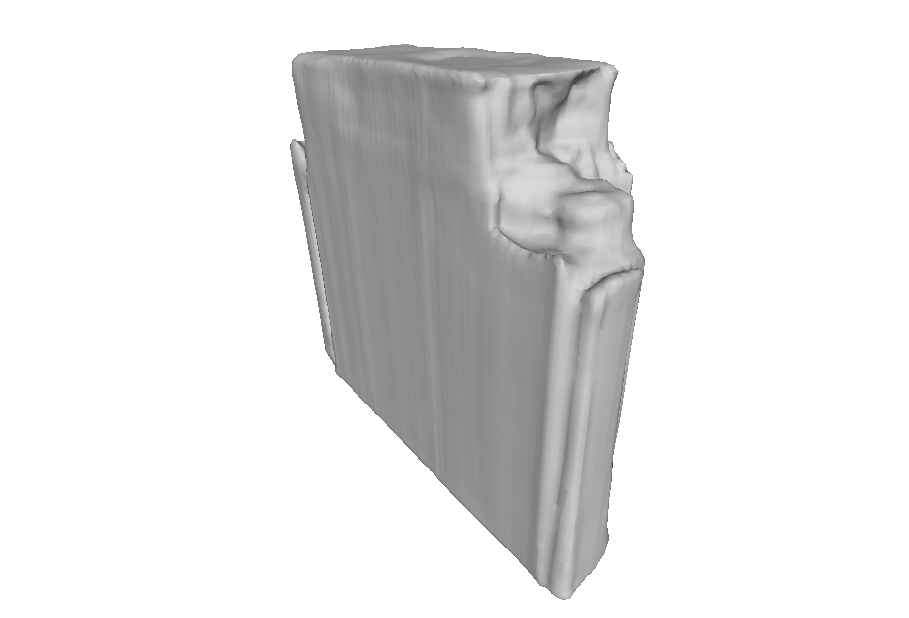}  
  \\ \hline
  
  \includegraphics[scale=0.35]{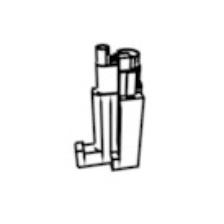} &
  \includegraphics[scale=0.35]{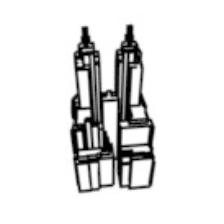} &
  \includegraphics[scale=0.35]{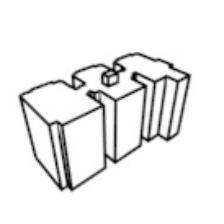} &
  \includegraphics[scale=0.35]{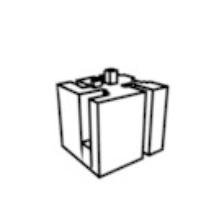} &
  
  \\ 
 \includegraphics[scale=0.15]{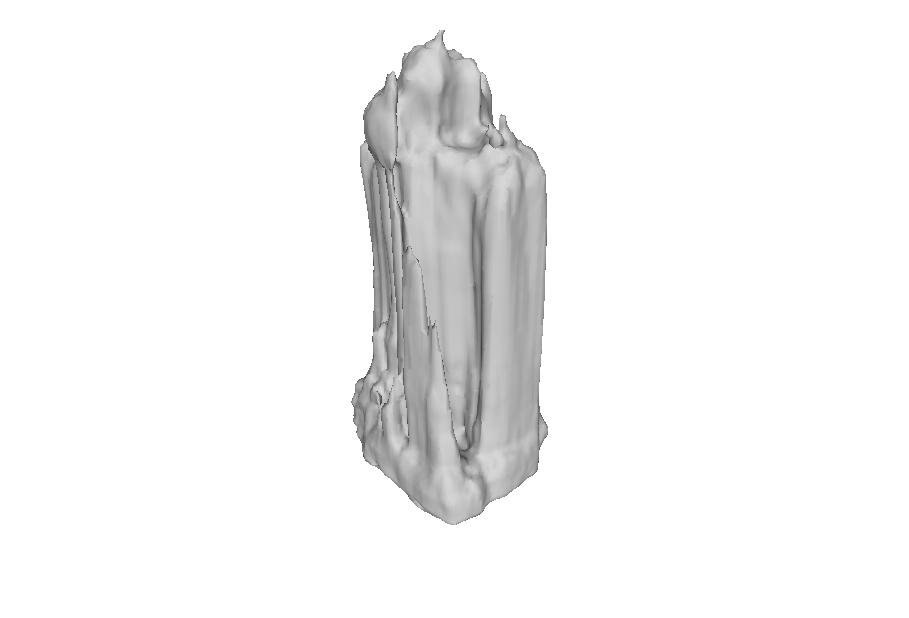} &
  \includegraphics[scale=0.15]{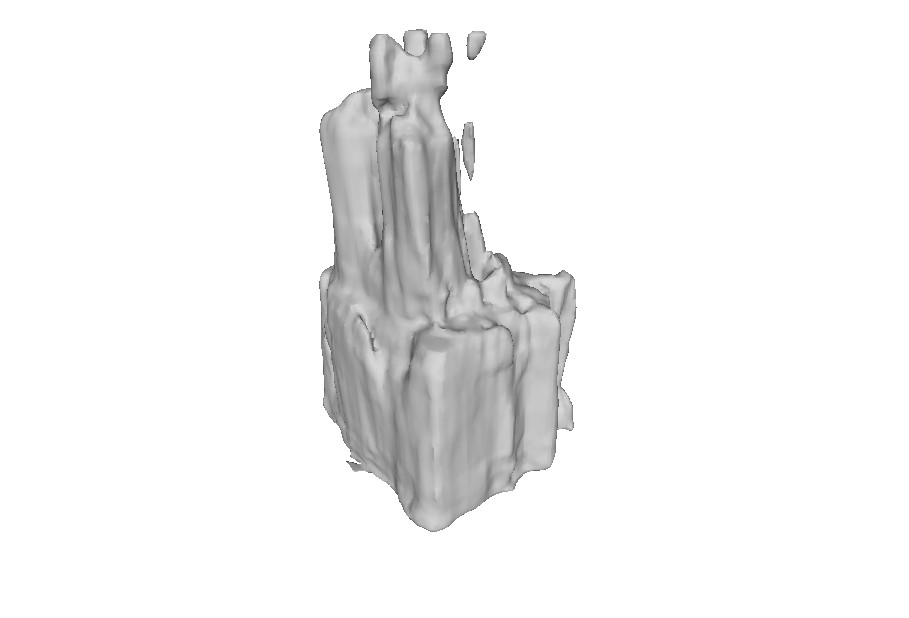} &
  \includegraphics[scale=0.15]{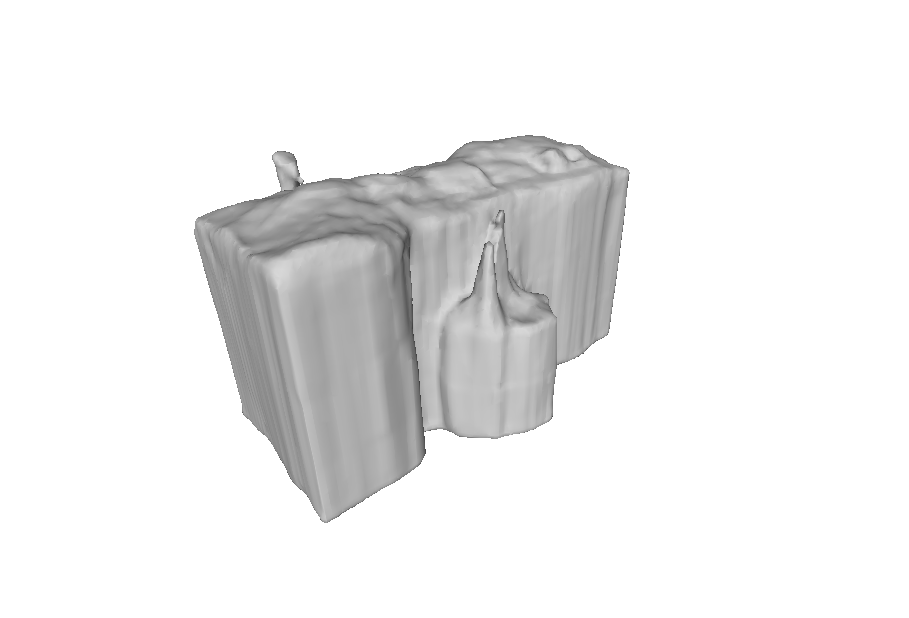} &
\includegraphics[scale=0.15]{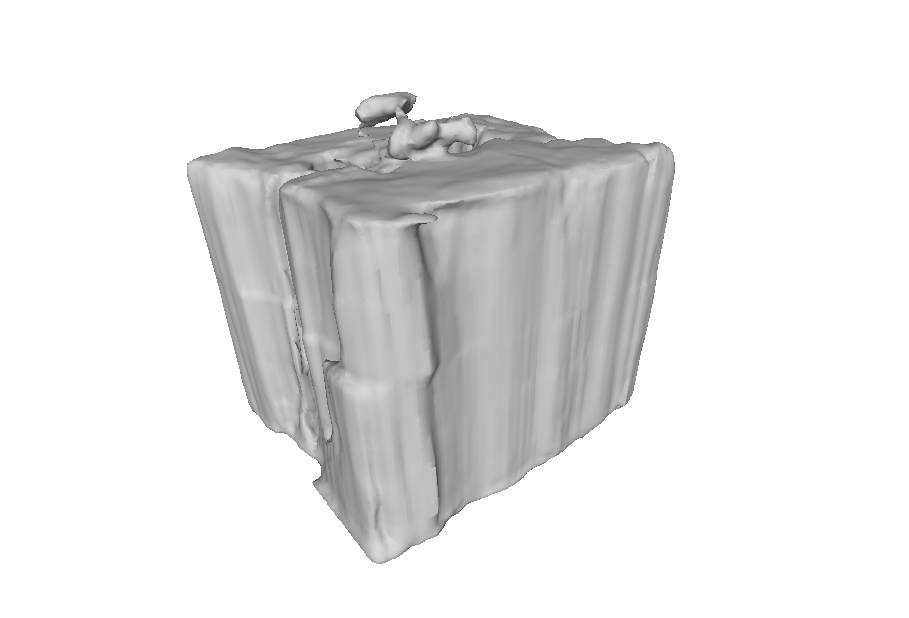} &
 \\   \\ \hline

\end{tabular}%
}
\caption{PFT 3D mesh reconstruction for a pretrained and fine tuned model results divide based on complexity level. To note that the first and third building are aligned differently, the model capture their differeces. This model contained more artifacts, due to the issue in the voxelization, but at the same time is capable to capture more fine-grain details, like in building 8, also other edges in a more expressive way. Qualitative speaking the best model overall}

\end{table*}

\begin{table*}[]
\centering
\resizebox{\textwidth}{!}{%
\begin{tabular}{ccccc}
 \\ \hline
  \includegraphics[scale=0.35]{images/sketch_rec/29_in.jpg} &
  \includegraphics[scale=0.35]{images/sketch_rec/11_in.jpg} &
  \includegraphics[scale=0.35]{images/sketch_rec/03_in.jpg} &
  \includegraphics[scale=0.35]{images/sketch_rec/05_in.jpg} &
  \includegraphics[scale=0.35]{images/sketch_rec/07_in.jpg} 
  \\ 
  \includegraphics[scale=0.15]{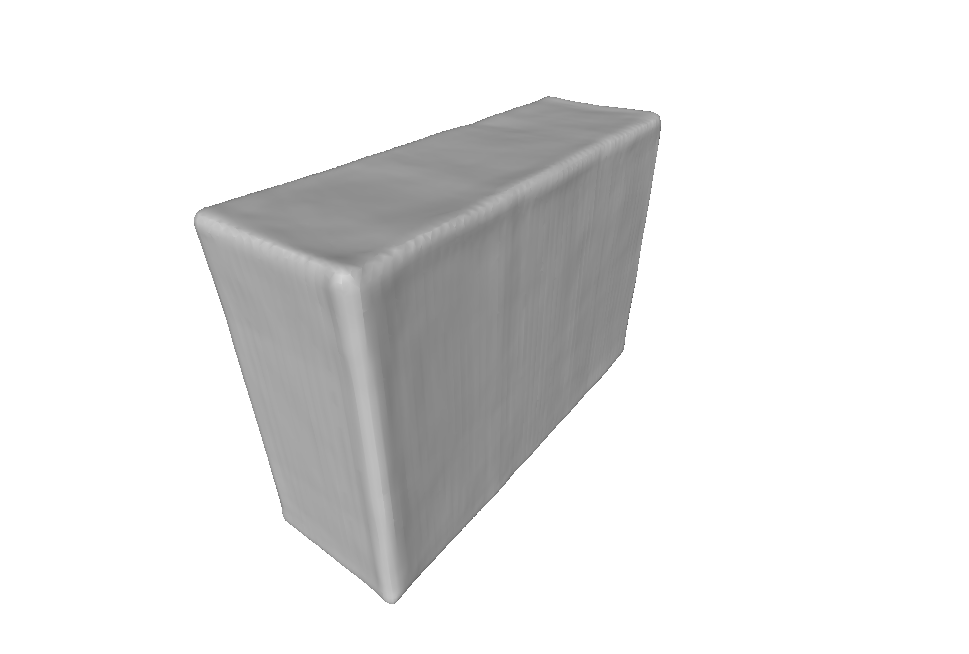} &
  \includegraphics[scale=0.15]{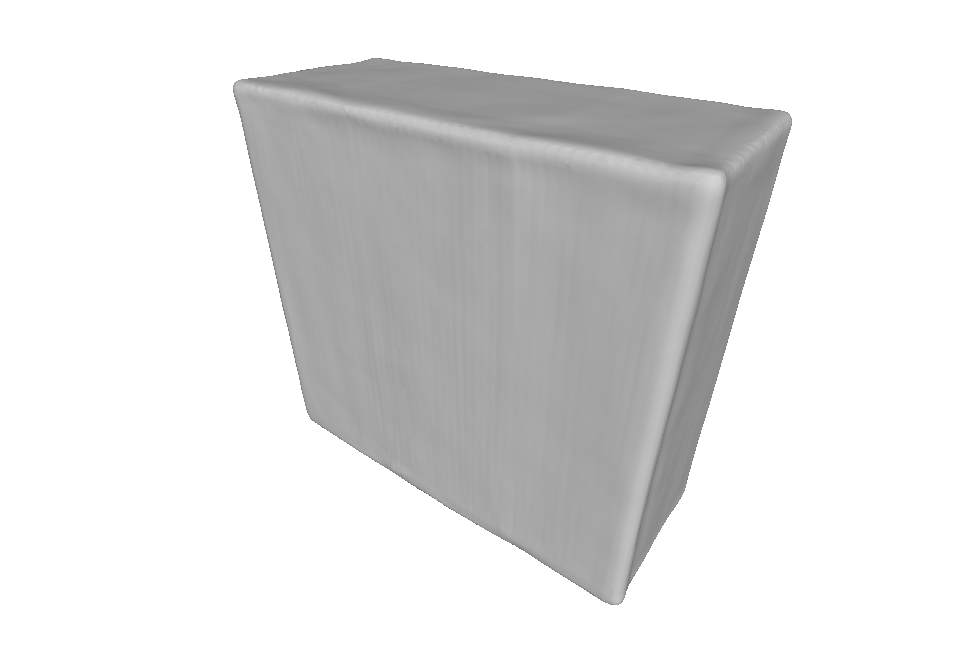} &
  \includegraphics[scale=0.15]{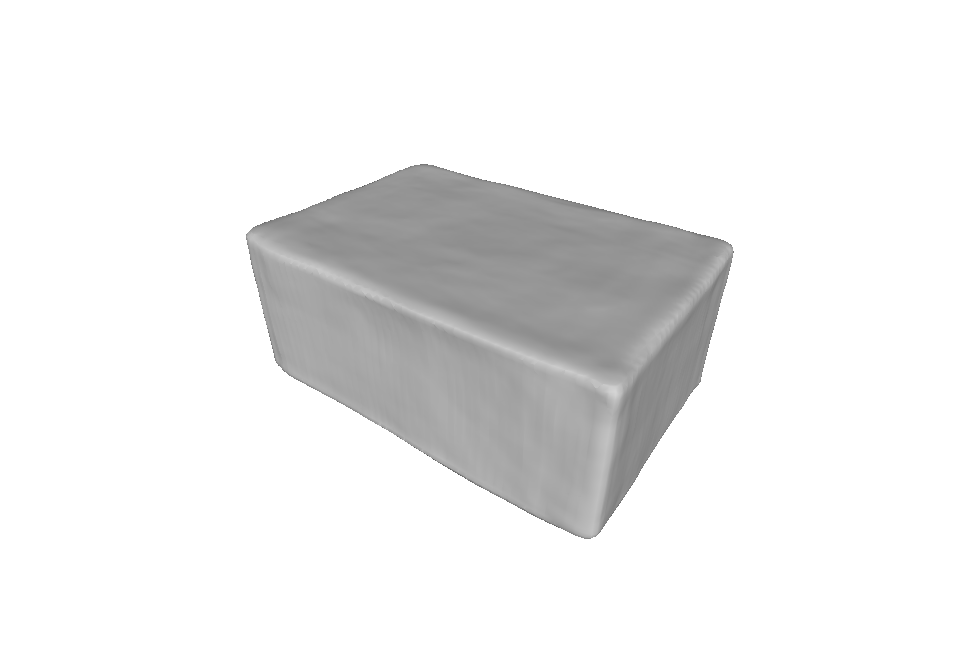} &
  \includegraphics[scale=0.15]{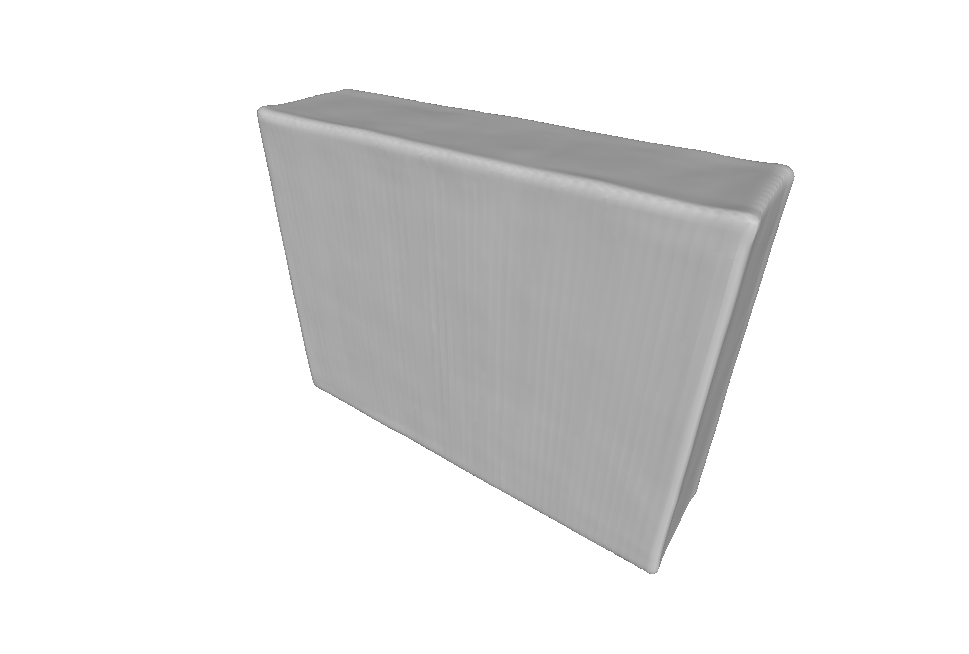} &
  \includegraphics[scale=0.15]{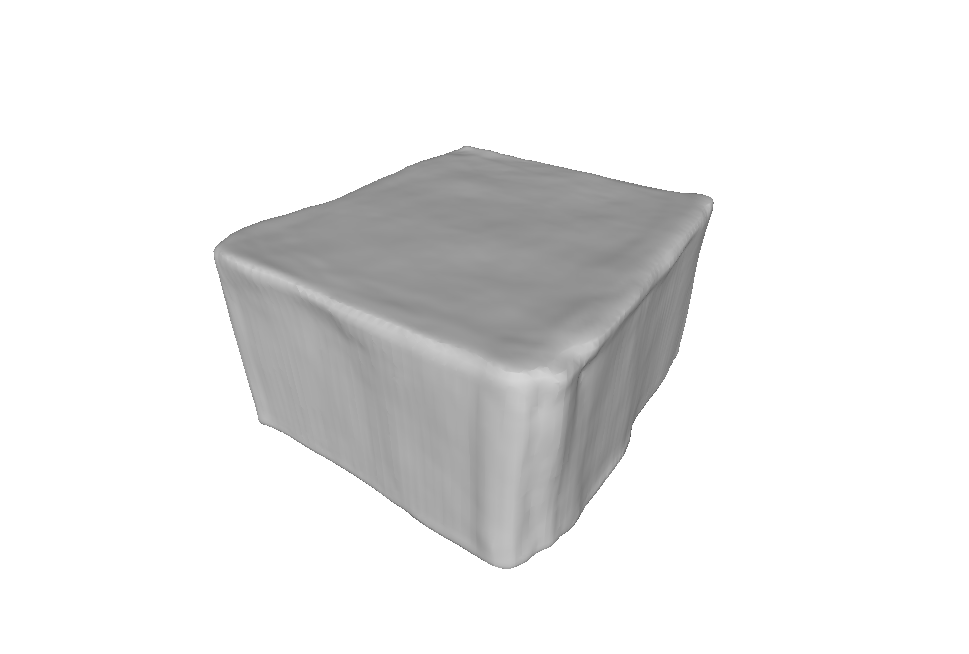} 
  \\ \hline
  
  \includegraphics[scale=0.35]{images/sketch_rec/14_in.jpg} &
  \includegraphics[scale=0.35]{images/sketch_rec/19_in.jpg} &
  \includegraphics[scale=0.35]{images/sketch_rec/23_in.jpg} &
  \includegraphics[scale=0.35]{images/sketch_rec/01_in.jpg} &
  \includegraphics[scale=0.35]{images/sketch_rec/10_in.jpg} 
  \\
  \includegraphics[scale=0.15]{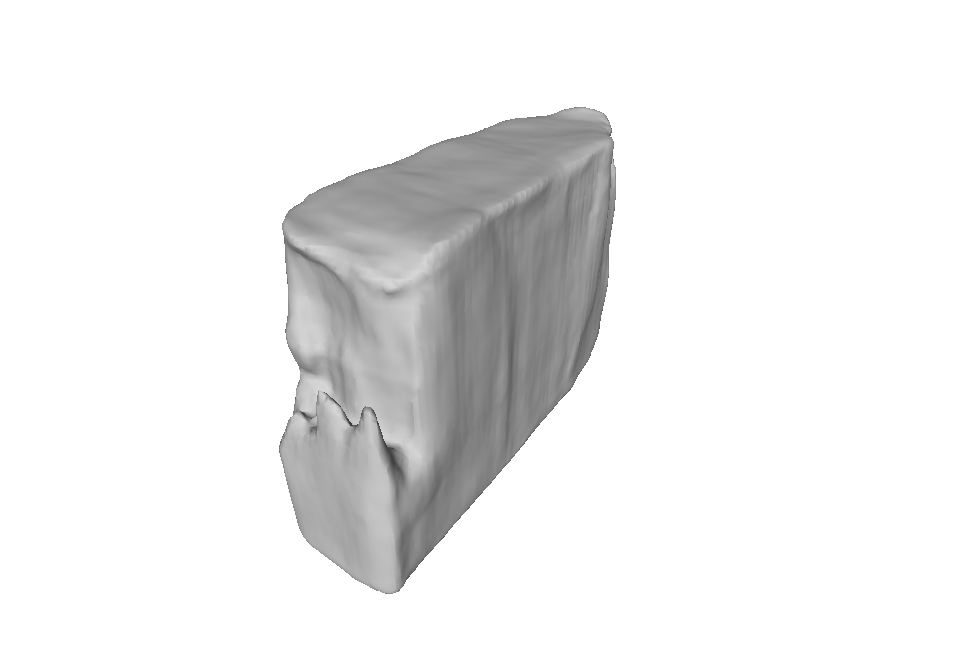} &
  \includegraphics[scale=0.15]{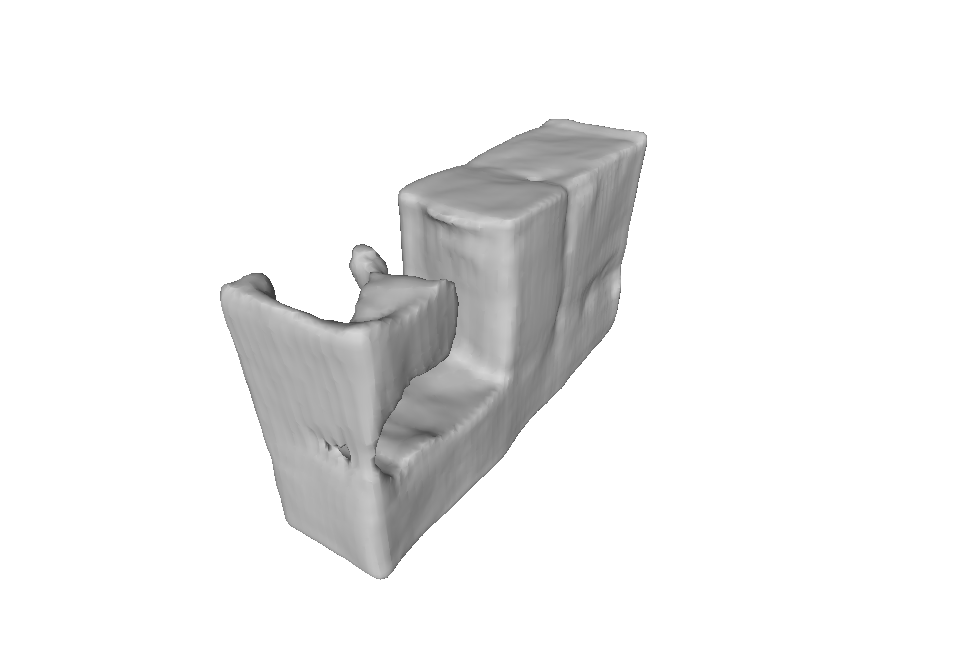} &
  \includegraphics[scale=0.15]{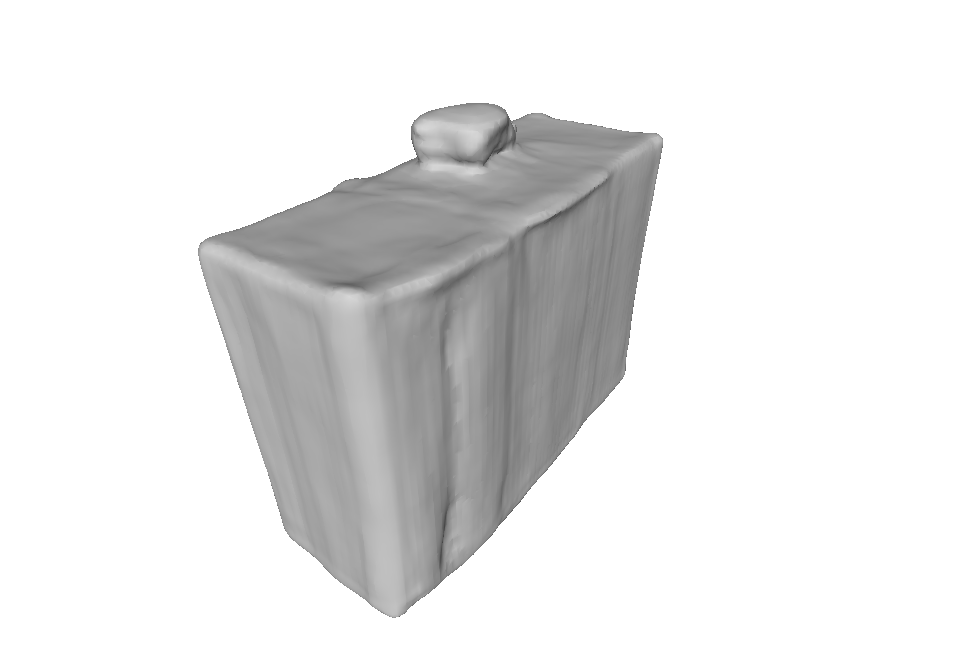} &
  \includegraphics[scale=0.15]{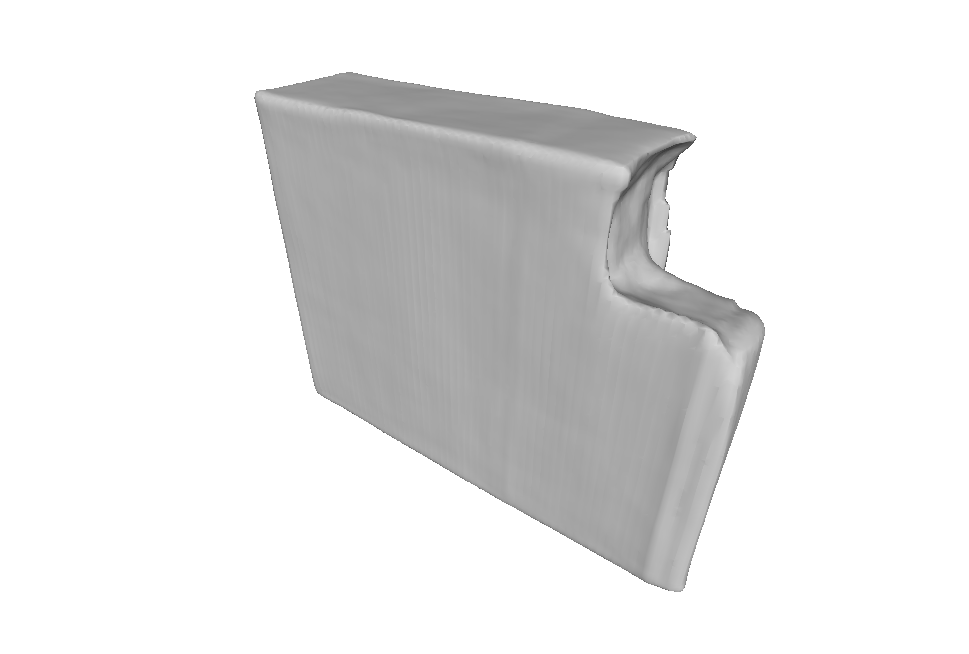} &
  \includegraphics[scale=0.15]{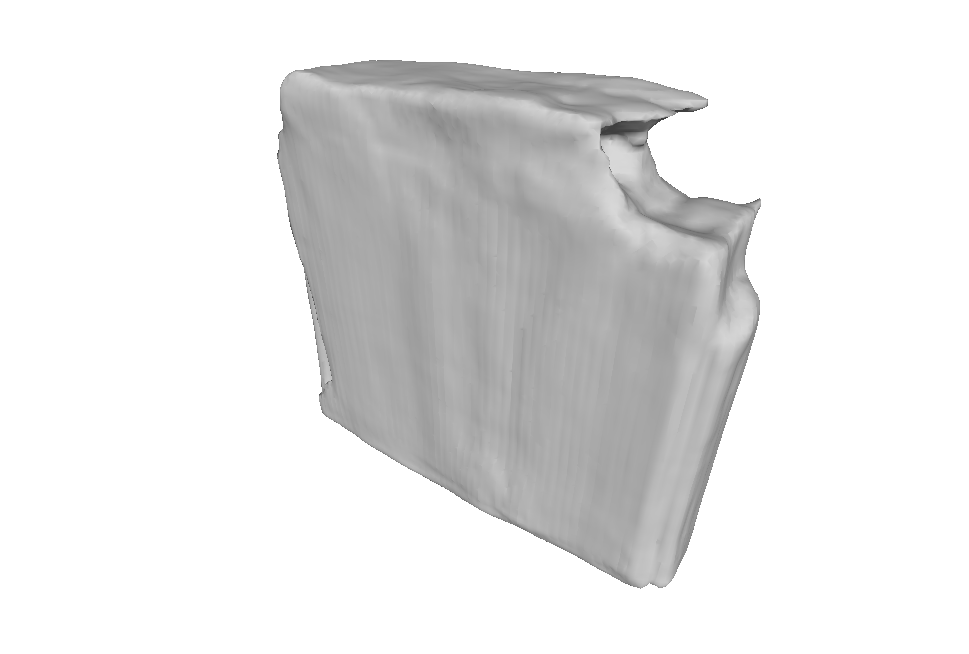}  
  \\ \hline
  
  \includegraphics[scale=0.35]{images/sketch_rec/02_in.jpg} &
  \includegraphics[scale=0.35]{images/sketch_rec/25_in.jpg} &
  \includegraphics[scale=0.35]{images/sketch_rec/08_in.jpg} &
  \includegraphics[scale=0.35]{images/sketch_rec/27_in.jpg} &
  
  \\ 
 \includegraphics[scale=0.15]{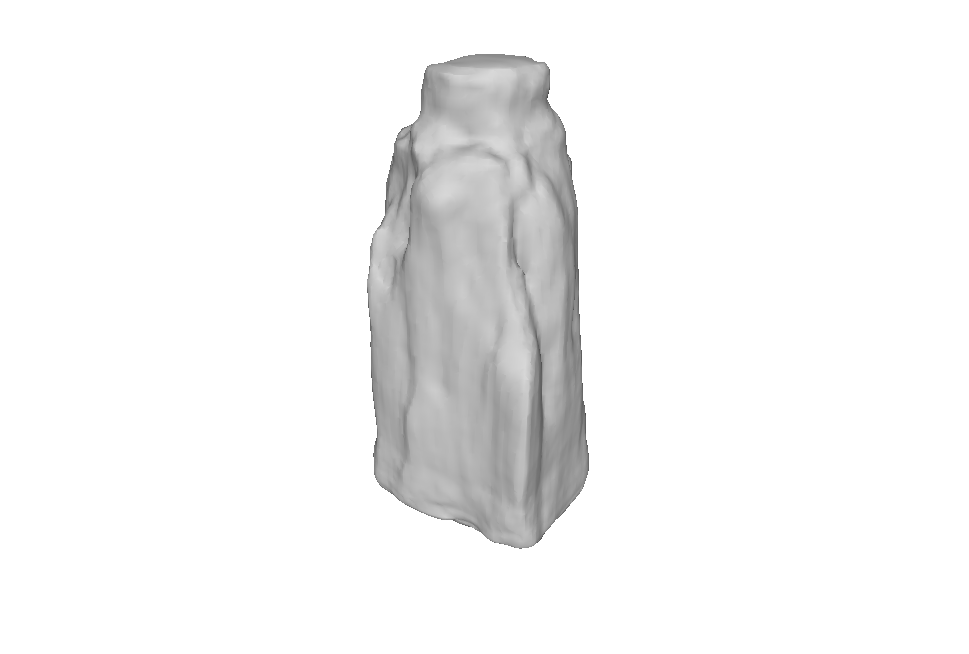} &
  \includegraphics[scale=0.15]{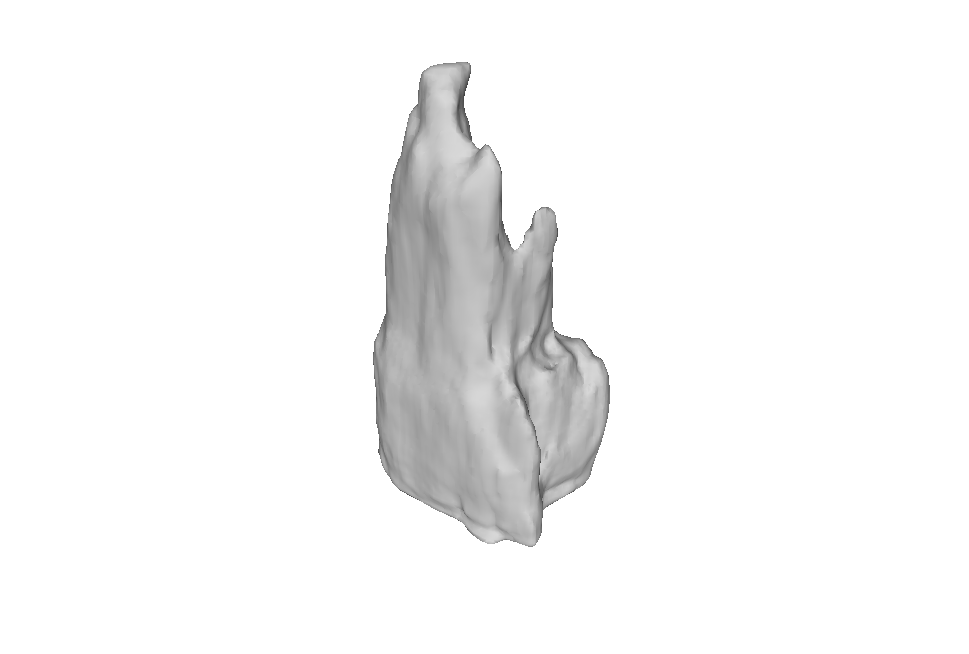} &
  \includegraphics[scale=0.15]{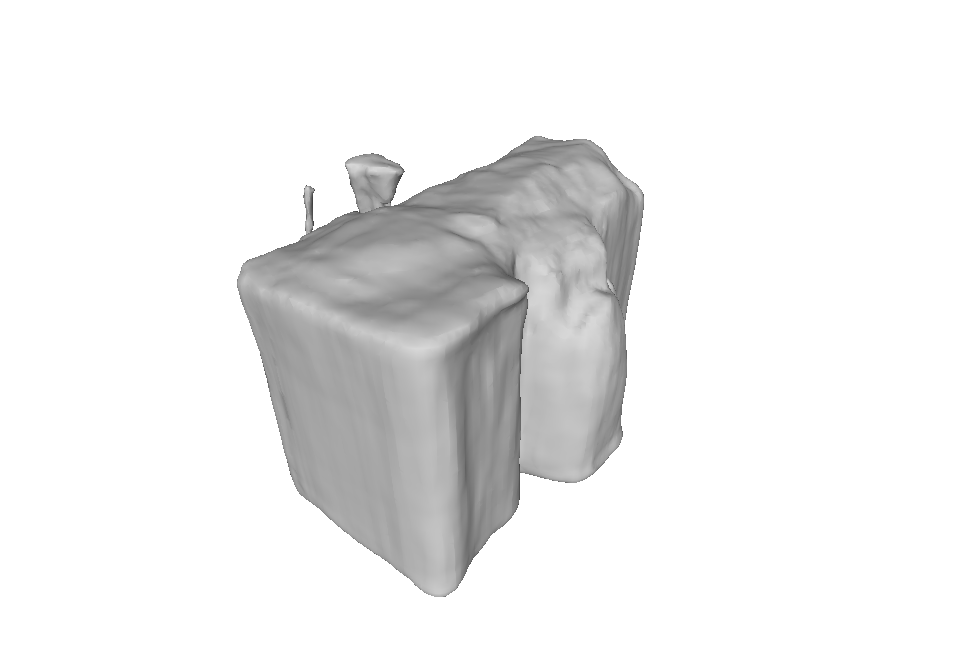} &
\includegraphics[scale=0.15]{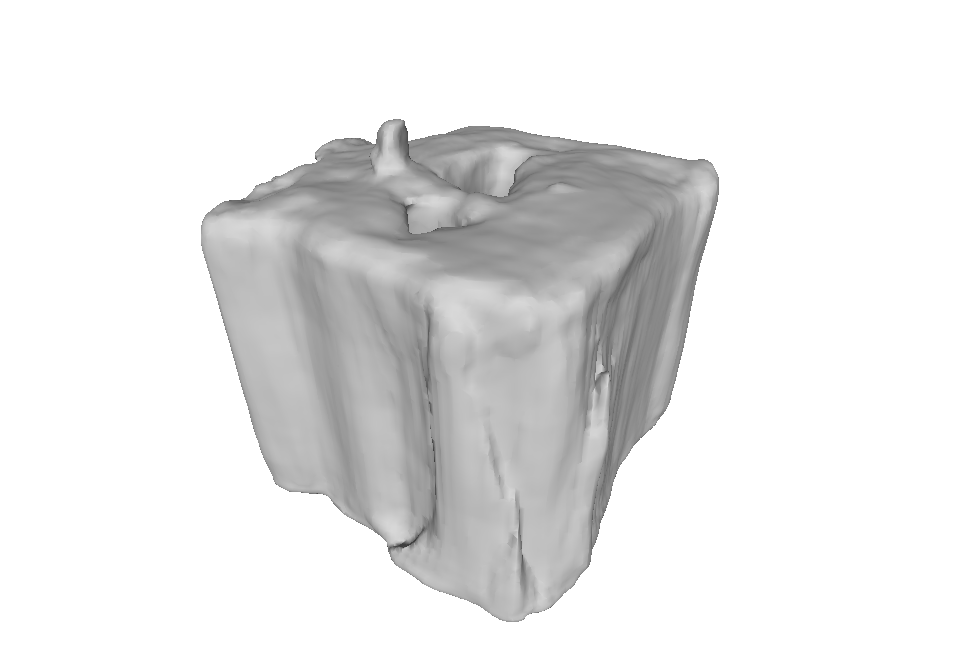} &
  \\   \\ \hline

\end{tabular}%
}
\caption{VT best model, loss got around 13.23. Also here we can see that training with the original orientation of the buildings allow the model to reconstruct also with some dependency of the view, some buildings follow the main direction of the sketch buildings}
\end{table*}

\begin{table*}[]
\centering
\resizebox{\textwidth}{!}{%
\begin{tabular}{ccccc}
 \\ \hline
  \includegraphics[scale=0.35]{images/sketch_rec/29_in.jpg} &
  \includegraphics[scale=0.35]{images/sketch_rec/11_in.jpg} &
  \includegraphics[scale=0.35]{images/sketch_rec/03_in.jpg} &
  \includegraphics[scale=0.35]{images/sketch_rec/05_in.jpg} &
  \includegraphics[scale=0.35]{images/sketch_rec/07_in.jpg} 
  \\ 
  \includegraphics[scale=0.15]{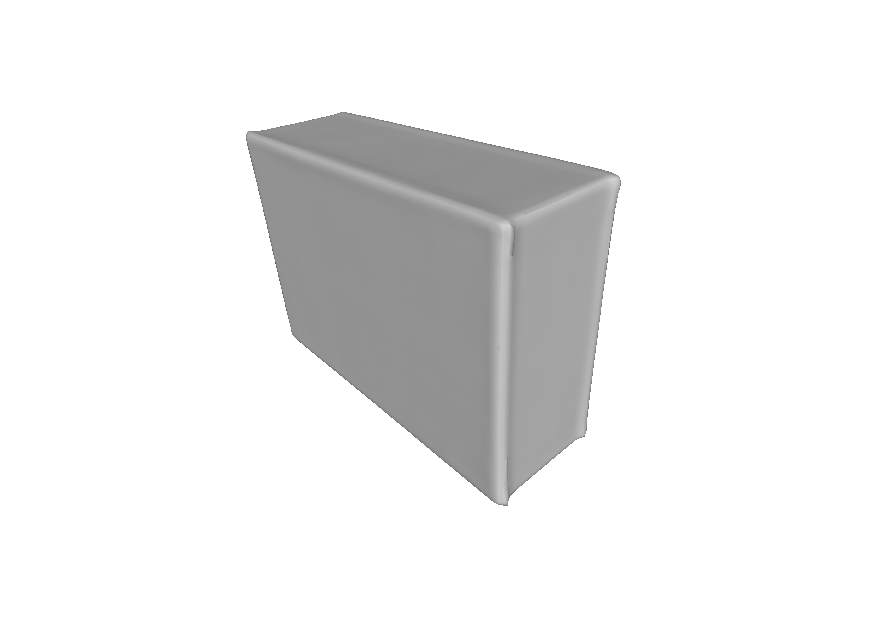} &
  \includegraphics[scale=0.15]{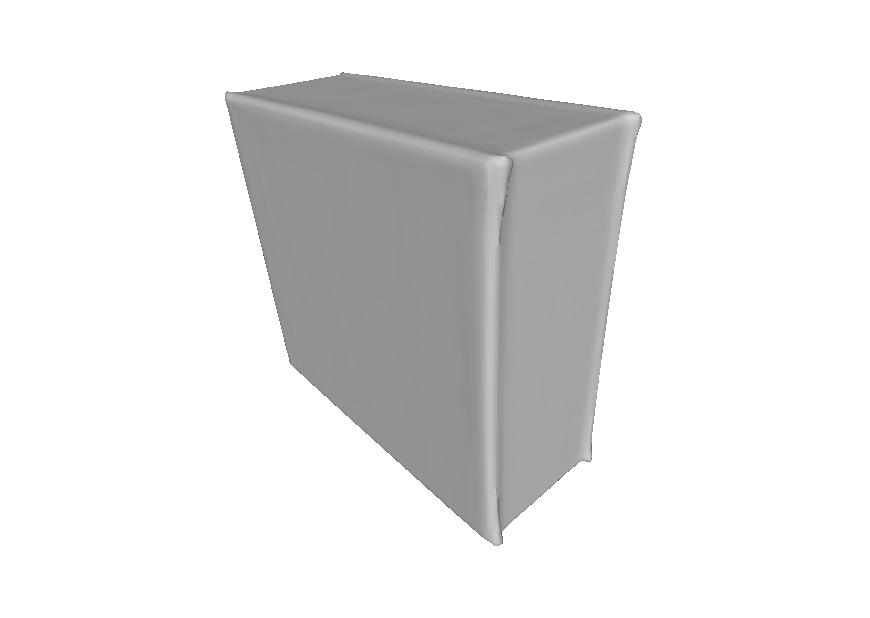} &
  \includegraphics[scale=0.15]{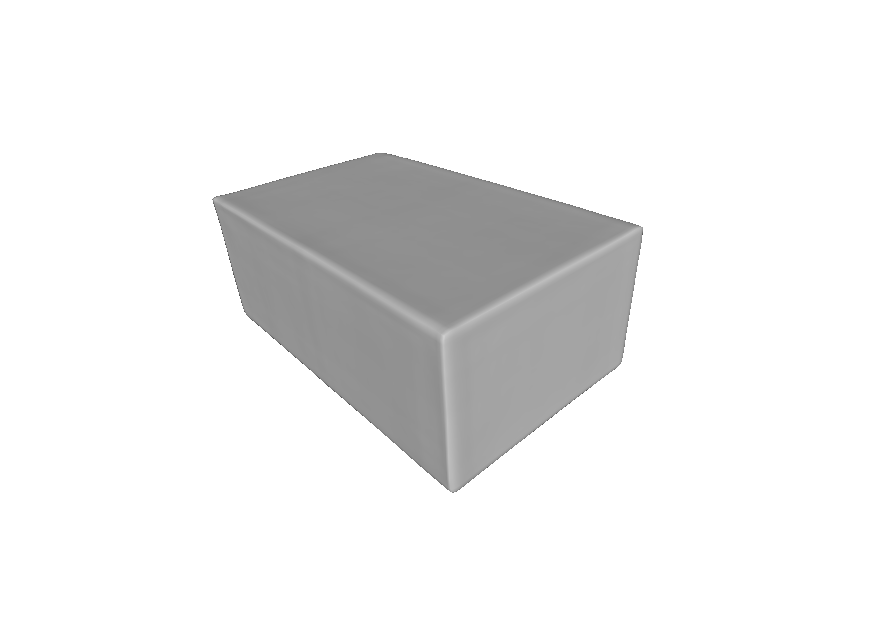} &
  \includegraphics[scale=0.15]{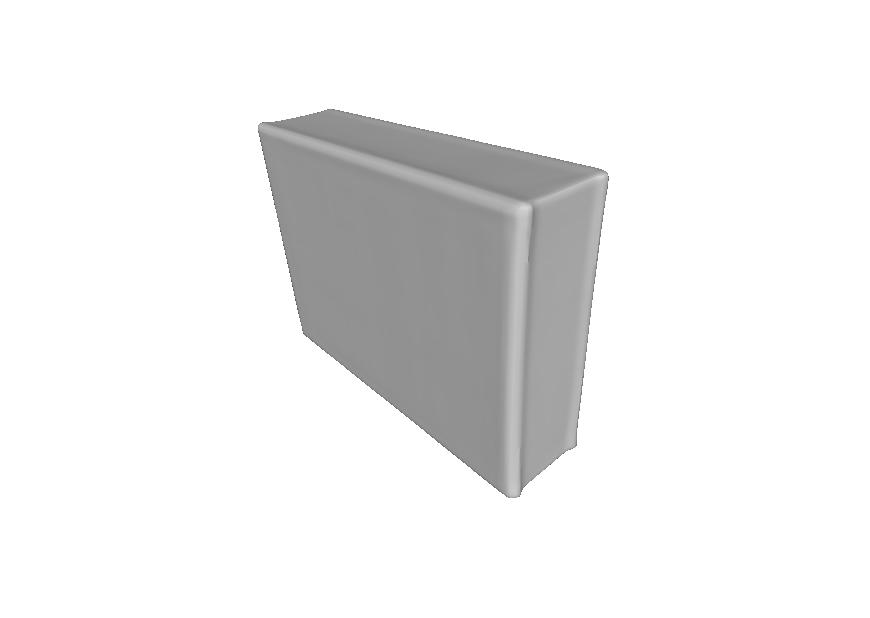} &
  \includegraphics[scale=0.15]{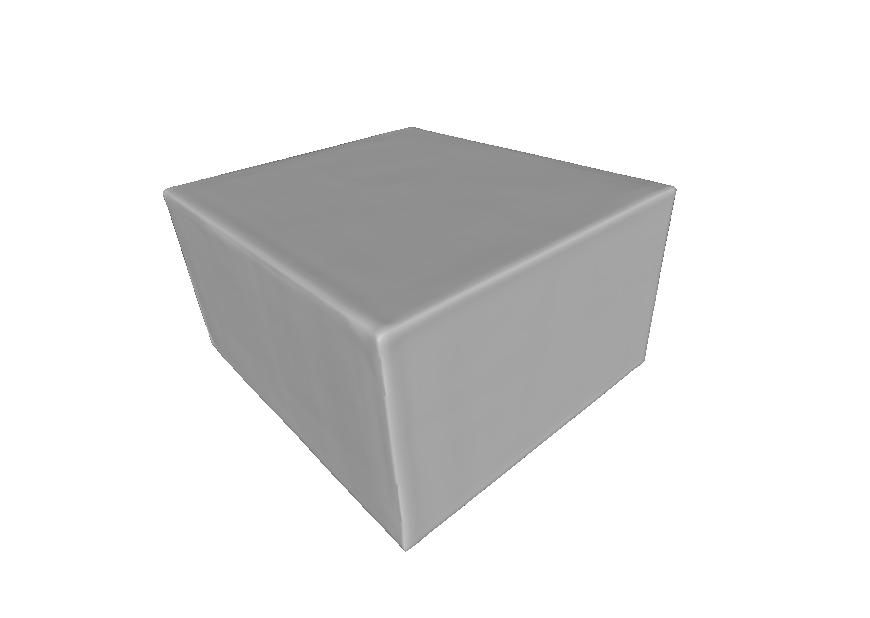} 
  \\ \hline
  
  \includegraphics[scale=0.35]{images/sketch_rec/14_in.jpg} &
  \includegraphics[scale=0.35]{images/sketch_rec/19_in.jpg} &
  \includegraphics[scale=0.35]{images/sketch_rec/23_in.jpg} &
  \includegraphics[scale=0.35]{images/sketch_rec/01_in.jpg} &
  \includegraphics[scale=0.35]{images/sketch_rec/10_in.jpg} 
  \\
  \includegraphics[scale=0.15]{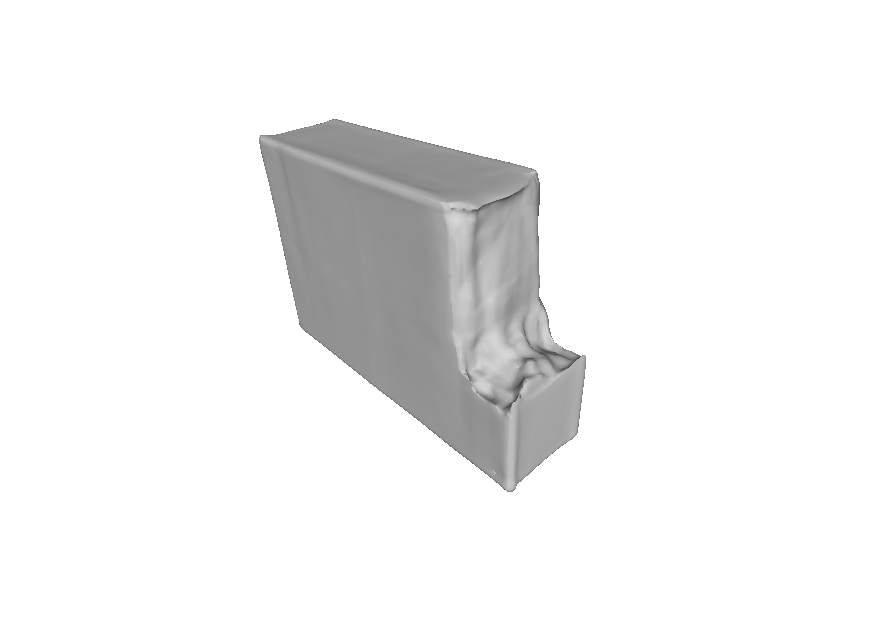} &
  \includegraphics[scale=0.15]{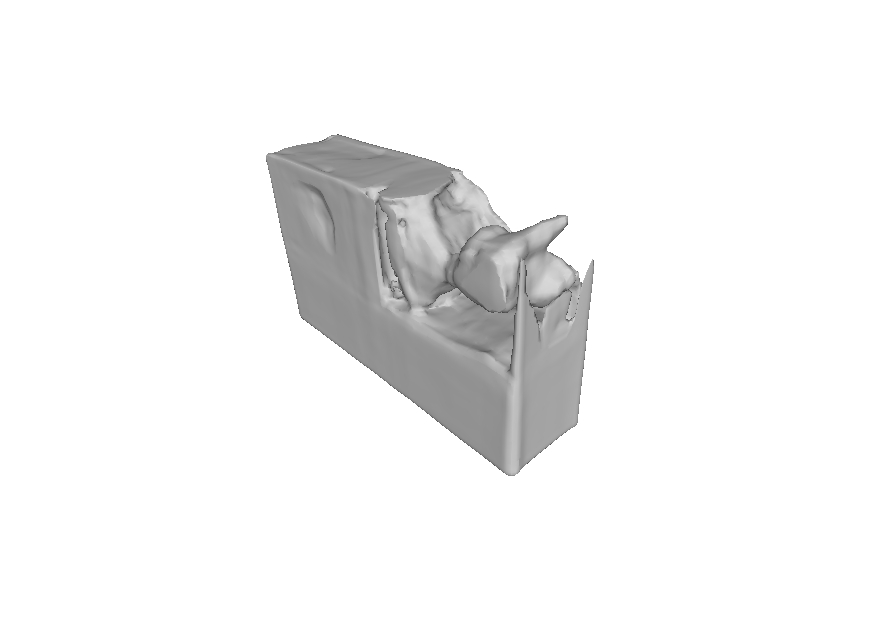} &
  \includegraphics[scale=0.15]{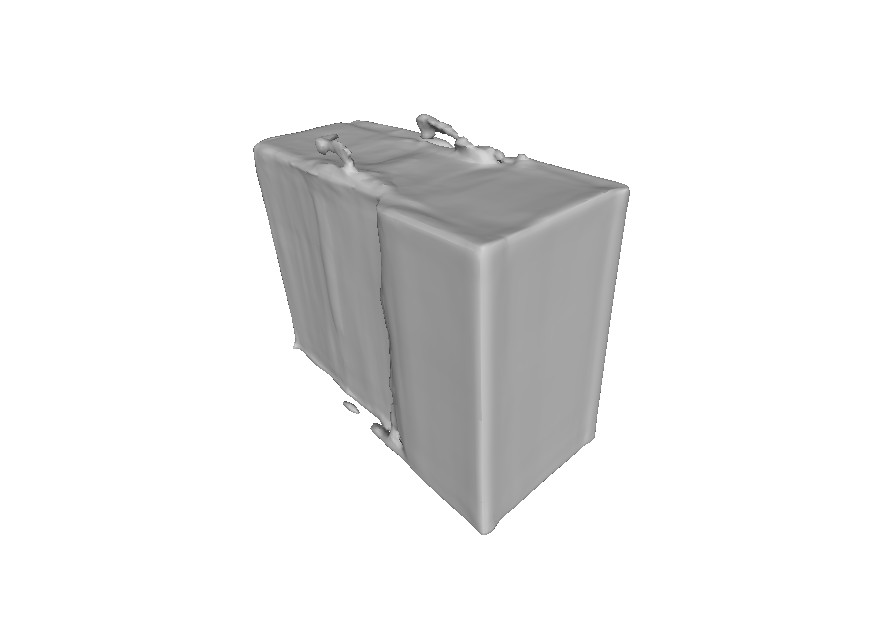} &
  \includegraphics[scale=0.15]{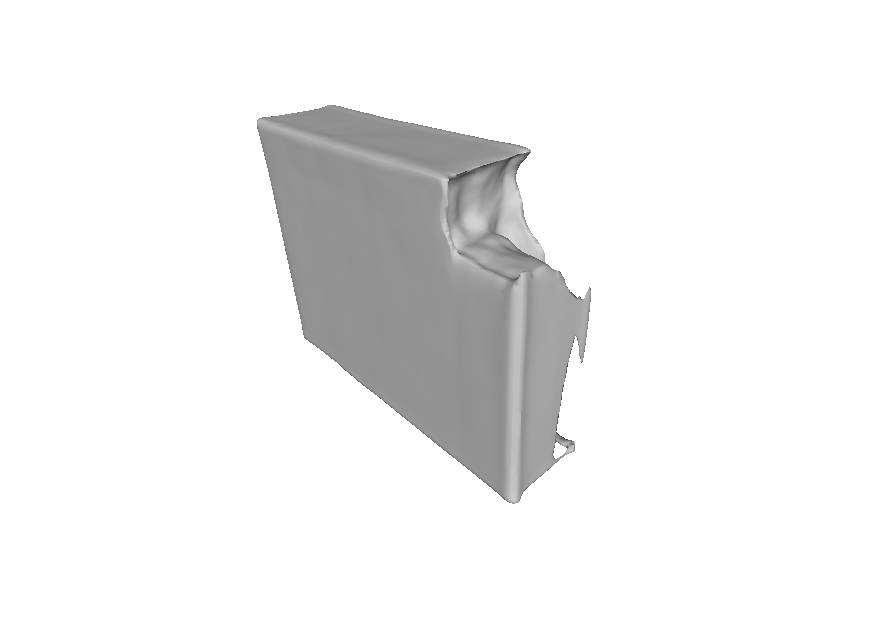} &
  \includegraphics[scale=0.15]{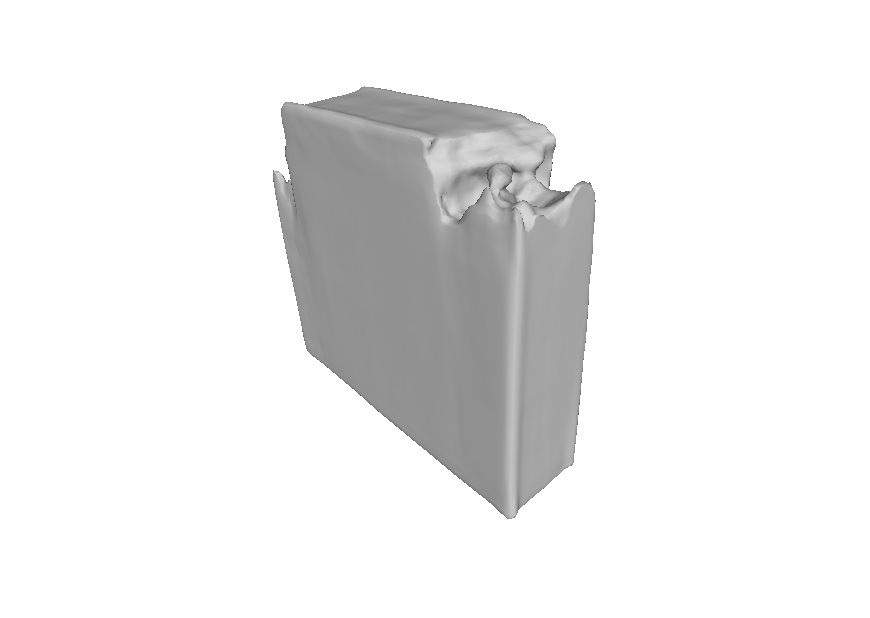}  
  \\ \hline
  
  \includegraphics[scale=0.35]{images/sketch_rec/02_in.jpg} &
  \includegraphics[scale=0.35]{images/sketch_rec/25_in.jpg} &
  \includegraphics[scale=0.35]{images/sketch_rec/08_in.jpg} &
  \includegraphics[scale=0.35]{images/sketch_rec/27_in.jpg} &
 
  \\ 
 \includegraphics[scale=0.15]{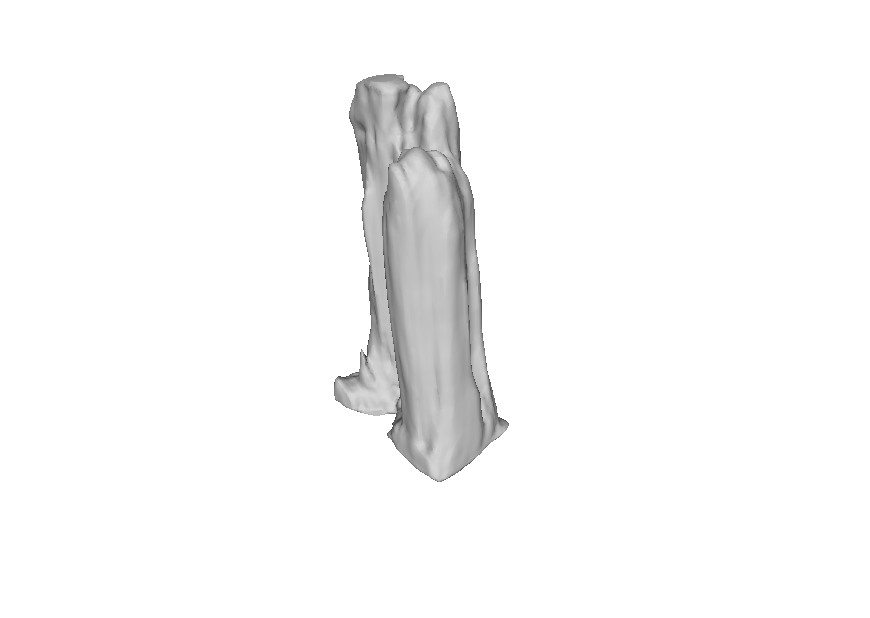} &
  \includegraphics[scale=0.15]{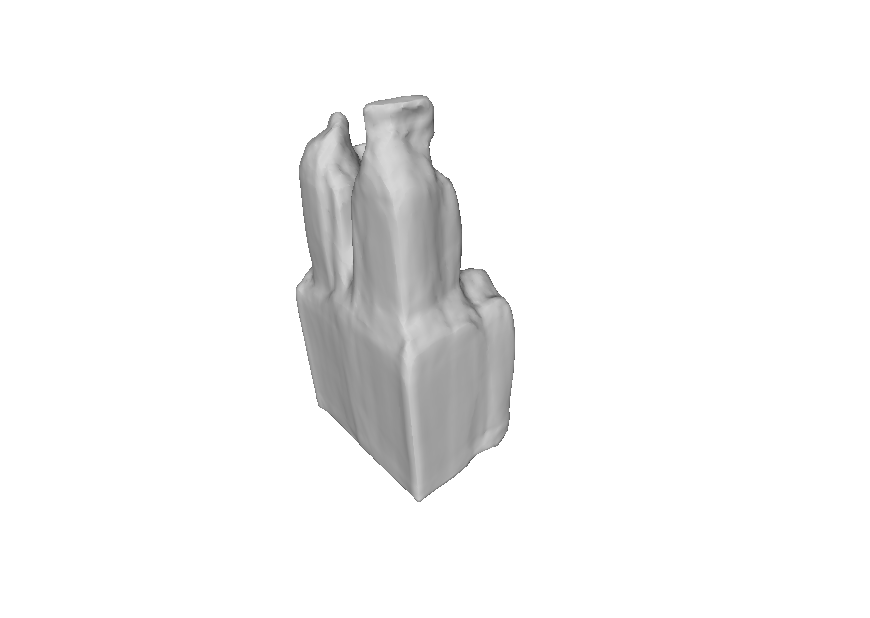} &
  \includegraphics[scale=0.15]{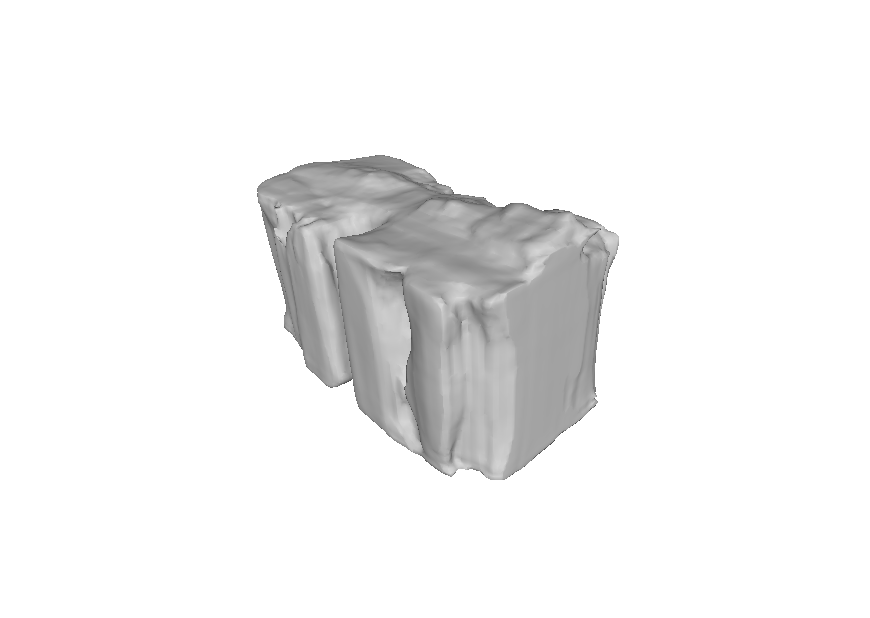} &
\includegraphics[scale=0.15]{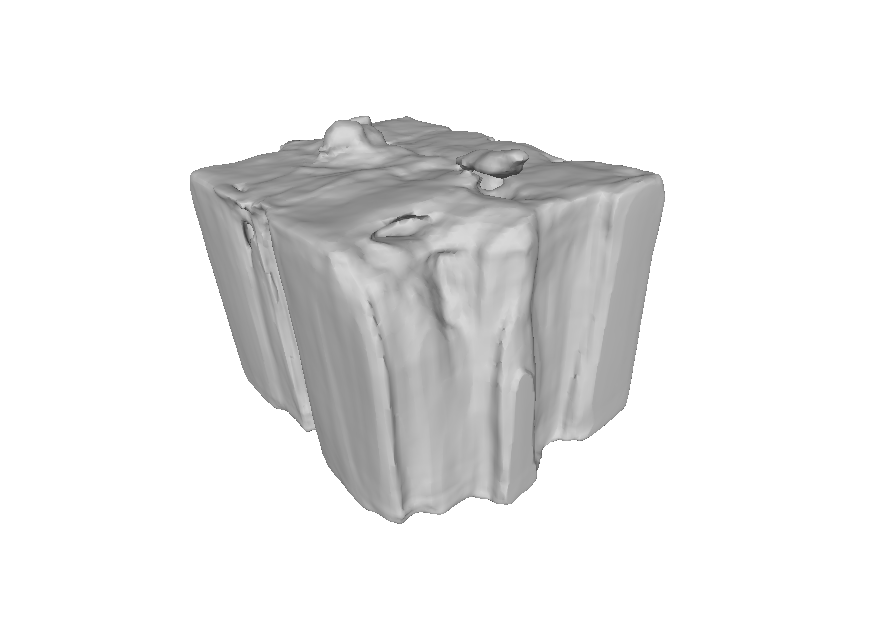} &
  \\   \\ \hline

\end{tabular}%
}
\caption{PFTA best model}
\end{table*}

\begin{table*}[]
\centering
\resizebox{\textwidth}{!}{%
\begin{tabular}{ccccc}
 \\ \hline
  \includegraphics[scale=0.35]{images/sketch_rec/29_in.jpg} &
  \includegraphics[scale=0.35]{images/sketch_rec/11_in.jpg} &
  \includegraphics[scale=0.35]{images/sketch_rec/03_in.jpg} &
  \includegraphics[scale=0.35]{images/sketch_rec/05_in.jpg} &
  \includegraphics[scale=0.35]{images/sketch_rec/07_in.jpg} 
  \\ 
  \includegraphics[scale=0.15]{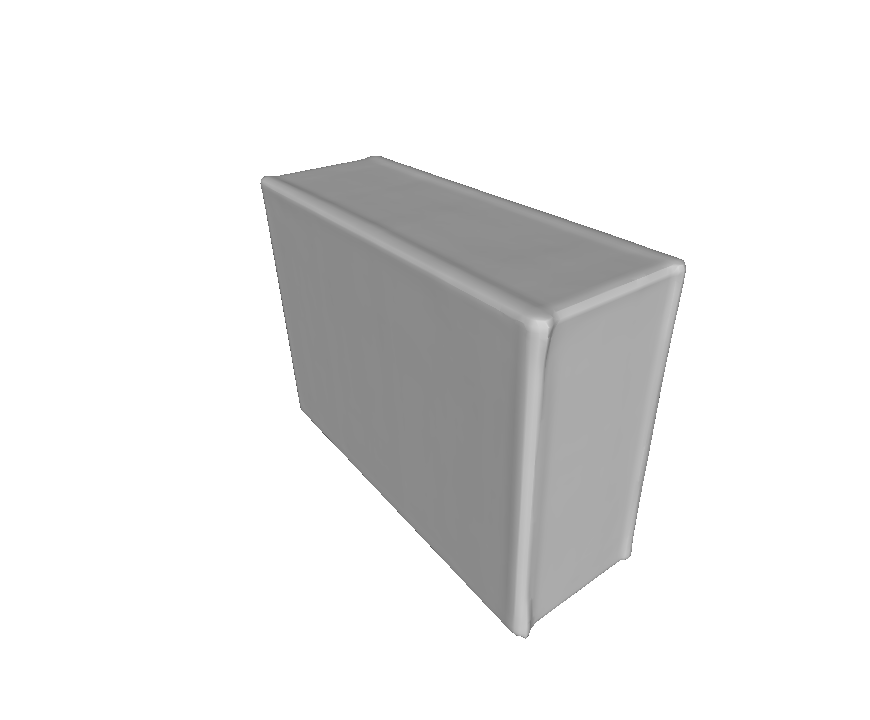} &
  \includegraphics[scale=0.15]{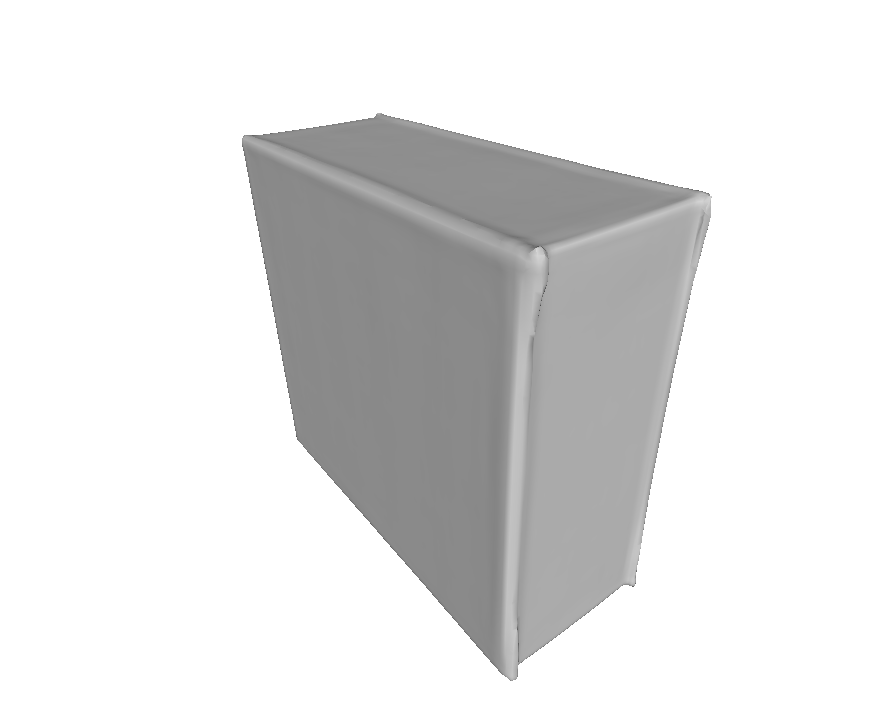} &
  \includegraphics[scale=0.15]{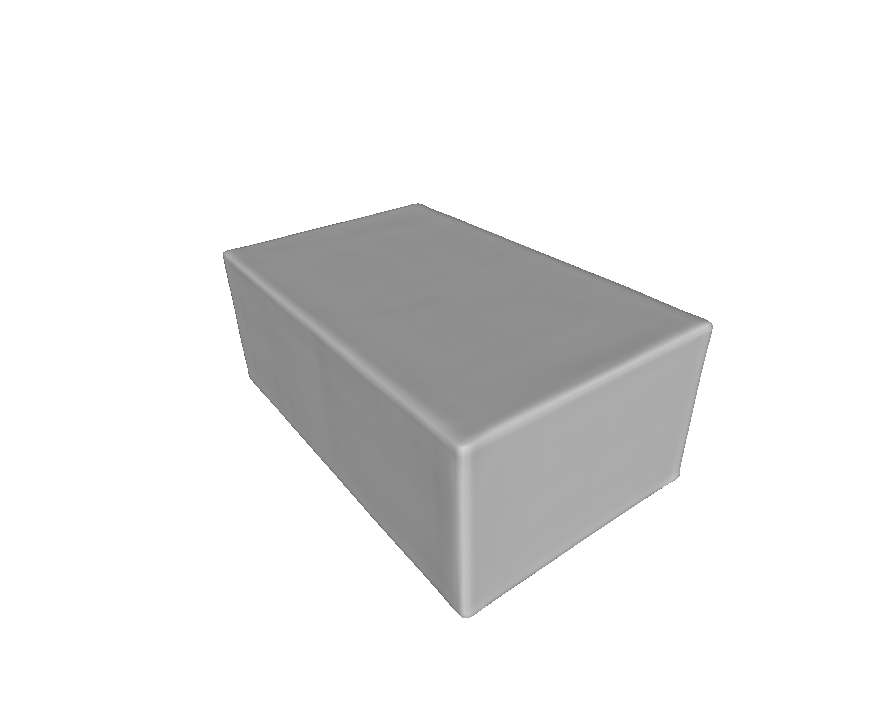} &
  \includegraphics[scale=0.15]{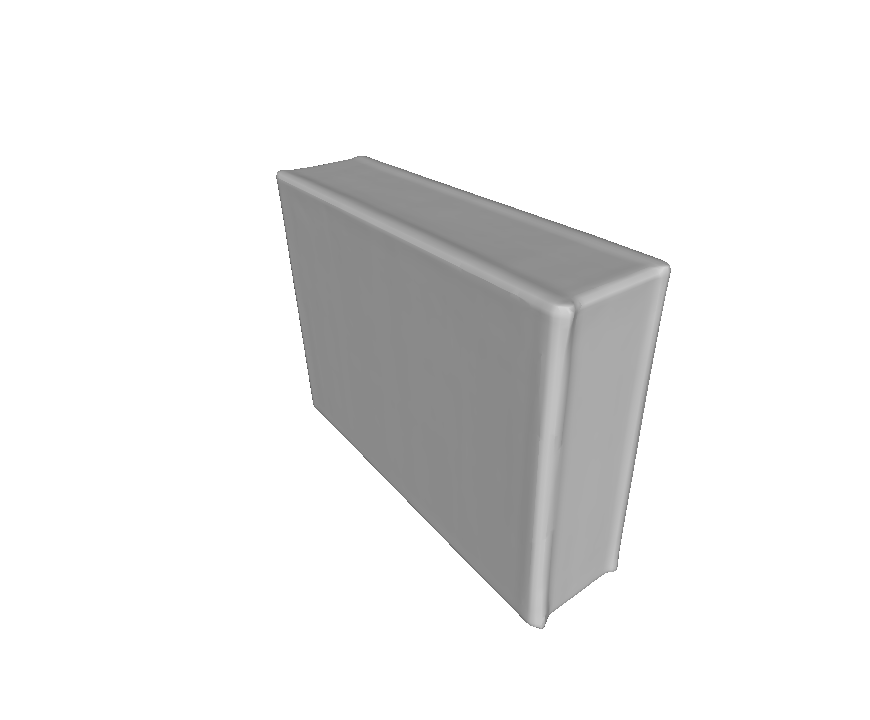} &
  \includegraphics[scale=0.15]{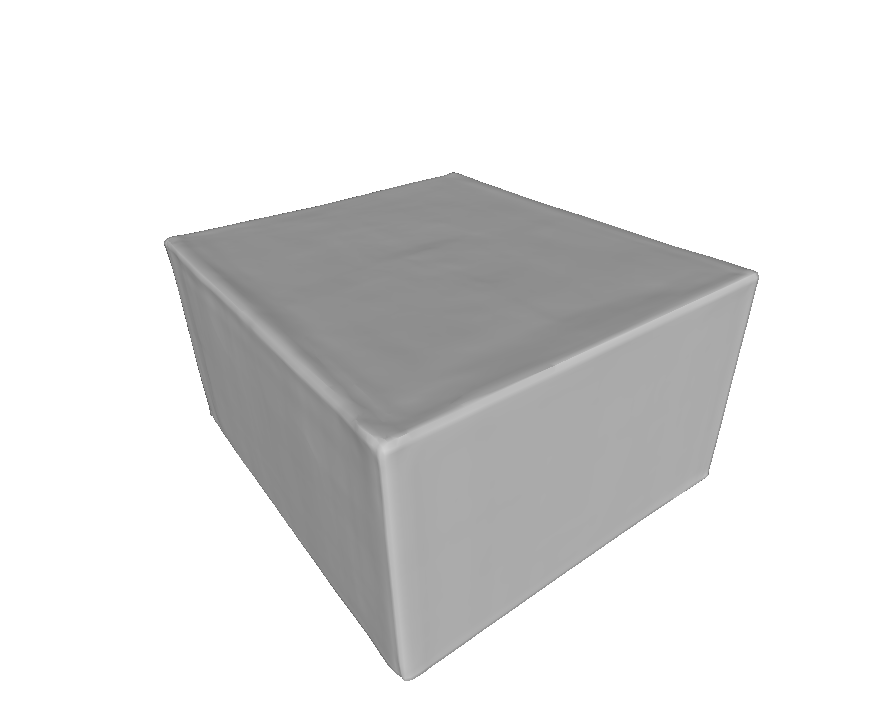} 
  \\ \hline
  
  \includegraphics[scale=0.35]{images/sketch_rec/14_in.jpg} &
  \includegraphics[scale=0.35]{images/sketch_rec/19_in.jpg} &
  \includegraphics[scale=0.35]{images/sketch_rec/23_in.jpg} &
  \includegraphics[scale=0.35]{images/sketch_rec/01_in.jpg} &
  \includegraphics[scale=0.35]{images/sketch_rec/10_in.jpg} 
  \\
  \includegraphics[scale=0.15]{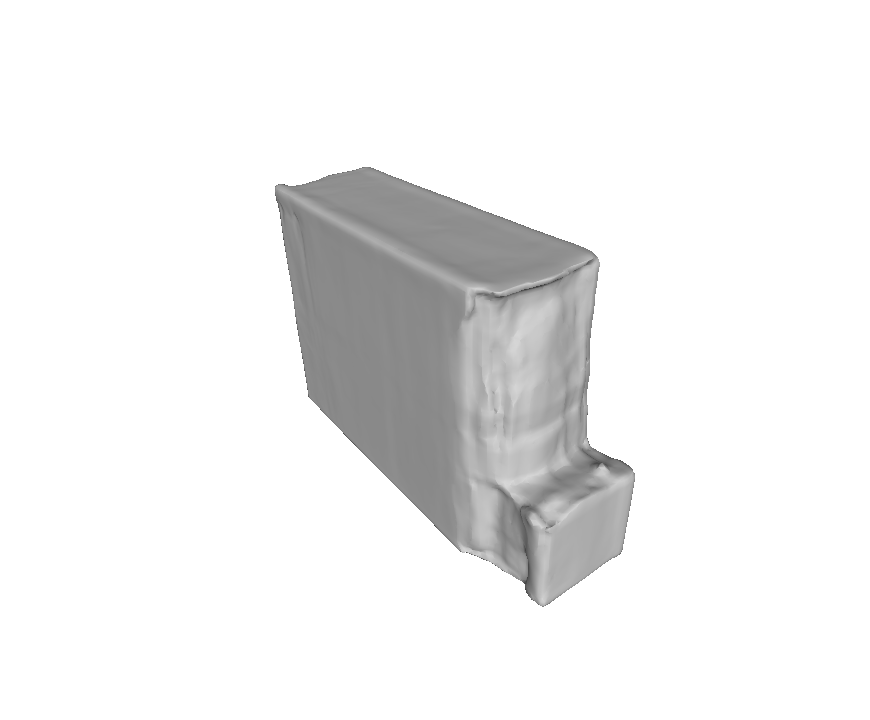} &
  \includegraphics[scale=0.15]{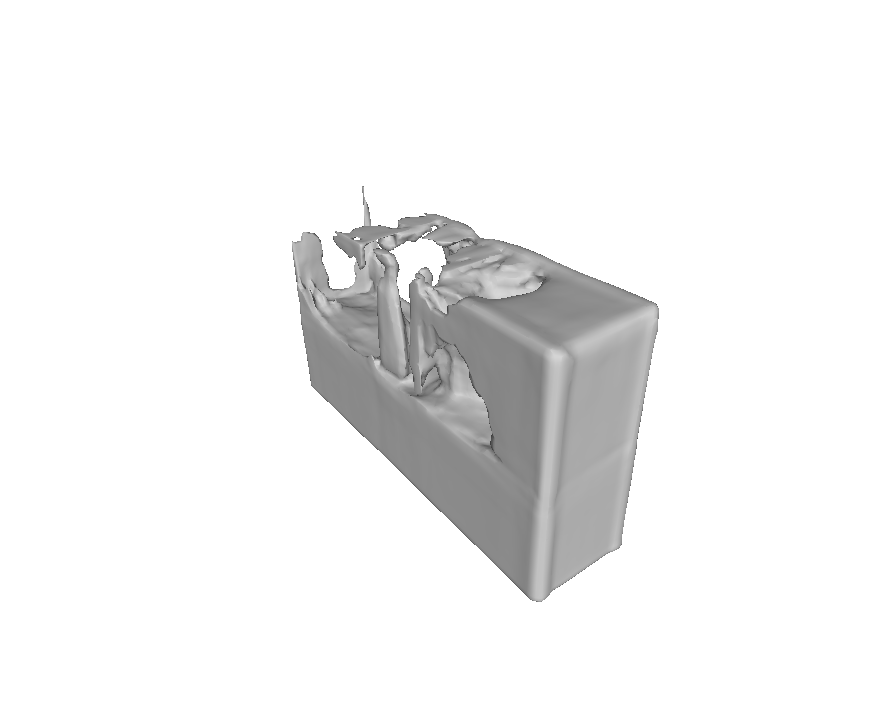} &
  \includegraphics[scale=0.15]{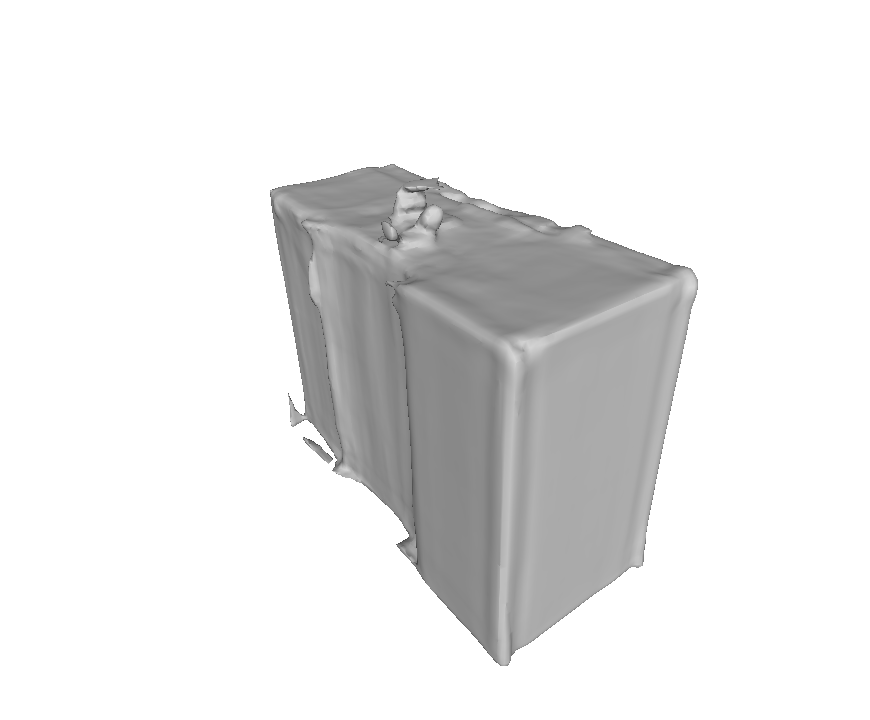} &
  \includegraphics[scale=0.15]{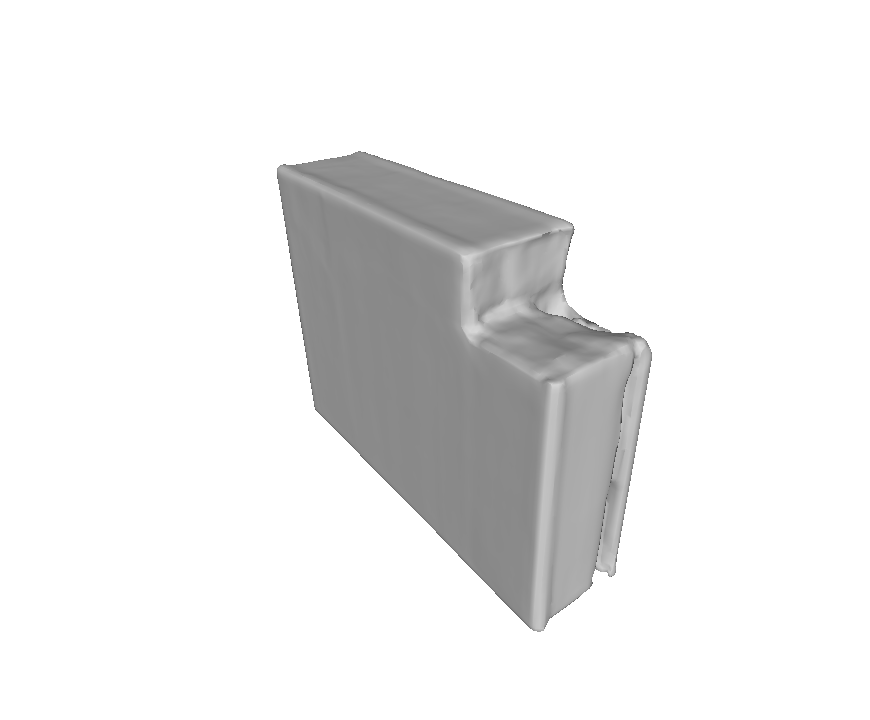} &
  \includegraphics[scale=0.15]{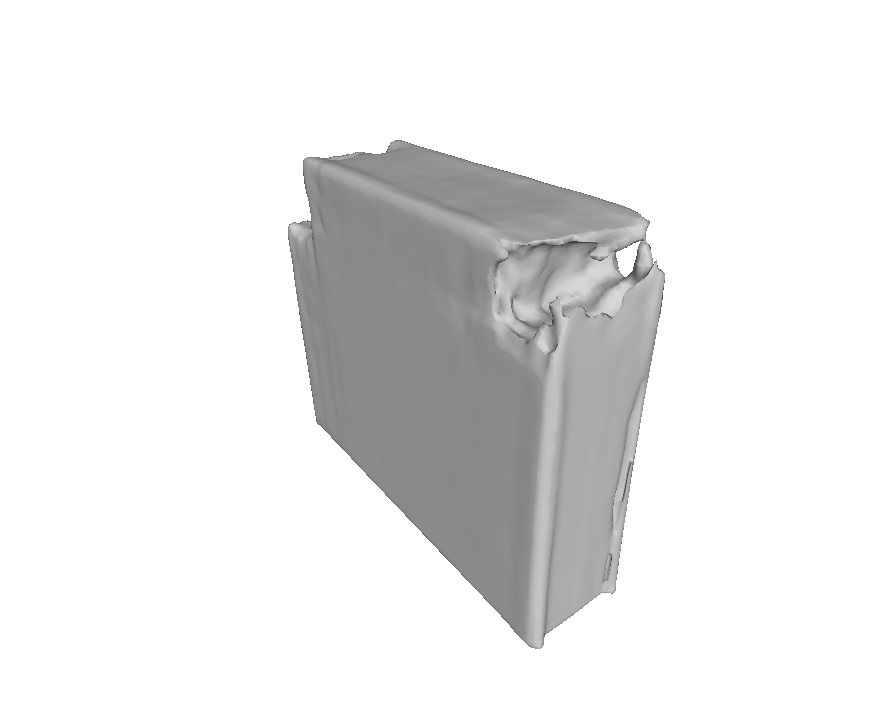}  
  \\ \hline
  
  \includegraphics[scale=0.35]{images/sketch_rec/02_in.jpg} &
  \includegraphics[scale=0.35]{images/sketch_rec/25_in.jpg} &
  \includegraphics[scale=0.35]{images/sketch_rec/08_in.jpg} &
  \includegraphics[scale=0.35]{images/sketch_rec/27_in.jpg} &
 
  \\ 
 \includegraphics[scale=0.15]{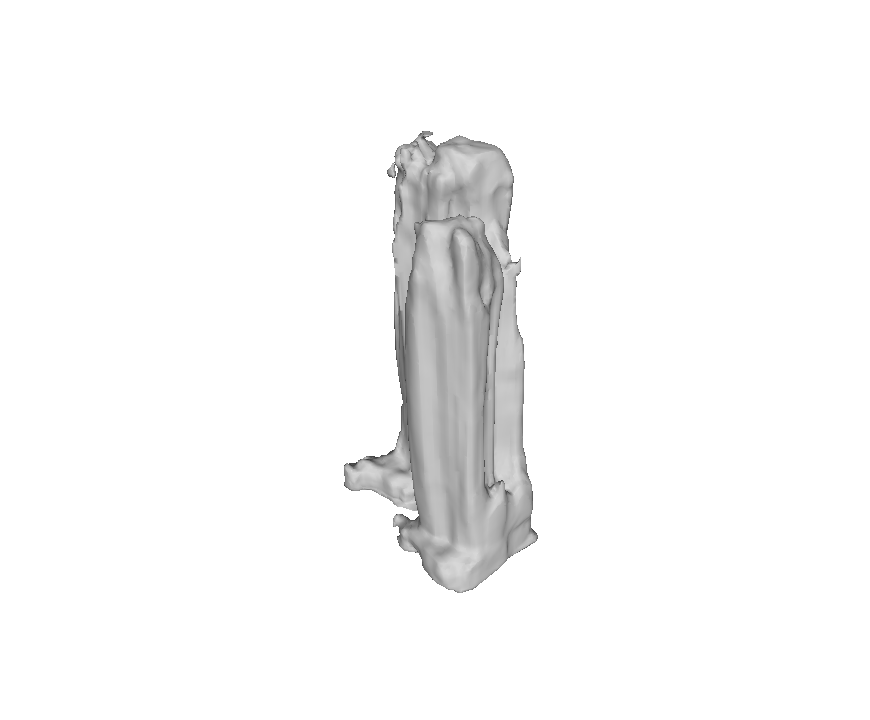} &
  \includegraphics[scale=0.15]{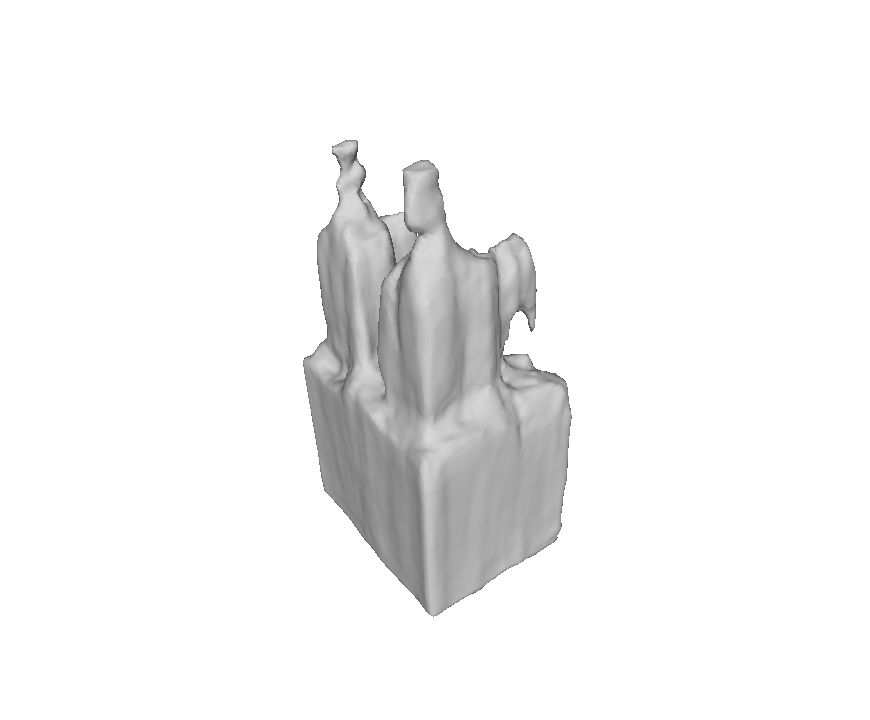} &
  \includegraphics[scale=0.15]{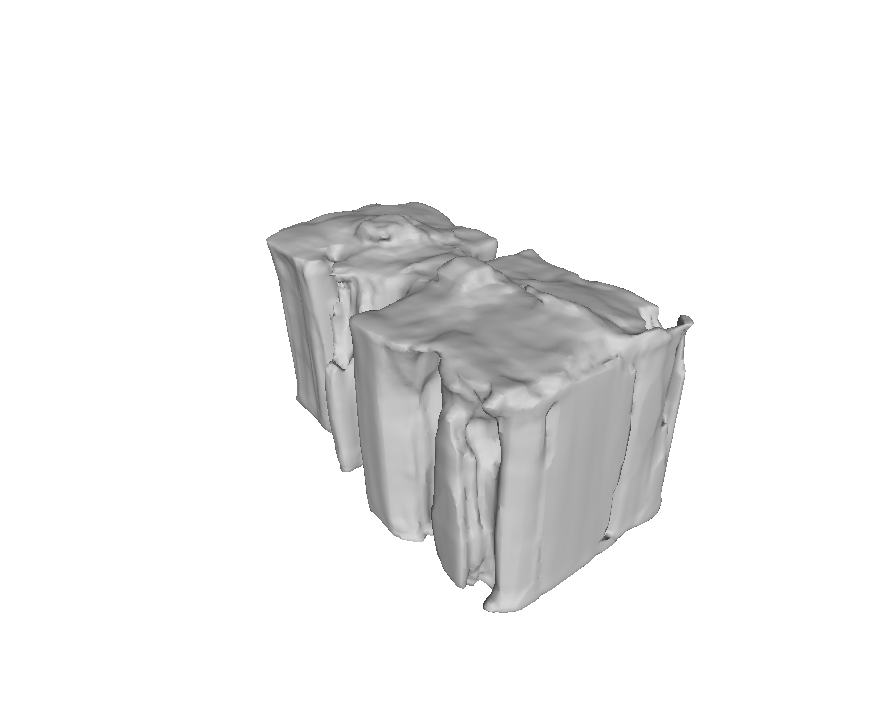} &
\includegraphics[scale=0.15]{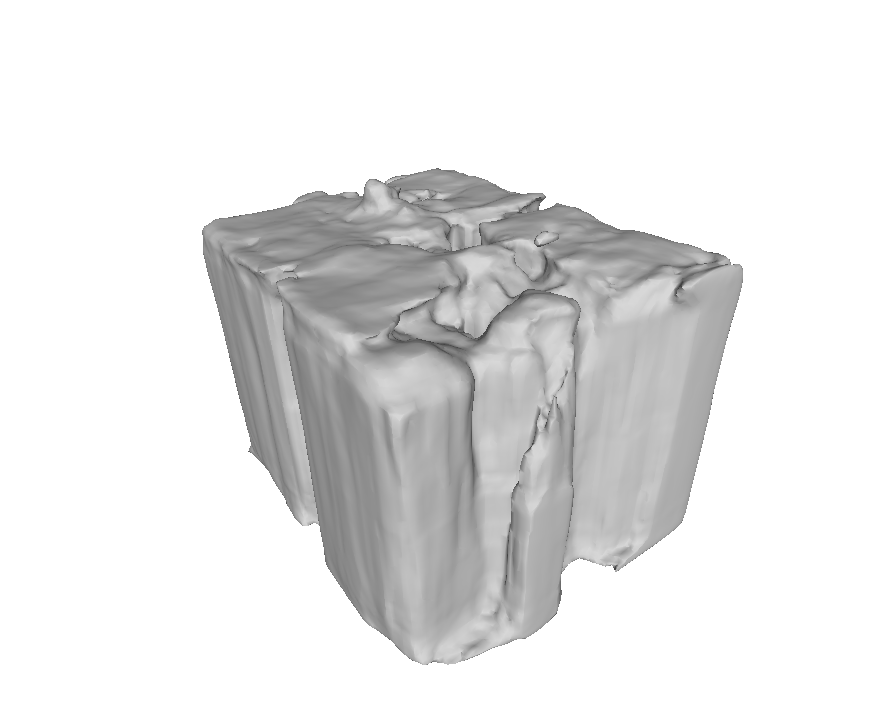} &
  \\   \\ \hline

\end{tabular}%
}
\caption{PFTA latest model, maximize IoU, capable to capture better fine-grain details in the more complex buildings, like the two towers and the edges in the other buildings}
\end{table*}

\begin{table*}[]
\centering
\resizebox{\textwidth}{!}{%
\begin{tabular}{ccccc}
 \\ \hline
  \includegraphics[scale=0.35]{images/sketch_rec/29_in.jpg} &
  \includegraphics[scale=0.35]{images/sketch_rec/11_in.jpg} &
  \includegraphics[scale=0.35]{images/sketch_rec/03_in.jpg} &
  \includegraphics[scale=0.35]{images/sketch_rec/05_in.jpg} &
  \includegraphics[scale=0.35]{images/sketch_rec/07_in.jpg} 
  \\ 
  \includegraphics[scale=0.15]{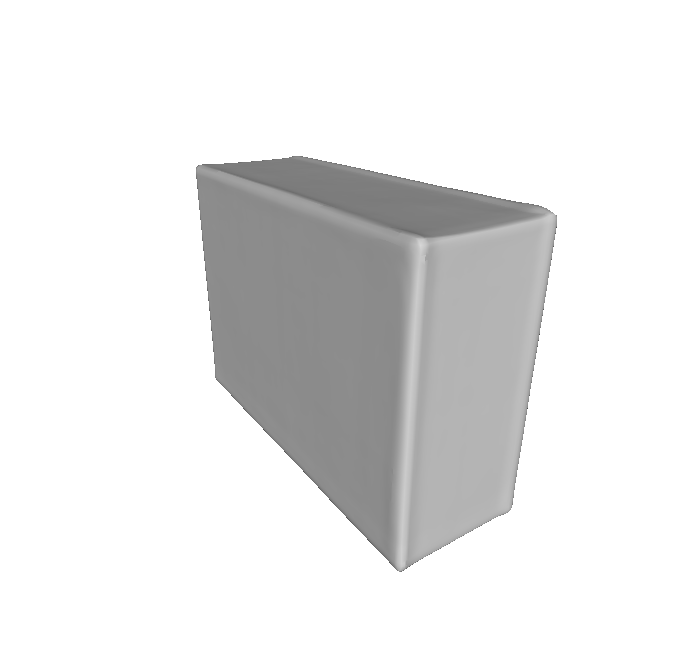} &
  \includegraphics[scale=0.15]{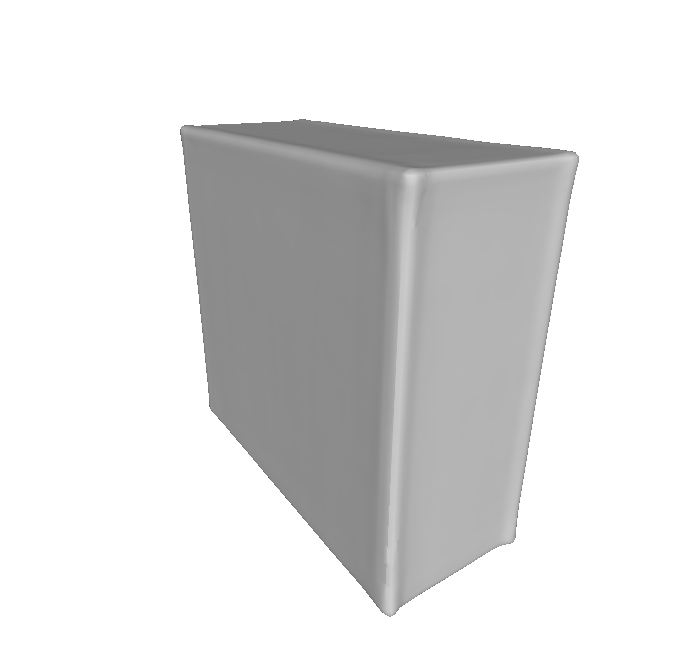} &
  \includegraphics[scale=0.15]{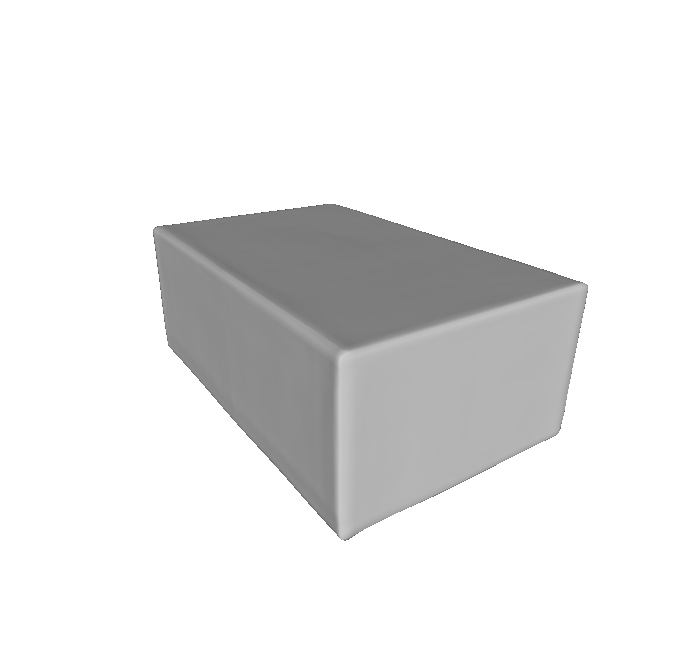} &
  \includegraphics[scale=0.15]{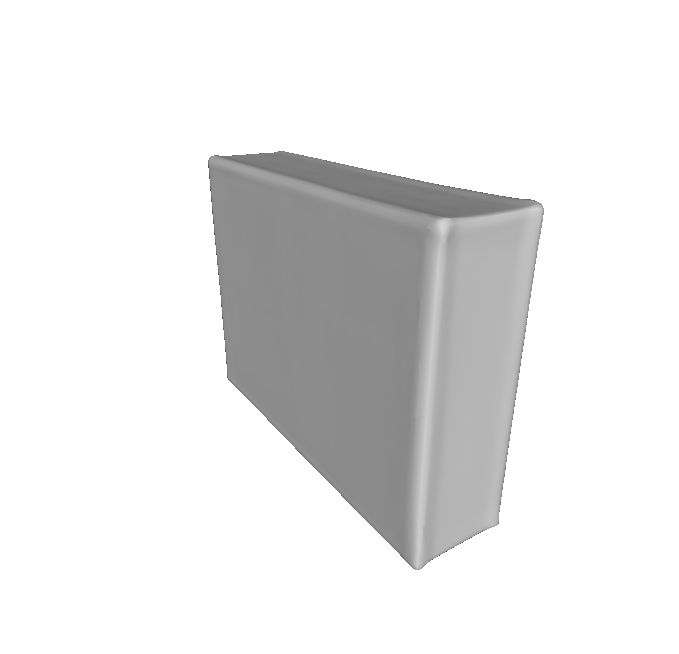} &
  \includegraphics[scale=0.15]{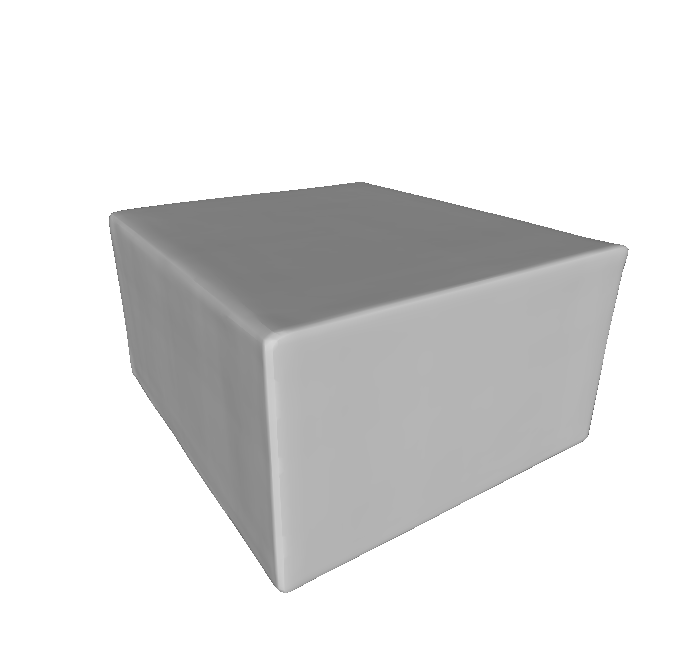} 
  \\ \hline
  
  \includegraphics[scale=0.35]{images/sketch_rec/14_in.jpg} &
  \includegraphics[scale=0.35]{images/sketch_rec/19_in.jpg} &
  \includegraphics[scale=0.35]{images/sketch_rec/23_in.jpg} &
  \includegraphics[scale=0.35]{images/sketch_rec/01_in.jpg} &
  \includegraphics[scale=0.35]{images/sketch_rec/10_in.jpg} 
  \\
  \includegraphics[scale=0.15]{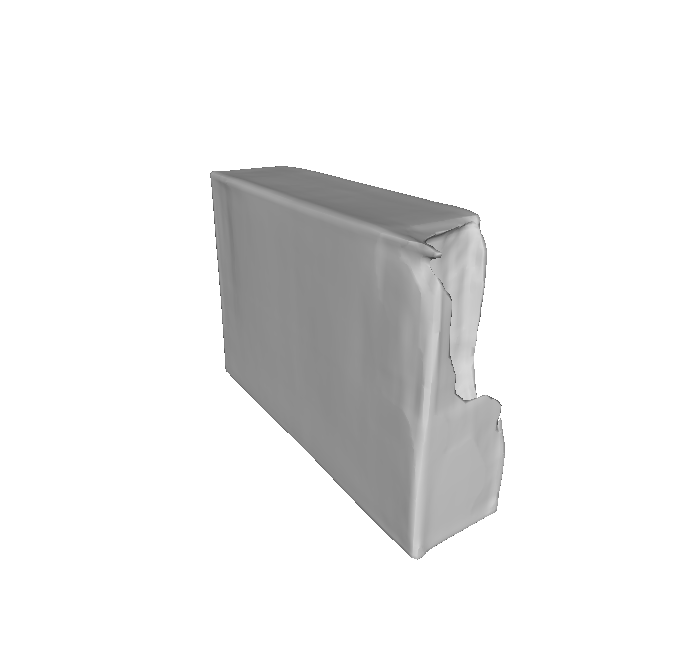} &
  \includegraphics[scale=0.15]{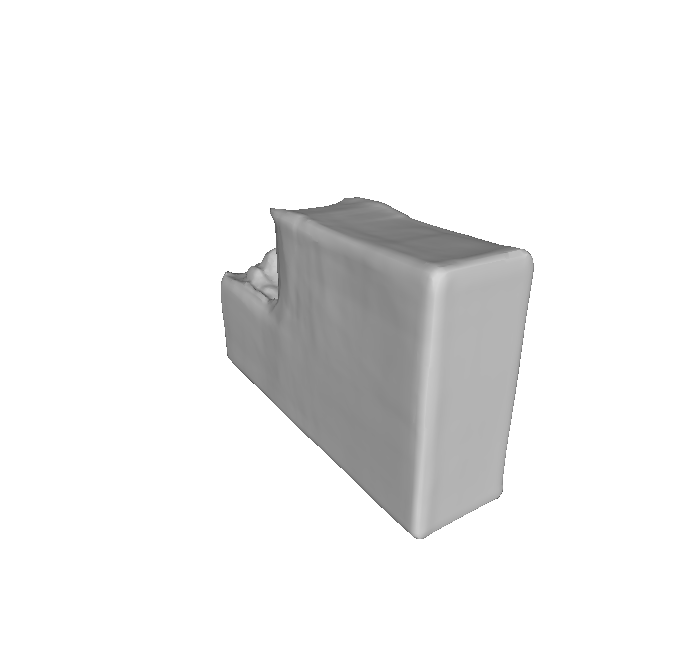} &
  \includegraphics[scale=0.15]{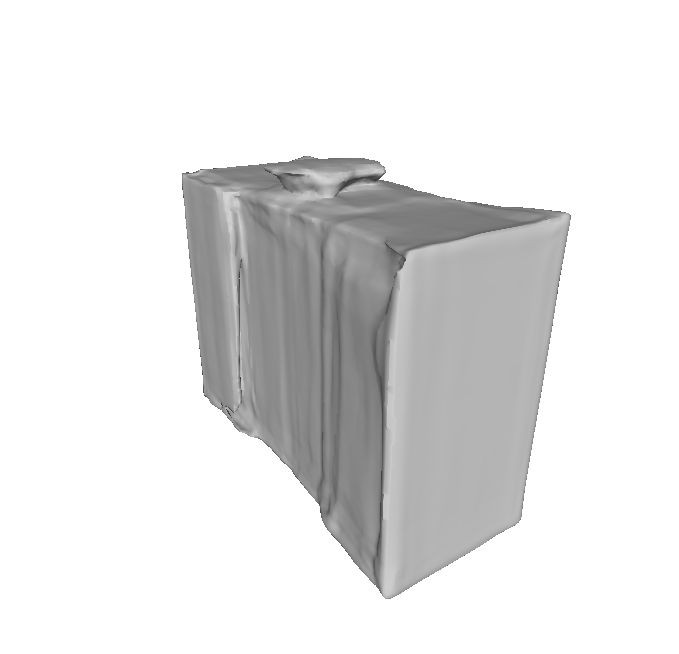} &
  \includegraphics[scale=0.15]{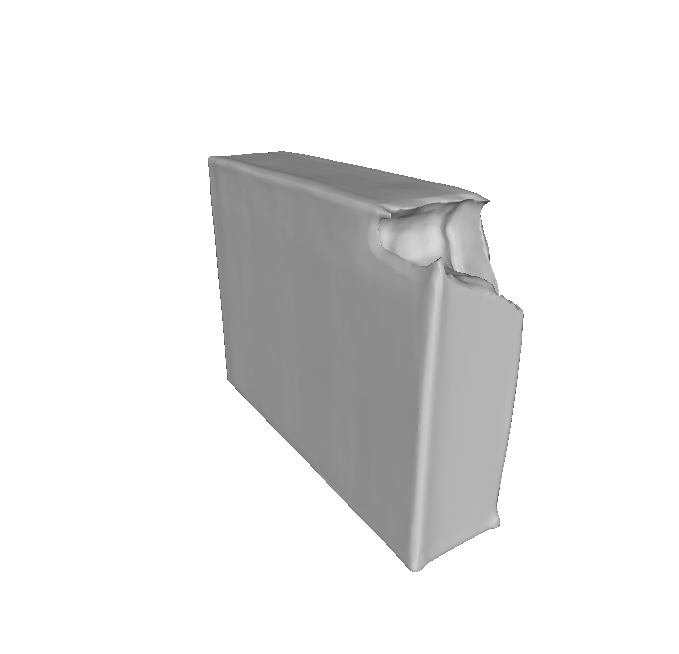} &
  \includegraphics[scale=0.15]{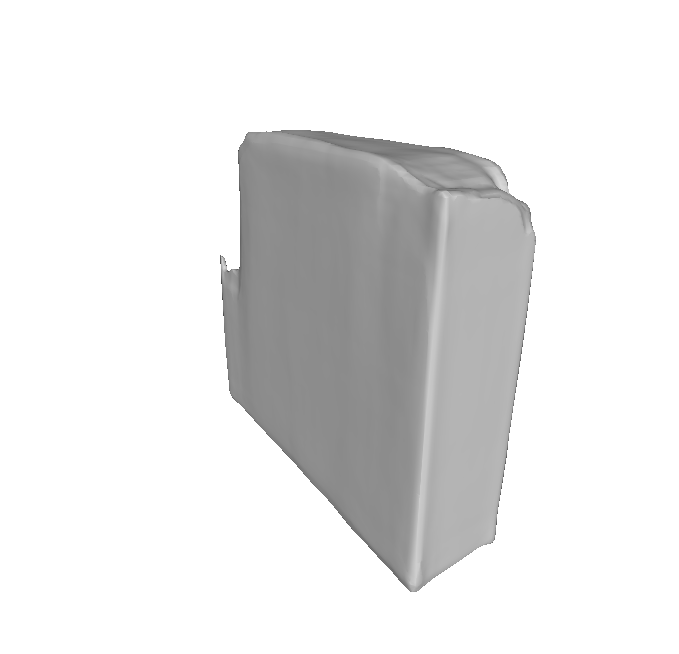}  
  \\ \hline
  
  \includegraphics[scale=0.35]{images/sketch_rec/02_in.jpg} &
  \includegraphics[scale=0.35]{images/sketch_rec/25_in.jpg} &
  \includegraphics[scale=0.35]{images/sketch_rec/08_in.jpg} &
  \includegraphics[scale=0.35]{images/sketch_rec/27_in.jpg} &
 
  \\ 
 \includegraphics[scale=0.15]{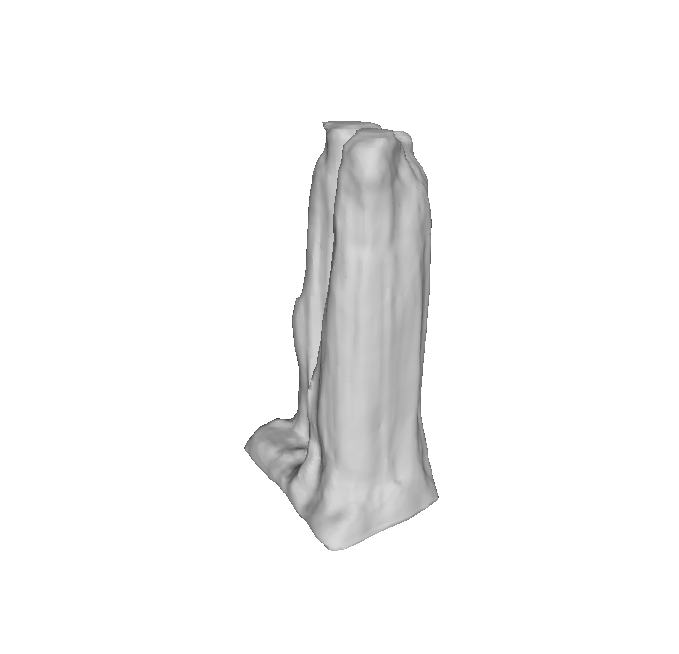} &
  \includegraphics[scale=0.15]{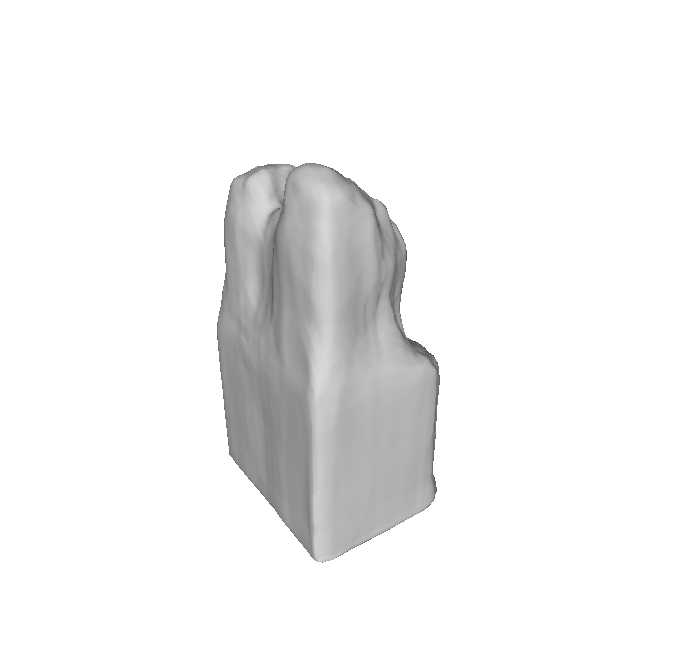} &
  \includegraphics[scale=0.15]{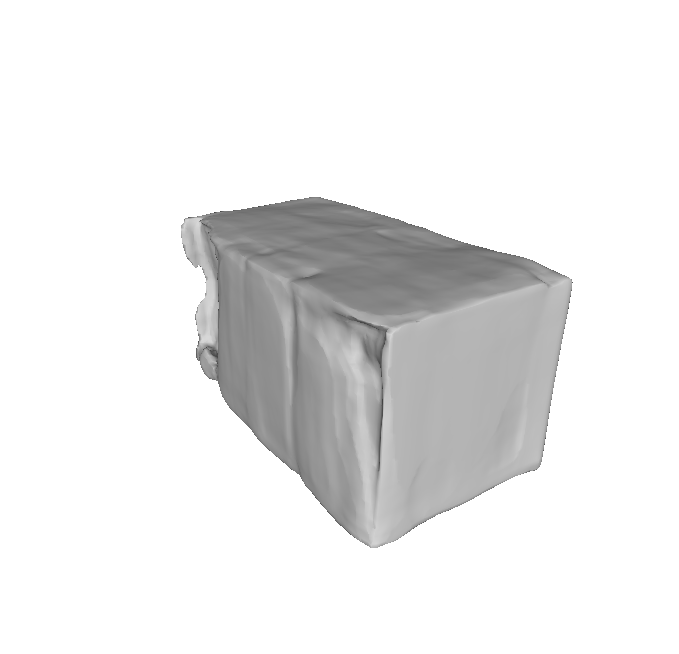} &
\includegraphics[scale=0.15]{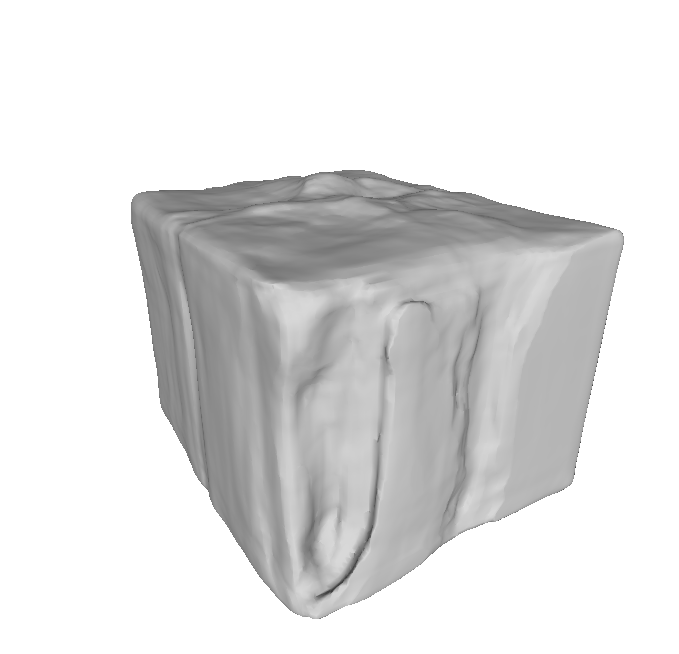} &
  \\   \\ \hline

\end{tabular}%
}
\caption{PFTA 30 best model, maximize IoU. Since it has been trained only for 30 minutes, we can definitely see that it hasn't received enough training, also the loss is pretty high compared with the other methods}
\end{table*}

\begin{table*}[]
\centering
\resizebox{\textwidth}{!}{%
\begin{tabular}{ccccc}
 \\ \hline
  \includegraphics[scale=0.35]{images/sketch_rec/29_in.jpg} &
  \includegraphics[scale=0.35]{images/sketch_rec/11_in.jpg} &
  \includegraphics[scale=0.35]{images/sketch_rec/03_in.jpg} &
  \includegraphics[scale=0.35]{images/sketch_rec/05_in.jpg} &
  \includegraphics[scale=0.35]{images/sketch_rec/07_in.jpg} 
  \\ 
  \includegraphics[scale=0.15]{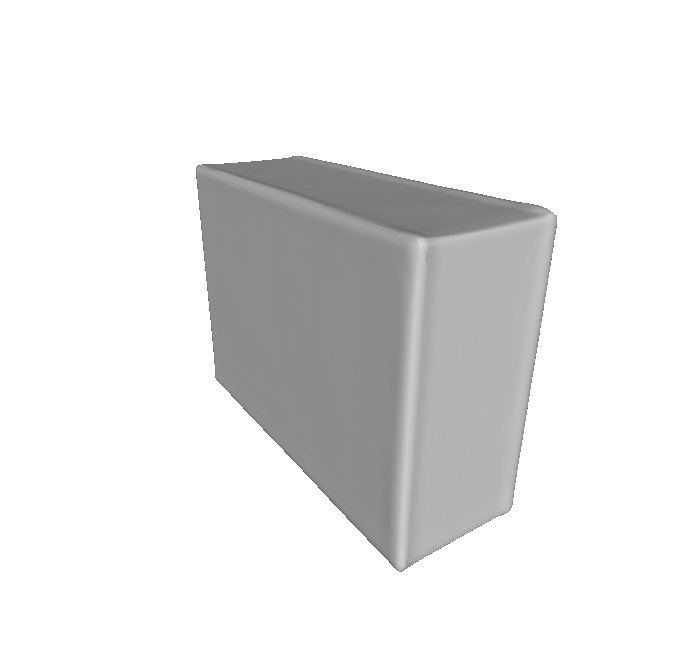} &
  \includegraphics[scale=0.15]{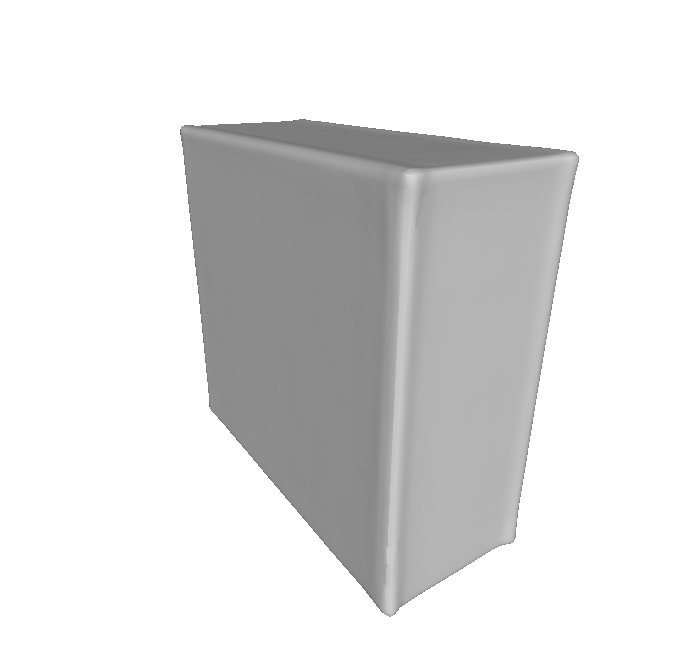} &
  \includegraphics[scale=0.15]{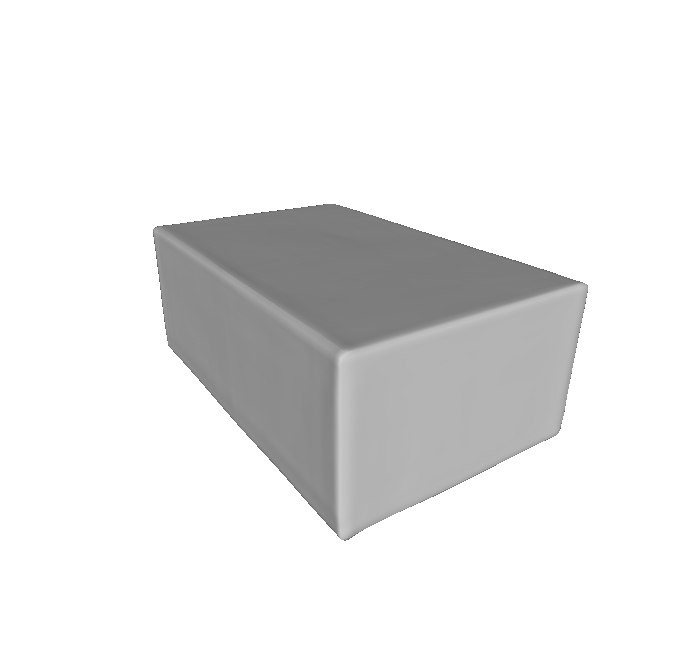} &
  \includegraphics[scale=0.15]{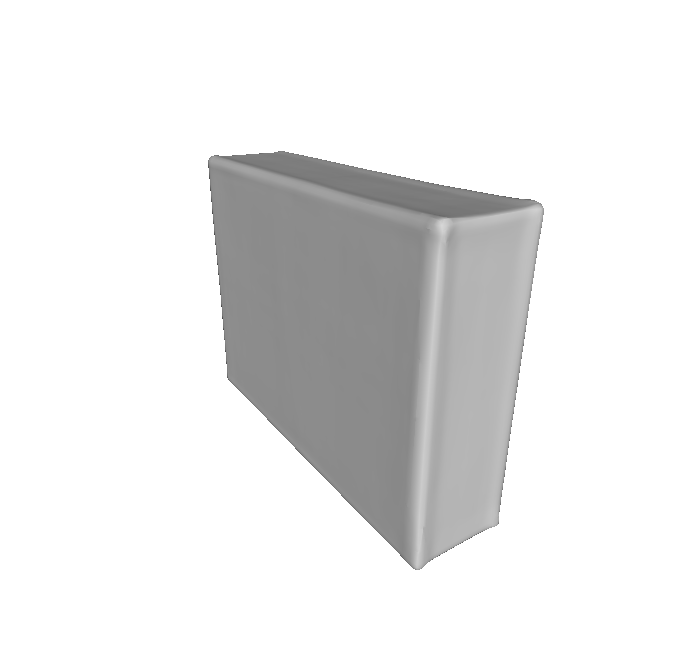} &
  \includegraphics[scale=0.15]{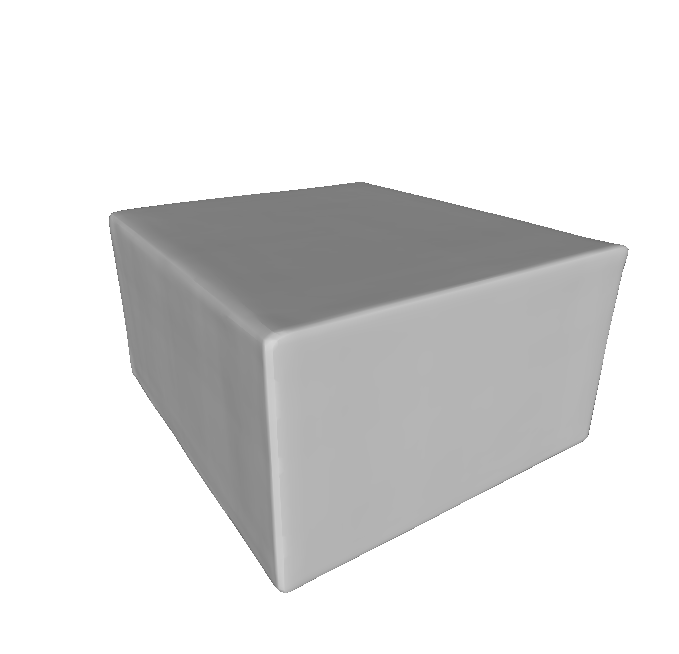} 
  \\ \hline
  
  \includegraphics[scale=0.35]{images/sketch_rec/14_in.jpg} &
  \includegraphics[scale=0.35]{images/sketch_rec/19_in.jpg} &
  \includegraphics[scale=0.35]{images/sketch_rec/23_in.jpg} &
  \includegraphics[scale=0.35]{images/sketch_rec/01_in.jpg} &
  \includegraphics[scale=0.35]{images/sketch_rec/10_in.jpg} 
  \\
  \includegraphics[scale=0.15]{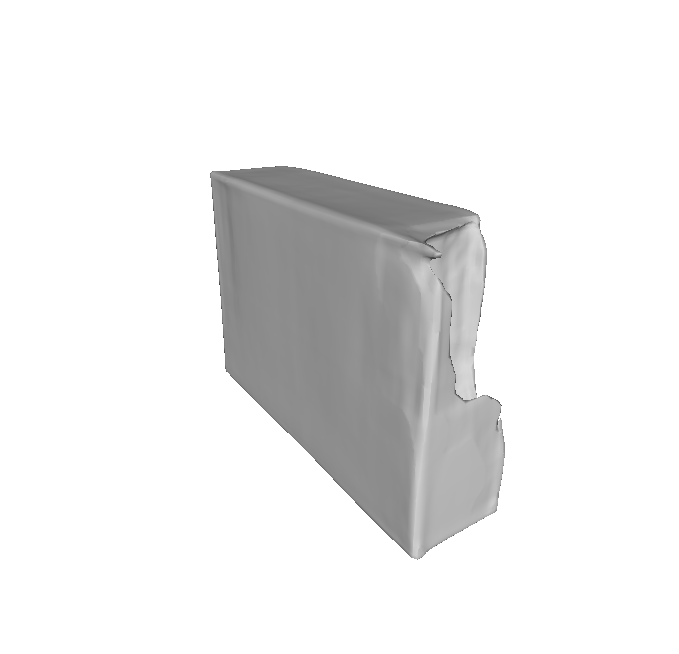} &
  \includegraphics[scale=0.15]{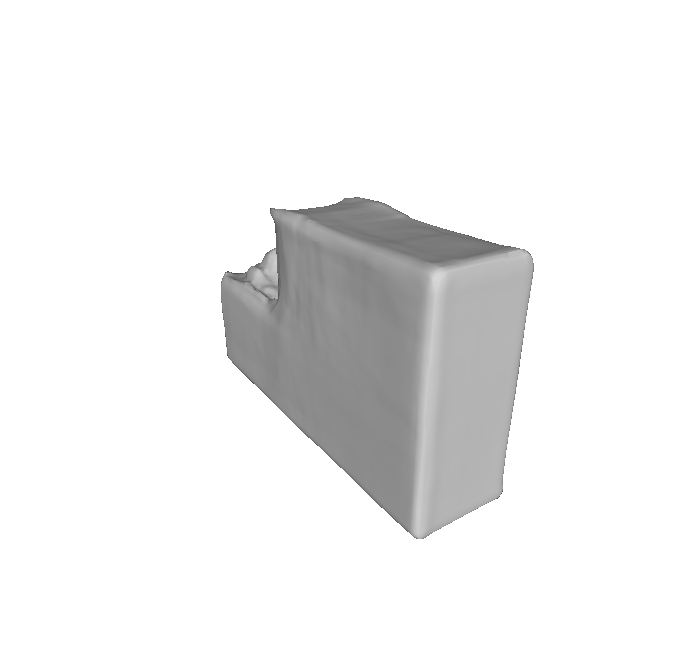} &
  \includegraphics[scale=0.15]{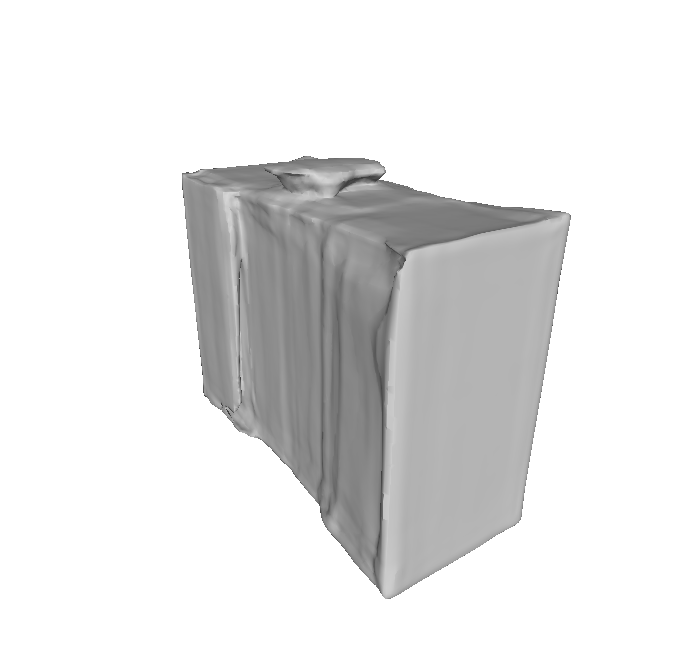} &
  \includegraphics[scale=0.15]{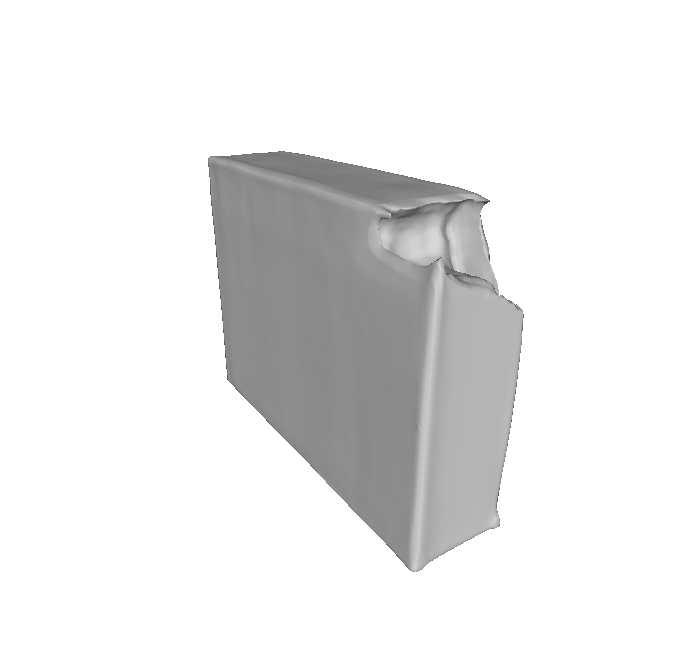} &
  \includegraphics[scale=0.15]{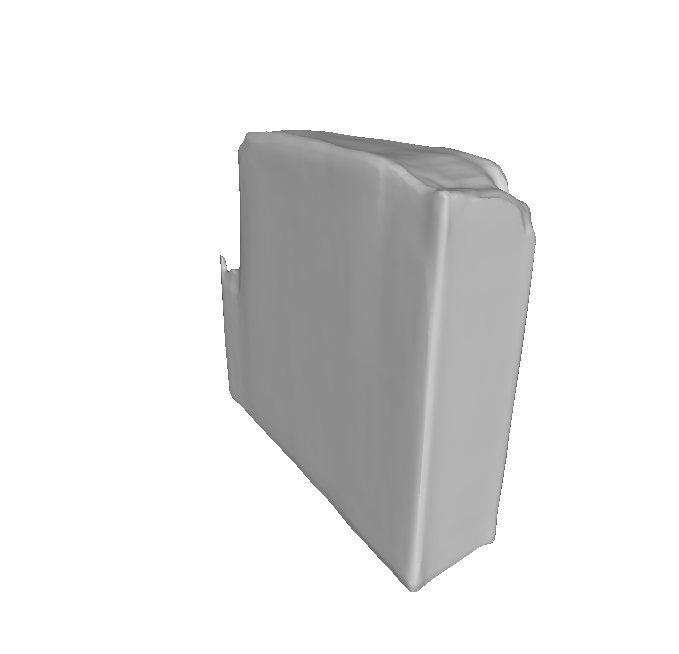}  
  \\ \hline
  
  \includegraphics[scale=0.35]{images/sketch_rec/02_in.jpg} &
  \includegraphics[scale=0.35]{images/sketch_rec/25_in.jpg} &
  \includegraphics[scale=0.35]{images/sketch_rec/08_in.jpg} &
  \includegraphics[scale=0.35]{images/sketch_rec/27_in.jpg} &
 
  \\ 
 \includegraphics[scale=0.15]{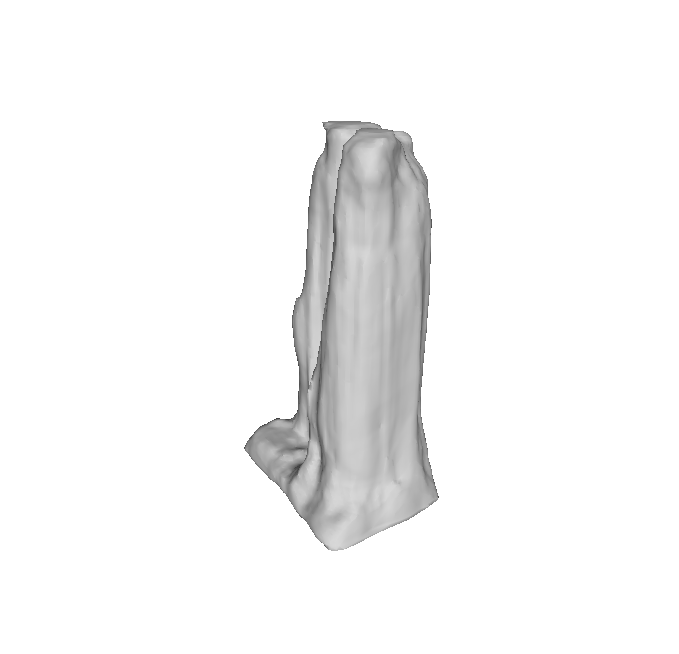} &
  \includegraphics[scale=0.15]{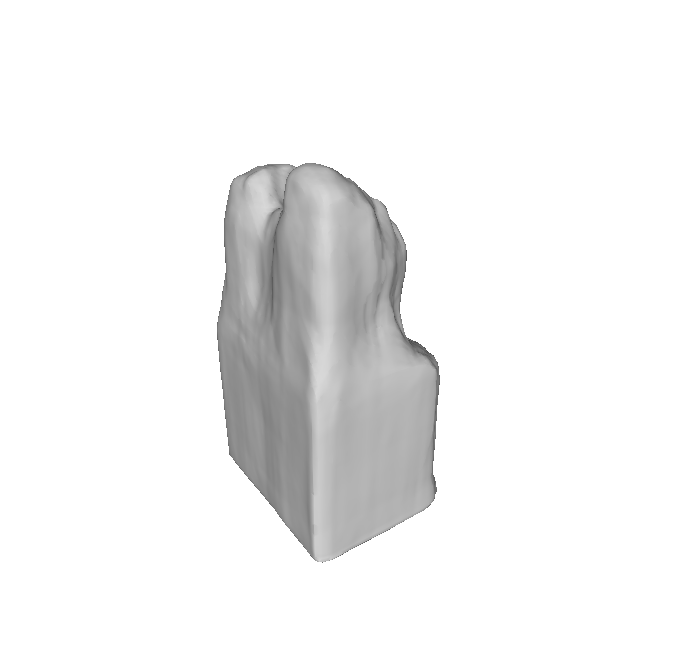} &
  \includegraphics[scale=0.15]{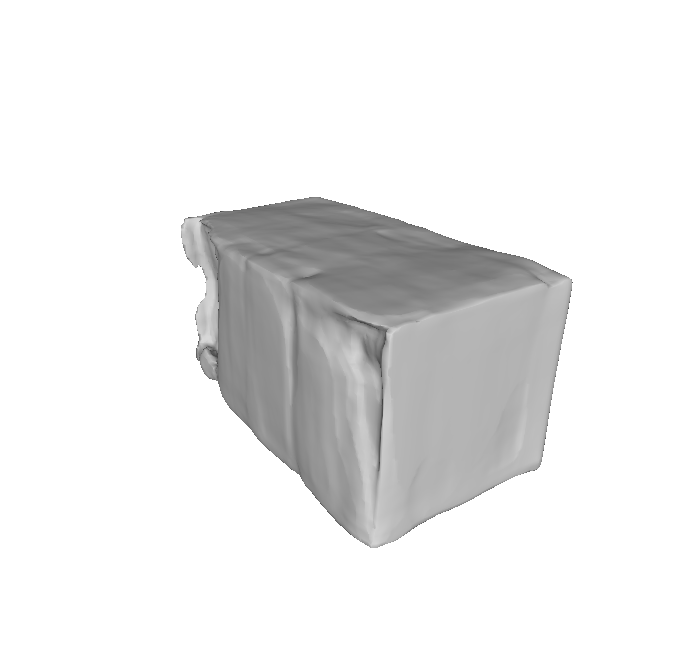} &
\includegraphics[scale=0.15]{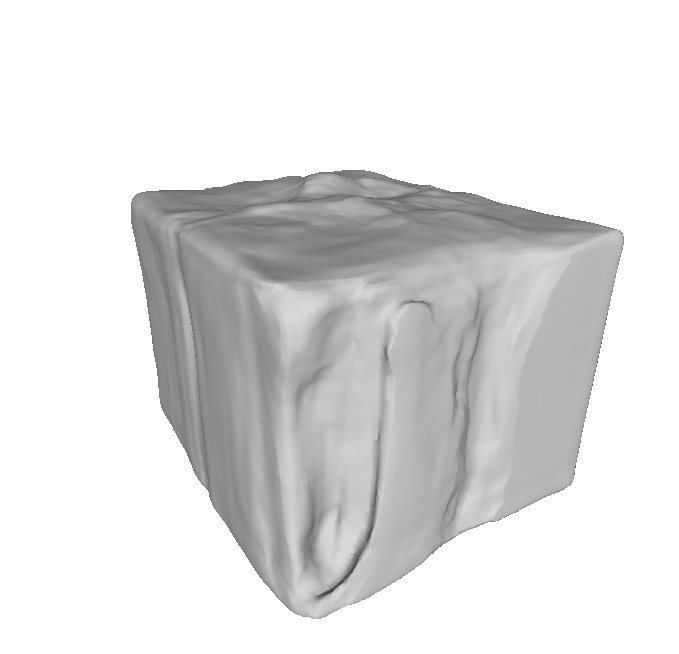} &
 \\   \\ \hline

\end{tabular}%
}
\caption{VA best model, maximize IoU}
\end{table*}

\begin{figure*}
  \centering
  \graphicspath{ {./images/} }
  \includegraphics[width=1\textwidth]{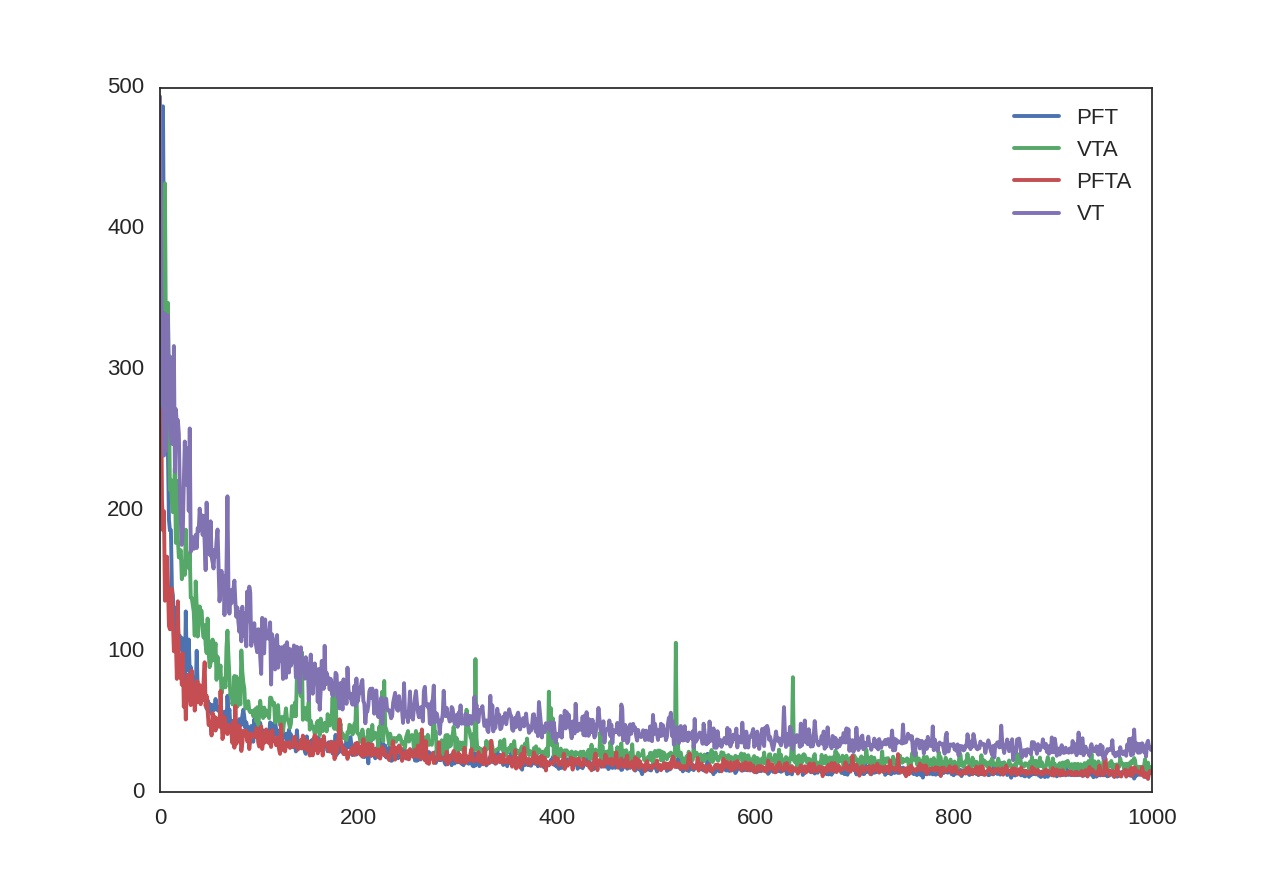}
   \caption{Initial training curves for the 4 categories. The vanilla training with aligned buildings shows to have more "wiggles" during training,  we trained with batch of 32. The pretrain and fine tune models are showing better training curves}
  
\end{figure*}

 \begin{figure}[hbt]
  \centering
  \graphicspath{ {./images/} }
  \includegraphics[width=1\textwidth]{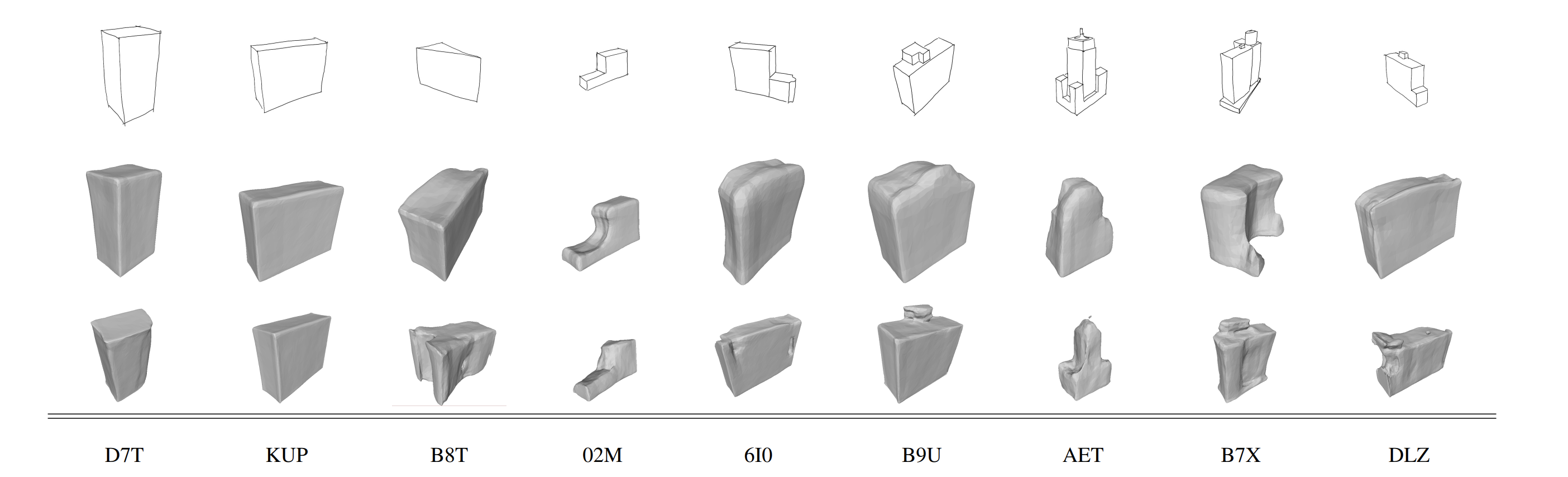}
  \caption{Tests with real sketches and their reconstruction. First VTARN and then VTAR. VTAR, with CBN, produces more accurate representation. Adding the regularization improved the qualitative analysis.}

\end{figure}

 \begin{figure}
  \centering
  \graphicspath{ {./images/} }
  \includegraphics[width=0.9\textwidth]{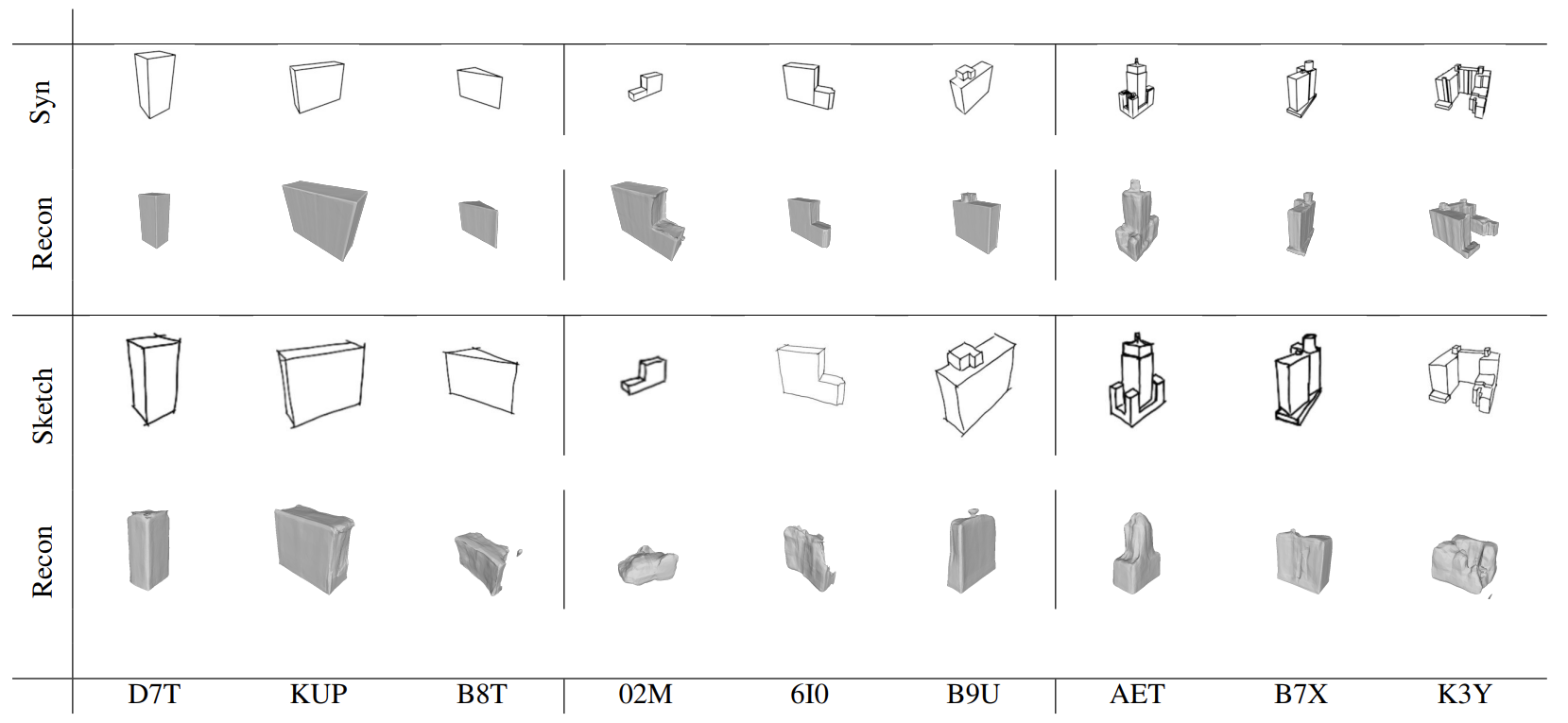}
  \caption{Best reconstruction with synthetic sketch and also with real sketches draw from a similar point of view. This stroke ablation study highlights the several issues for real-sketch reconstruction. It is extremely sensitive to line thickness, background, image size and many other features extract from the convolution }

\end{figure}

\begin{table*}[t]
\begin{minipage}{1.0\linewidth}
    \centering
    \resizebox{\textwidth}{!}{
    \begin{tabular}{c|cccccccccc}
        \textbf{Experiment} & \textbf{Acc}$\uparrow$ & \textbf{Acc2} $\uparrow$ & \textbf{C-L1}$\downarrow$ & \textbf{C-L2}$\downarrow$ & \textbf{Compl}$\uparrow$ & \textbf{Compl2} $\uparrow$& \textbf{IoU} $\uparrow$ & \textbf{N} $\uparrow$  & \textbf{N Acc} $\uparrow$ & \textbf{N Compl} $\uparrow$ \\ \hline
        PFT & 0.0401 & 0.0049 & 0.0362 & 0.0040 & 0.0324 & 0.0030 & 0.6396 & 0.7788 & 0.7726 & 0.7850 \\
        VT & 0.0374 & 0.0041 & 0.0359 & 0.0037 & 0.0344 & 0.0032 & 0.6500 & 0.7936 & 0.7941 & 0.7931 \\
        \hline
        PFTA & 0.0264 & 0.0025 & \textbf{0.0242} & \textbf{0.0019} & 0.0221 & 0.0014 & \textbf{0.7369} &  0.8438 &  0.84385 &  0.8438 
        \\
        VTA & 0.0289 & 0.0029 & 0.0268 & 0.0024 & 0.0247 & 0.0018 & 0.7223 & 0.8427 & 0.8464 & 0.8391 
        \\
        \hline
        VTAR  & 0.0275 & 0.0025 & 0.0260 & 0.0021 & 0.0246 & 0.0017 & 0.72098 & 0.8448 &  0.8490 & 0.8407 \\
        VTARN & 0.0285 & 0.0026 & 0.0272 & 0.0022 & 0.0259 & 0.0019 & 0.72062 &  \textbf{0.8525} &  \textbf{0.8606} & \textbf{0.8445}
        \\
        \hline
        VT MobileNetV3  & \textbf{0.0553} & \textbf{0.0083} & 0.0506 & 0.0070 & \textbf{0.0458} & \textbf{0.0058} & 0.5618 & 0.7262 & 0.7223 & 0.7301 \\\hline

    \end{tabular}}
    \caption{We identified that these metrics are not well correlated with the physical appearance and practical performance. We noticed overall better qualitative performance with data augmentation for "random" rotations. VTAR stands for VTA with L2 regularization and instead VTARN = VTAR without the CBN}

\end{minipage}
\end{table*}


\twocolumn

\begin{figure}
  \centering
  \graphicspath{ {./images/} }
  \includegraphics[width=0.4\textwidth]{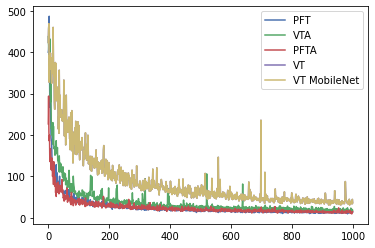}
   \caption{Initial training curves for the 4 categories. The vanilla training with aligned buildings shows to have more "wiggles" during training,  we trained with batch of 32. The pretrain and fine tune models are showing better training curves}

\end{figure}

\begin{figure}
  \centering
  \graphicspath{ {./images/} }
  \includegraphics[trim={0 0 0 1.2cm},clip,width=0.5\textwidth]{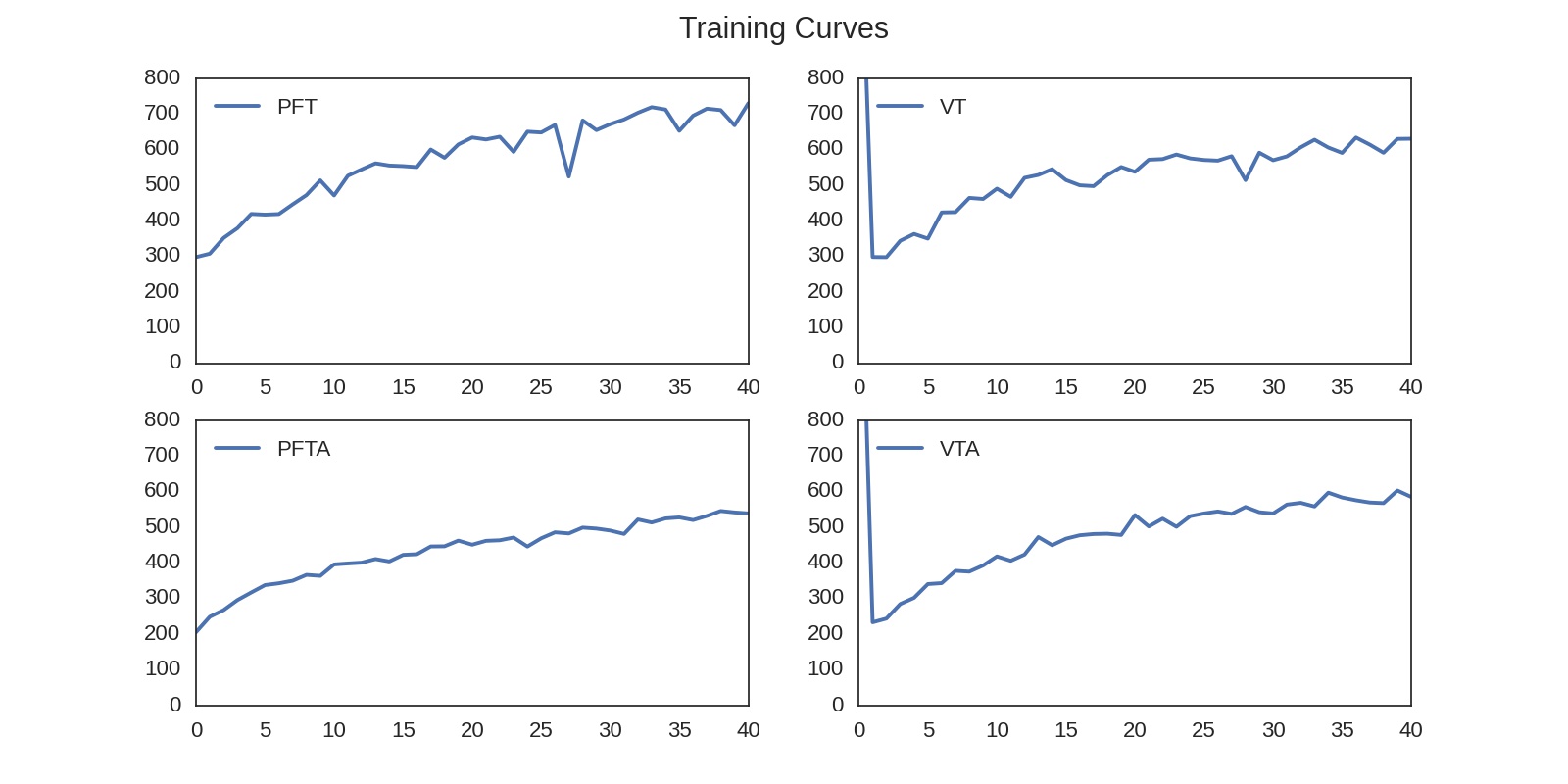}
  \caption{Validation loss showing overfitting.}
  \label{fig:val}
\end{figure}

  

\begin{figure}[hb]
  \centering
  \graphicspath{ {./images/} }
  \includegraphics[width=0.45\textwidth]{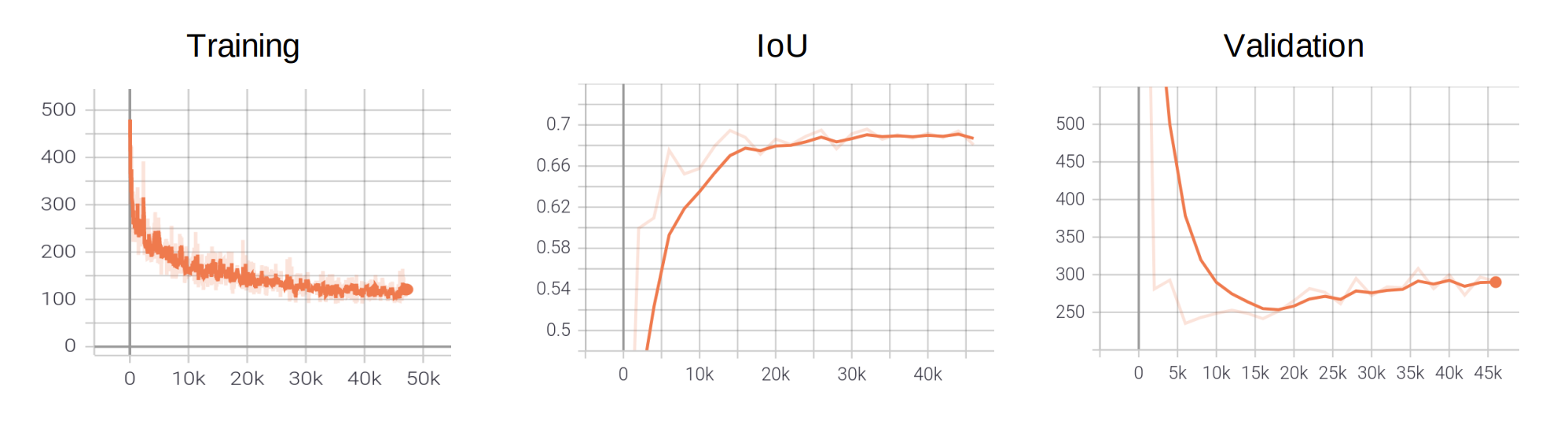}
   \caption{Initial training, IoU, validation curves with overfitting - Supplementary Material for more analysis.}

\end{figure}



\begin{table}[h]
        \resizebox{\columnwidth}{!}{
        \begin{tabular}[b]{c|ccccc}
        \textbf{Trials}  & $IoU_{points}$ $\uparrow$ & $IoU_{voxels}$ $\uparrow$ & $e_{rec}$ $\downarrow$ \\
        \hline

        Vanilla Training                   & 0.649     & 0.225           & 518.54 \\
        
        Pretrain Fine Tuning               & 0.639     & \textbf{0.226}          & 967.4 \\
        
        \hline
        
        VT Aligned                & 0.720      &  0.087           &  \textbf{241.25}  \\

        PFT Aligned           & \textbf{0.736}      & 0.086            &  424.07 \\
        
        \end{tabular}}
        \caption{Reconstruction quality evaluation for the different models. The IoU voxel is sensitive to the alignment. The model performances for IoU points and reconstruction error. }
\label{tab:analysis}
\end{table}

In the trials presented in Table \ref{tab:analysis}, the Pre-Trained Fine Tune Model with the aligned building provides the best results for CD, IoU and Normals, but worst in Accuracy and Completion, making it harder to evaluate.

\begin{figure}[h]
  \centering
  \graphicspath{ {./images/} }
  \includegraphics[width=0.4\textwidth]{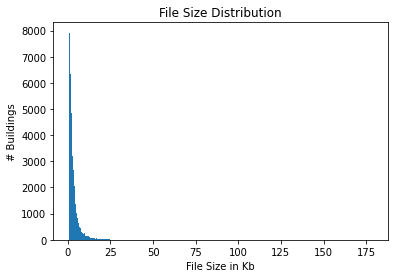}
   \caption{The $x$ axis has the file size and $y$ axis the number of shapes. In total, they represent the 47k model size distribution. Here "GUF7" is the largest model with 180kb, and "CEW2" is the smallest is just a few bites.}
   \label{fig:reconstruction}
\end{figure}

\begin{table}[hb] 
\begin{minipage}{\columnwidth}
\begin{center}
\begin{tabular}{ccc}
  
  \textbf{Small} & \textbf{Medium} & \textbf{Large} \\ 
   \includegraphics[scale=0.1]{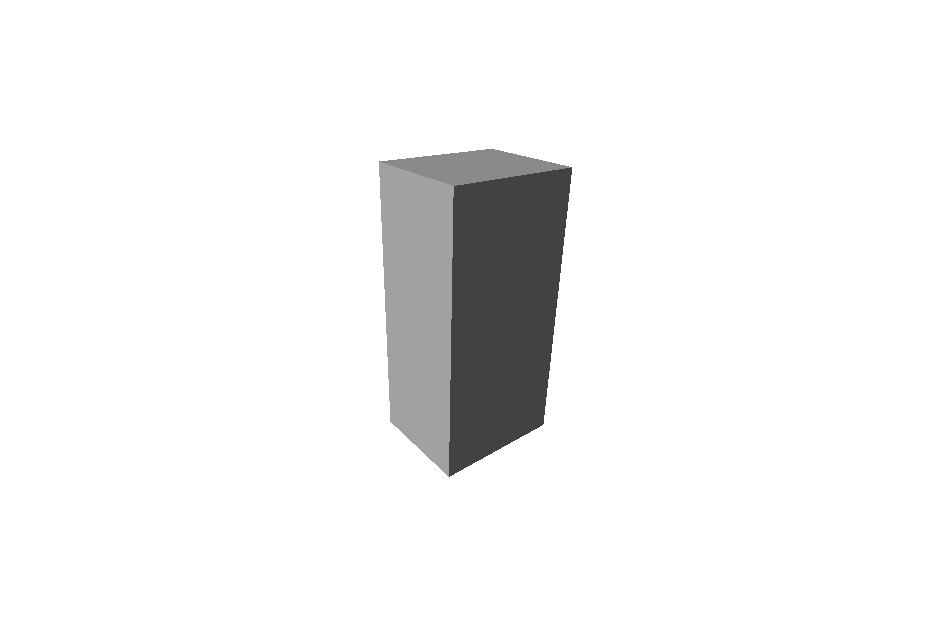}  &  \includegraphics[scale=0.1]{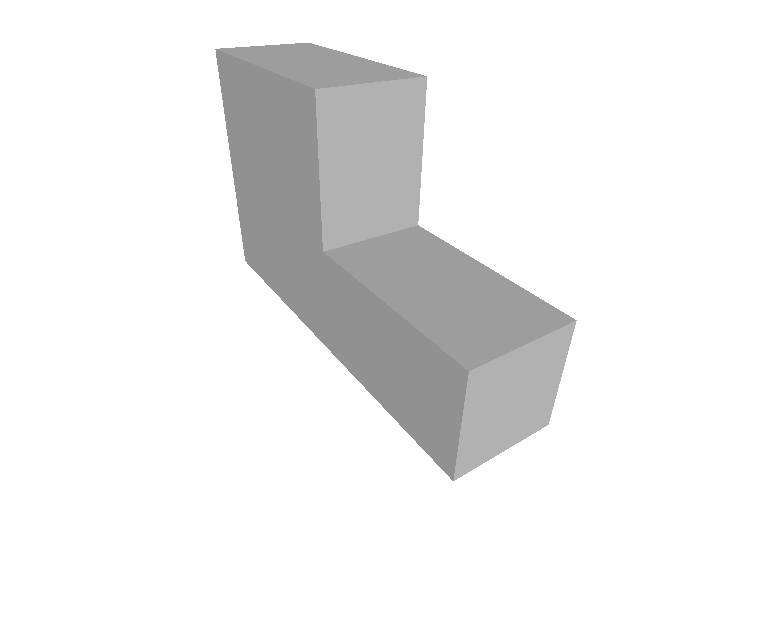} & 
   \includegraphics[scale=0.1]{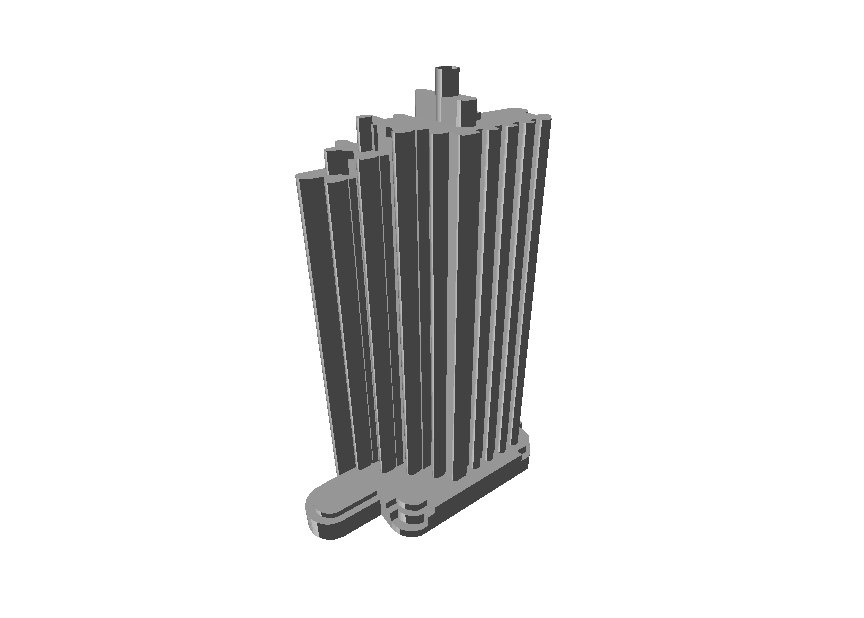} \\ 
   \includegraphics[scale=0.1]{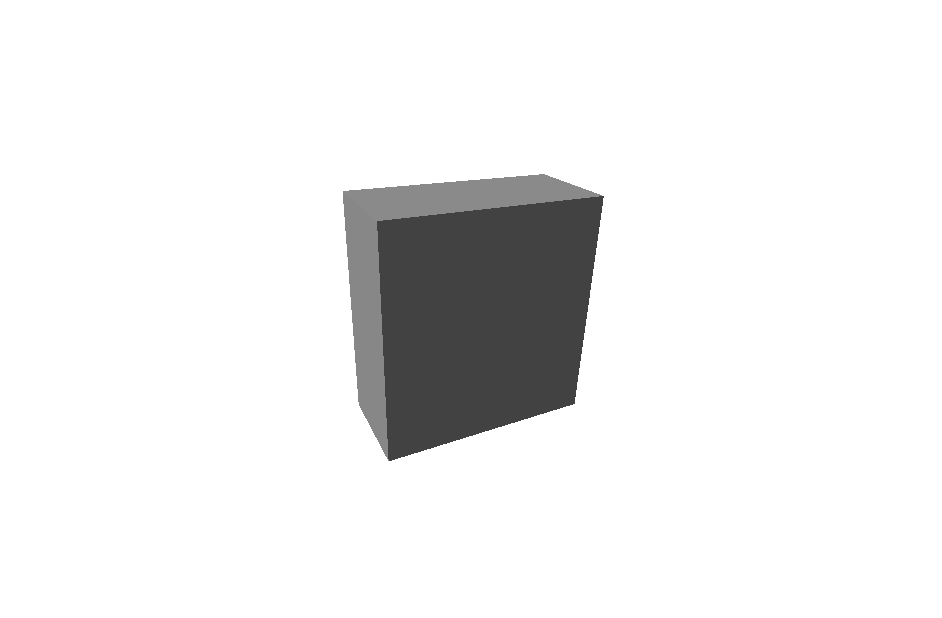}  &  \includegraphics[scale=0.1]{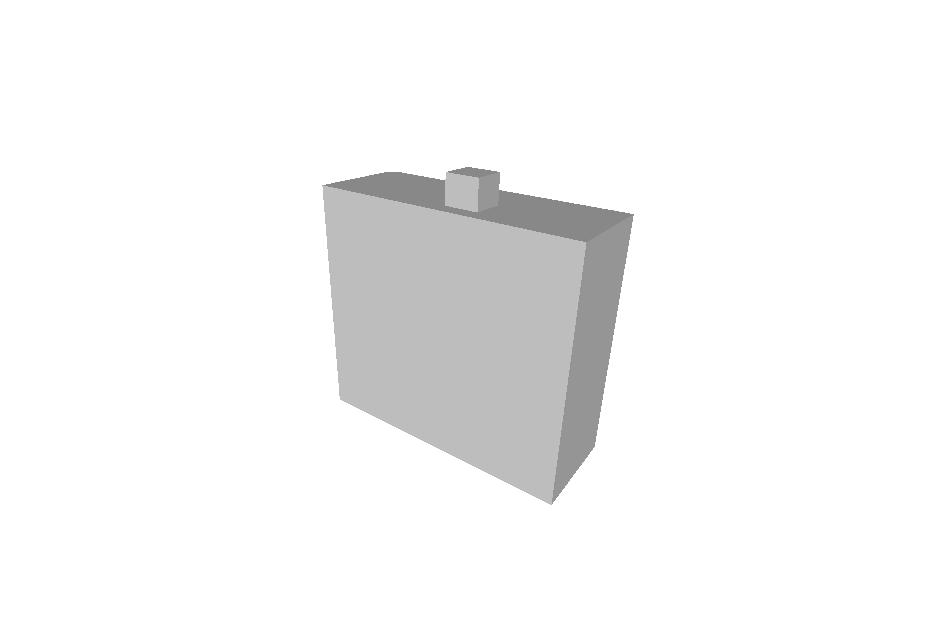} & 
   \includegraphics[scale=0.1]{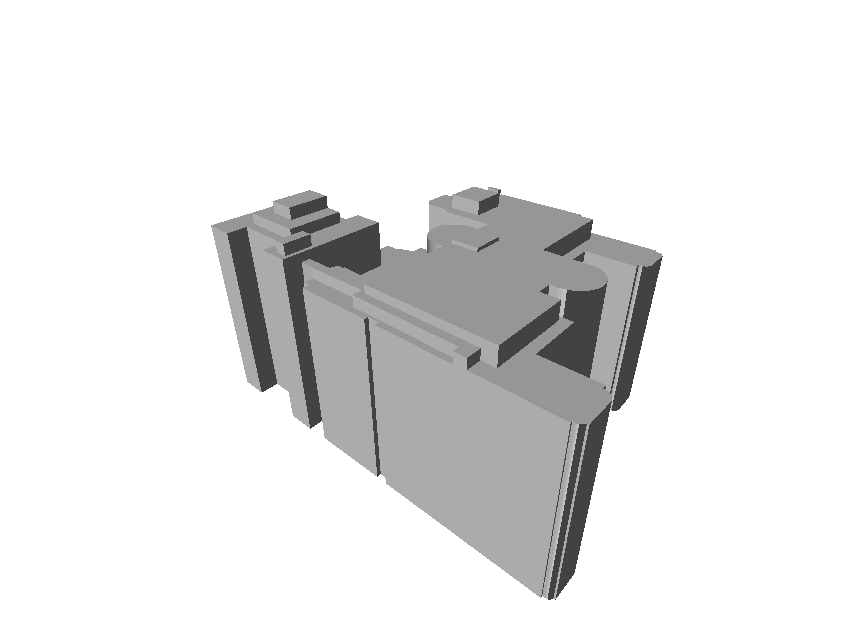}  \\ 
   \includegraphics[scale=0.1]{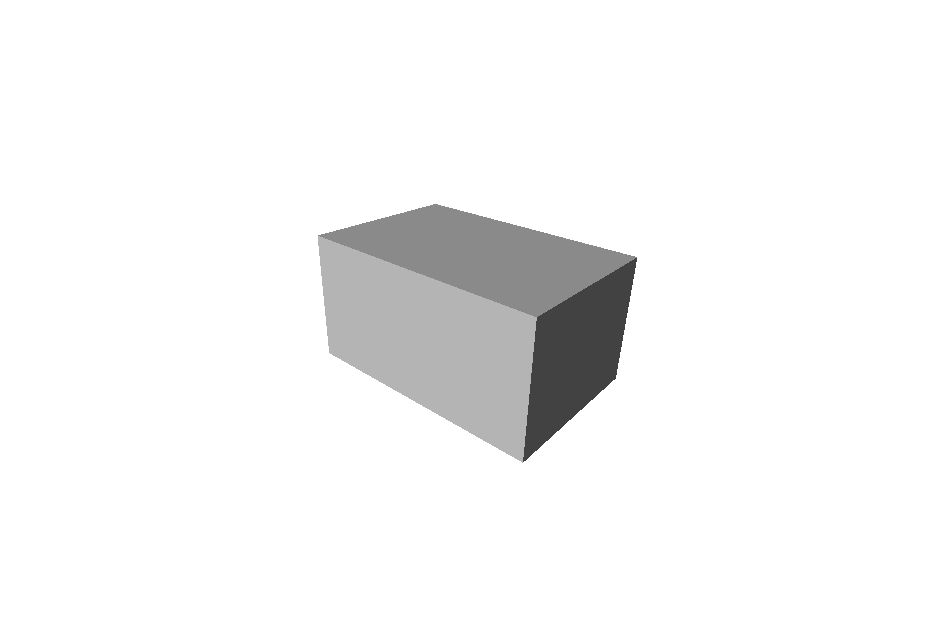}  &  \includegraphics[scale=0.1]{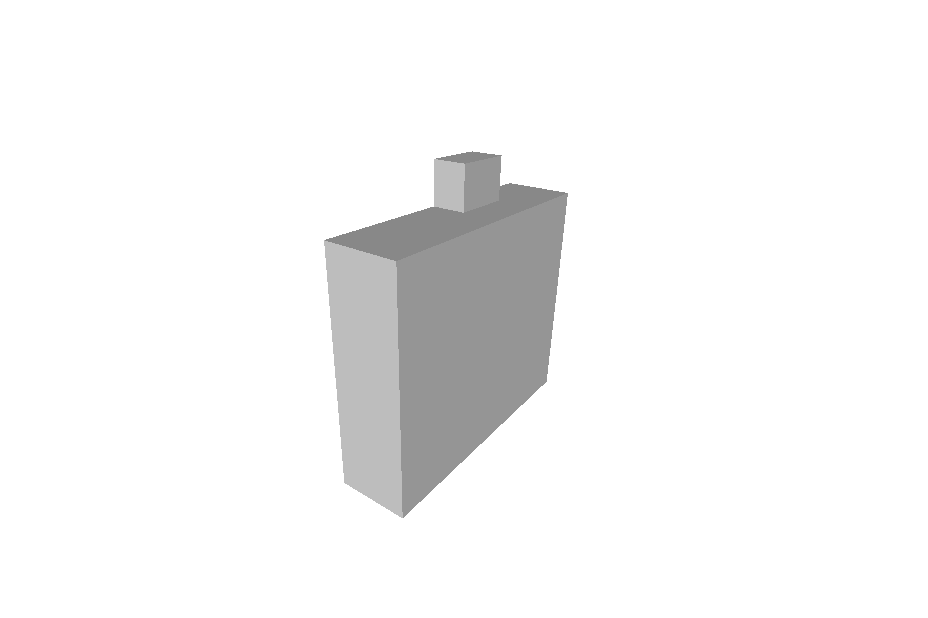} & 
   \includegraphics[scale=0.1]{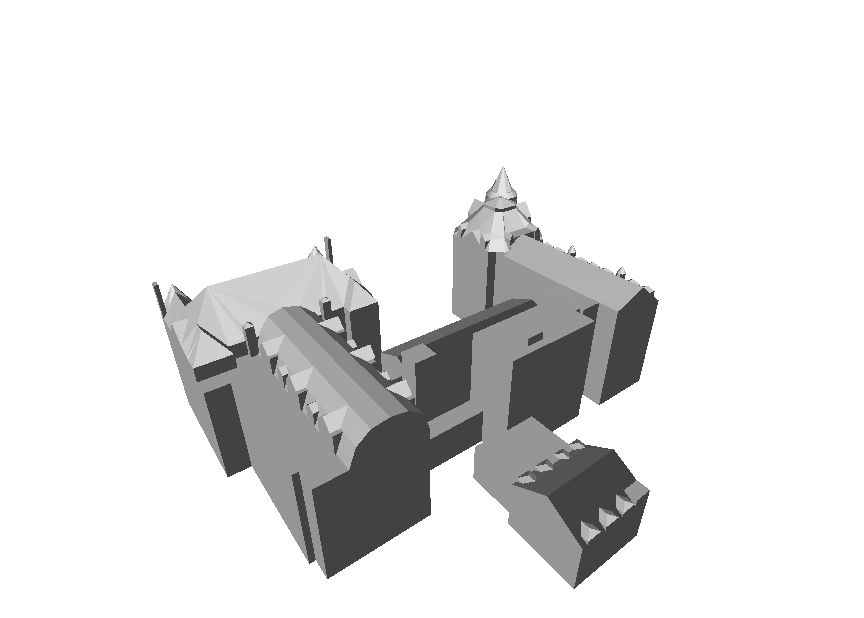}  \\ 
   \includegraphics[scale=0.1]{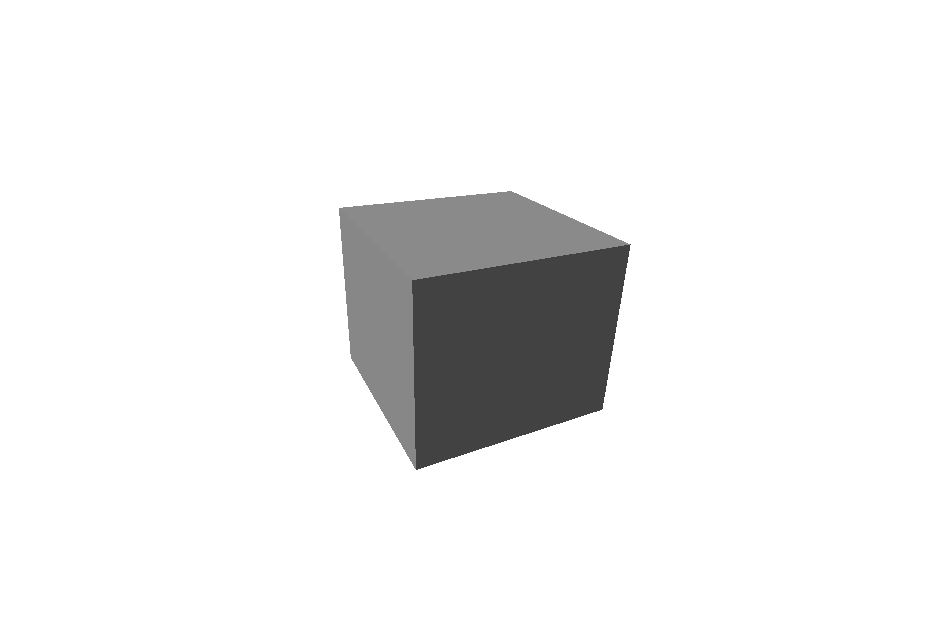}  &  \includegraphics[scale=0.1]{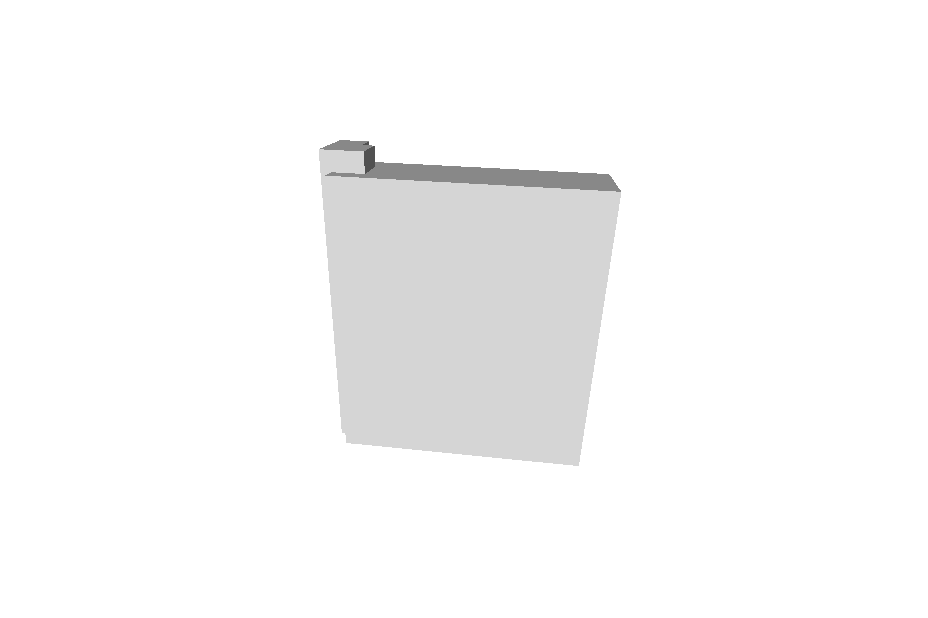} & 
   \includegraphics[scale=0.1]{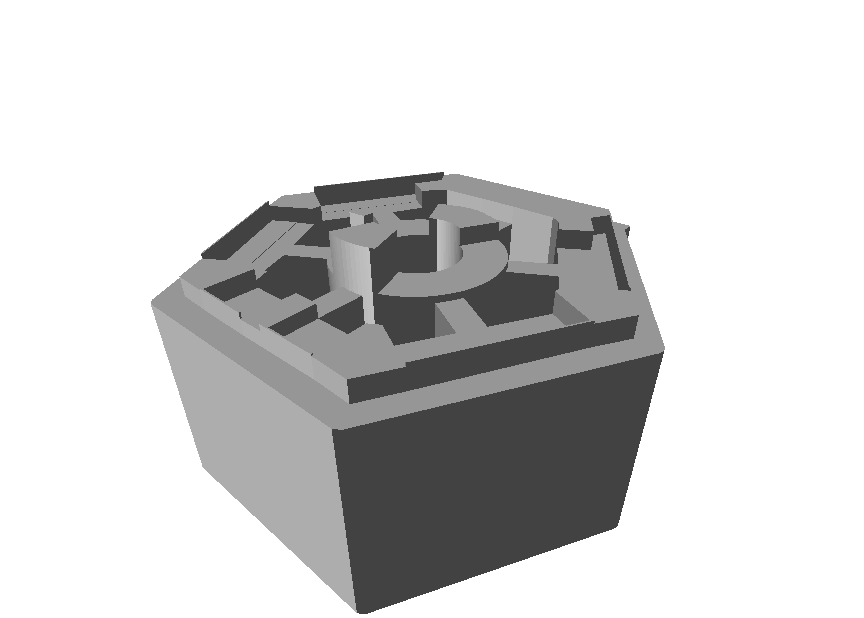}  \\ 
   \includegraphics[scale=0.1]{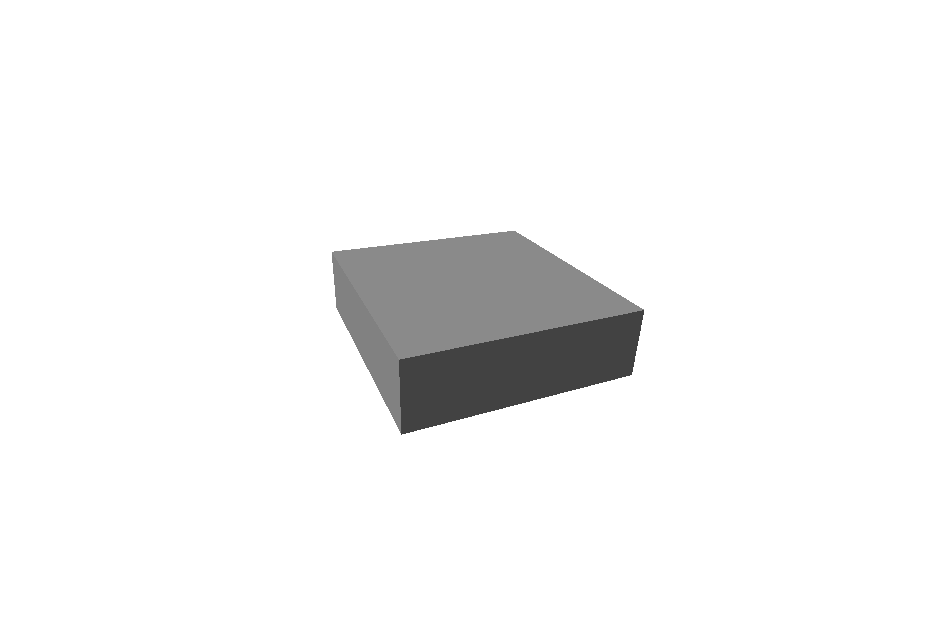}  &  \includegraphics[scale=0.1]{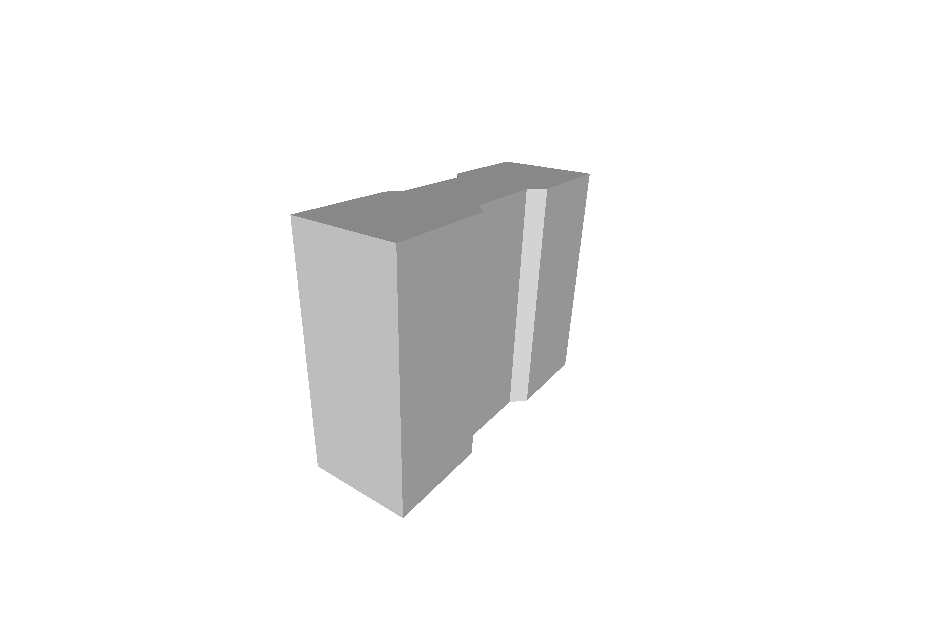} & 
   \includegraphics[scale=0.1]{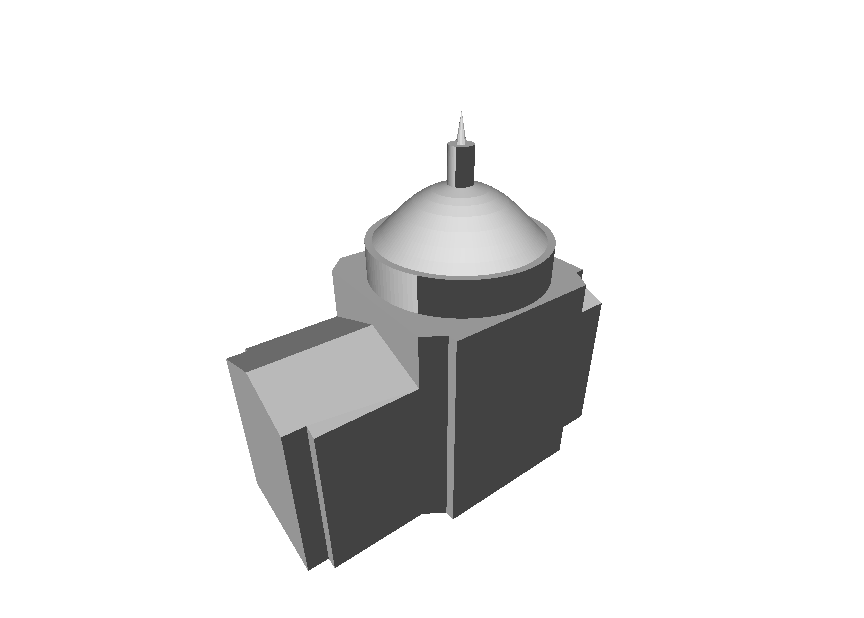}  \\ 
\includegraphics[scale=0.15]{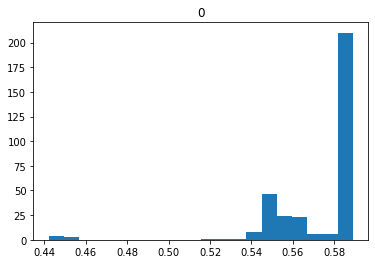}  &  \includegraphics[scale=0.15]{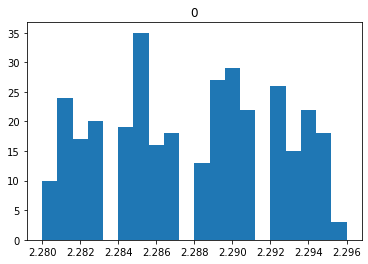} & 
   \includegraphics[scale=0.15]{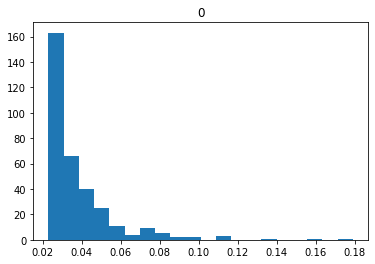}  
  
\end{tabular}
\end{center}
\caption{Sample of the 1k models based on file size with their respective distribution. The filename of the model identifies the building location}
\bigskip\centering
\end{minipage}
\end{table}

\end{document}